\documentclass[10pt,journal,compsoc]{IEEEtran}
\ifCLASSOPTIONcompsoc
  \usepackage[nocompress]{cite}
\else
  \usepackage{cite}
\fi
\ifCLASSINFOpdf
\else
\fi
\usepackage{times}
\usepackage{epsfig}
\usepackage{graphicx}
\usepackage{amsmath}
\usepackage{amssymb}

\usepackage{multirow}
\usepackage{graphicx}
\usepackage{subfig}
\usepackage{array}
\usepackage[colorlinks]{hyperref}
\usepackage{tablefootnote}
\usepackage{threeparttable}
\usepackage{pifont}% http://ctan.org/pkg/pifont
\newcommand{\cmark}{\ding{51}}%
\newcommand{\xmark}{\ding{55}}%

\newcommand{\etal}{\emph{et al. }}

\newcommand{\ie}{\emph{i.e., }}

\begin{document}
\title{A Lightweight Optical Flow CNN \textemdash \\Revisiting Data Fidelity and Regularization}

\author{Tak-Wai~Hui,
        Xiaoou~Tang,~\IEEEmembership{Fellow,~IEEE,}
        and~Chen Change Loy,~\IEEEmembership{Senior Member,~IEEE}
\IEEEcompsocitemizethanks{\IEEEcompsocthanksitem T.-W. Hui and X. Tang are with the Department of Information Engineering, The Chinese University of Hong Kong, Sha Tin, Hong Kong.\protect\\
E-mails: \{twhui, xtang\}@ie.cuhk.edu.hk
\IEEEcompsocthanksitem C. C. Loy is with the School of Computer Science and Engineering, Nanyang Technological University, Singapore.\protect\\
E-mail: ccloy@ntu.edu.sg}
}

\IEEEtitleabstractindextext{
\begin{abstract}
Over four decades, the majority addresses the problem of optical flow estimation using variational methods. With the advance of machine learning, some recent works have attempted to address the problem using convolutional neural network (CNN) and have showed promising results. FlowNet2~\cite{Ilg17}, the state-of-the-art CNN, requires over 160M parameters to achieve accurate flow estimation. Our LiteFlowNet2 outperforms FlowNet2 on Sintel and KITTI benchmarks, while being 25.3 times smaller in the model size and 3.1 times faster in the running speed.
LiteFlowNet2 is built on the foundation laid by conventional methods and resembles the corresponding roles as data fidelity and regularization in variational methods. 
We compute optical flow in a spatial-pyramid formulation as SPyNet~\cite{Ranjan17} but through a novel lightweight cascaded flow inference. It provides high flow estimation accuracy through early correction with seamless incorporation of descriptor matching. Flow regularization is used to ameliorate the issue of outliers and vague flow boundaries through feature-driven local convolutions. Our network also owns an effective structure for pyramidal feature extraction and embraces feature warping rather than image warping as practiced in FlowNet2 and SPyNet. 
Comparing to LiteFlowNet~\cite{Hui18-arxiv}, LiteFlowNet2 improves the optical flow accuracy on Sintel Clean by 23.3\%, Sintel Final by 12.8\%, KITTI 2012 by 19.6\%, and KITTI 2015 by 18.8\%, while being 2.2 times faster. Our network protocol and trained models are made publicly available on \url{https://github.com/twhui/LiteFlowNet2}.
\end{abstract}

% Note that keywords are not normally used for peerreview papers.
\begin{IEEEkeywords}
Convolutional neural network, cost volume, deep learning, optical flow, regularization, spatial pyramid, and warping.
\end{IEEEkeywords}}

% make the title area
\maketitle

% To allow for easy dual compilation without having to reenter the
% abstract/keywords data, the \IEEEtitleabstractindextext text will
% not be used in maketitle, but will appear (i.e., to be "transported")
% here as \IEEEdisplaynontitleabstractindextext when the compsoc 
% or transmag modes are not selected <OR> if conference mode is selected 
% - because all conference papers position the abstract like regular
% papers do.
\IEEEdisplaynontitleabstractindextext
% \IEEEdisplaynontitleabstractindextext has no effect when using
% compsoc or transmag under a non-conference mode.

% For peer review papers, you can put extra information on the cover
% page as needed:
% \ifCLASSOPTIONpeerreview
% \begin{center} \bfseries EDICS Category: 3-BBND \end{center}
% \fi
%
% For peerreview papers, this IEEEtran command inserts a page break and
% creates the second title. It will be ignored for other modes.
\IEEEpeerreviewmaketitle

%-------------------------------------------------------------------------
\IEEEraisesectionheading{\section{Introduction}}
\IEEEPARstart{O}{ptical flow}, which refers to the point correspondence across a pair of images, is induced by the spatial motion at any image position. Due to the well-known aperture problem, optical flow cannot be directly measured. The partial observability of optical flow is the major reason that makes it a challenging problem. 
The optical flow problem has attracted many attentions since the seminal works by Horn and Schunck~\cite{Horn81}, and Lucas and Kanade~\cite{Lucas81} about four decades ago. Most of the approaches estimate optical flow relying on an energy minimization method in a coarse-to-fine framework~\cite{Brox04,Papenberg06,Brox11}. Optical flow is refined iteratively using a numerical approach from the coarsest level towards the finest level by warping one of the images in the image pair towards the other using the flow estimate from the coarser level. The warping technique is theoretically justified to minimize the energy functional~\cite{Brox04, Papenberg06}. On the other hand, normal flow which is directly measurable is more ready for motion estimation~\cite{Hui13,Hui13a,Hui15}.

FlowNet~\cite{Dosovitskiy15} and FlowNet2~\cite{Ilg17}, are the pioneering works using convolutional neural network (CNN) for optical flow estimation. Their performances especially the successor are approaching to the state-of-the-art energy minimization approaches, while the speed is several orders of magnitude faster. To push the envelop of accuracy, FlowNet2 is designed as a cascade of variants of FlowNet, \ie FlowNetC and FlowNetS. Each network in the cascade refines the preceding flow field by contributing on the flow adjustment between the first image and the warped second image. The model, as a result, comprises over 160M parameters and has a slow runtime, which could be formidable in many applications. Another work, SPyNet~\cite{Ranjan17}, uses a spatial pyramid network with only 1.2M parameters by adopting image warping in each pyramid level. Nonetheless, its performance can only match that of FlowNet but not FlowNet2. 

\begin{figure}[t]
\centering
   \includegraphics[width=\linewidth]{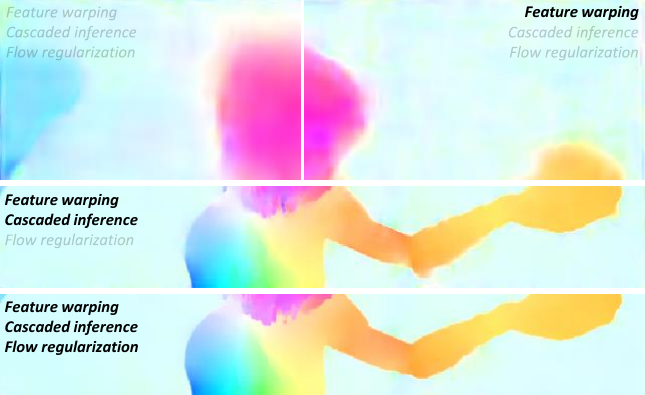}
\caption{Examples demonstrate the effectiveness of the proposed components in LiteFlowNet for i) feature warping, ii) cascaded flow inference, and iii) flow regularization. Enabled components are indicated with bold black fonts.}
\label{fig:overview}
\vspace{-0.5em}
\end{figure}

\begin{figure*}[ht]
\centering
   \includegraphics[width=\linewidth]{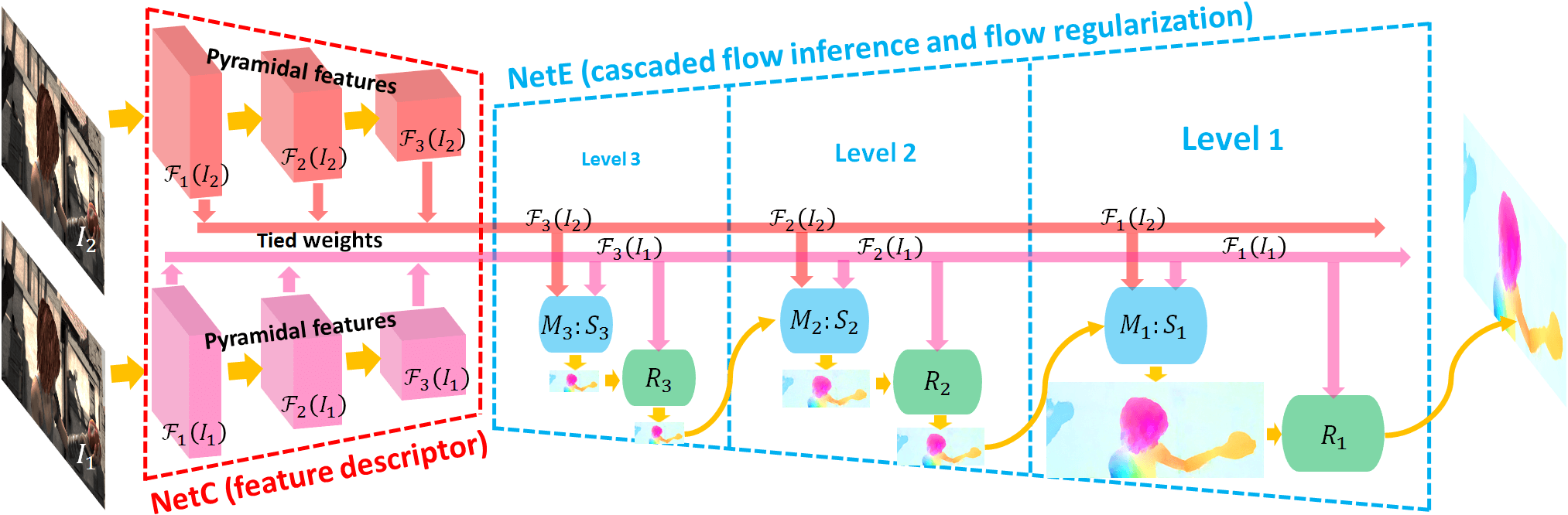}
\caption{The network structure of LiteFlowNet. For the ease of representation, only a design of 3-level pyramid is shown. Given an image pair ($I_{1}$ and $I_{2}$), NetC generates two pyramids of high-level features ($\mathcal F_{k}(I_{1})$ in pink and $\mathcal F_{k}(I_{2})$ in red, $k \in [1, 3]$). 
NetE yields multi-scale flow fields such that each of them is generated by a cascaded flow inference module $M$:$S$ (in blue color, including a descriptor matching unit $M$ and a sub-pixel refinement unit $S$) and a regularization module $R$ (in green color). 
Flow inference and regularization modules correspond to data fidelity and regularization terms in conventional energy minimization methods, respectively.}
\label{fig:network structure}
\vspace{-0.5em}
\end{figure*}

FlowNet2~\cite{Ilg17} and SPyNet~\cite{Ranjan17} showed the potential of solving the optical flow problem by using CNNs. Our earlier work, LiteFlowNet~\cite{Hui18}, is inspired by their successes, but we further drill down some of the key elements of solving the flow problem by adopting \textit{data fidelity} and \textit{regularization} in classical variational methods to CNN more closely. In this work, we provide more details on the correspondences between conventional methods and optical flow CNNs. We also present LiteFlowNet2 that has a better flow accuracy and a faster runtime by optimizing the network architecture and training protocols of LiteFlowNet. 

In the following, we first discuss the motivations, namely i) data fidelity, ii) image warping, and iii) regularization, from classical variational methods on the design of LiteFlowNet. Then, we highlight the more specific differences between our design and the state-of-the-art optical flow CNNs.

\vspace{0.1cm}
\noindent 
\textbf{Data Fidelity.}
Point correspondence across two images is generally constrained by the classical brightness constancy~\cite{Horn81}. Gradient~\cite{Brox04} and higher-order brightness constancy~\cite{Papenberg06} assumptions are also widely used in the literature. The above constancy assumptions are collectively known as data fidelity and are often combined to form a hybrid data term~\cite{Brox04,Xu12}. Although different matching quantities are proved to be useful in solving the optical flow problem, finding a correct proportion of their contributions in the hybrid data term is non-trivial and requires a highly engineered data fusion model~\cite{Kim13}. An improper mixture of the brightness and gradient terms can severely affect the performance~\cite{Xu12}.   
To avoid the aforementioned difficulties, feature descriptors that are not explicitly defined are learned in variational setting~\cite{Sun08}. We use a CNN to train a \textit{pyramidal feature descriptor} (\ie a feature encoder)~\cite{Dosovitskiy15,Ilg17} which resembles data fidelity in variational methods and is prepared for establishing robust point correspondence later. Specifically, a given image pair is transformed from the spatial domain to the learned feature space in the form of two pyramids of multi-scale high-dimensional feature maps. 

\vspace{0.1cm}
\noindent
\textbf{Image Warping.}
It is proved that image warping effectively minimizes an energy functional by using a numerical method in a coarse-to-fine framework~\cite{Brox04, Papenberg06}. Intuitively, at each iteration the numerical solver displaces every pixel value of the second image in the image pair according to the constraints imposed in the functional so that the warped image has a visual appearance close to the first image. Image warping is practiced in FlowNet2~\cite{Ilg17} and SPyNet~\cite{Ranjan17} between cascaded networks and pyramid levels, respectively.
However, warping an image and then generating the feature maps of the warped image as the above CNN-based methods are two ordered steps. We find that the two steps can be reduced to a single one by directly warping the feature maps of the second image, which have been provided by the feature encoder. This one-step \textit{feature warping} (f-warp) process reduces the more discriminative feature-space distance instead of the RGB-space distance between the two images. This makes LiteFlowNet more powerful and efficient in addressing the optical flow problem. To this end, we use the spatial transformer~\cite{Jaderberg15} for the warping.

\vspace{0.1cm}
\noindent
\textbf{Regularization.}
Merely using data fidelity for flow estimation is an ill-posed problem~\cite{Horn81}. One example is the one-to-many point correspondences in homogeneous regions of an image pair. With the co-occurrence between motion boundaries and intensity edges, flow estimate is often smoothed by an anisotropic image-driven regularization~\cite{Werlberger09, Xu12}. However, the image-driven strategies are prone to over-segmentation artifacts in the textured image regions since image edges do not necessarily correspond to flow edges. More advanced methods overcome the previous shortcomings through the use of an anisotropic image- and flow-driven regularization~\cite{Sun08} and a complementary regularizer~\cite{Zimmer11}.
With the motivation to establish robust point correspondence in the learned feature space, we generalize the use of regularization from the spatial space to the feature space. This allow the flow field to be regularized by a \textit{feature-driven local convolution} (f-lconv) at each pyramid level. The kernels of such a local convolution are adaptive to the pyramidal features from the encoder, flow estimate, and occlusion probability map. This makes the flow regularization to be both flow- and image-aware. We name it as the feature-driven local convolution layer in order to distinguish it from the local convolution (lconv) layer of which filter weights are locally fixed in conventional CNNs~\cite{Taigman14}. We use the feature-driven convolution~\cite{Brabandere16} in our framework to regularize flow fields.

\begin{table*}[ht]
\small
\centering
\caption{A comparison of the major components used in the state-of-the-art optical flow CNNs. (Notes: $^{1}$We use the convention that flow field at level 1 has the same spatial resolution as the given image pair. $^{2}$Flow inference from levels 7 to 3 is performed in each of the stacking networks except the fusion network. Flow fields resulting from FlowNet2-CSS and FlowNet2-SD are upsampled by a factor 4 (\ie from level 3 to level 1) and then used as the inputs to the fusion network. $^{3}$The authors excluded the use of residual connections in the publicly released model.)}\label{tab: flow cnn comparison}
\scalebox{0.83}{
\begin{tabular}{ccccccc}
\hline     
\multicolumn{1}{|c|}{}
&\multicolumn{1}{c|}{FlowNetS\cite{Dosovitskiy15}} 		     		
&\multicolumn{1}{c|}{FlowNetC\cite{Dosovitskiy15}}				
&\multicolumn{1}{c|}{FlowNet2\cite{Ilg17}} 
&\multicolumn{1}{c|}{SPyNet\cite{Ranjan17}} 
&\multicolumn{1}{c|}{PWC-Net\cite{Sun18}}  
&\multicolumn{1}{c|}{LiteFlowNet\cite{Hui18}} \\
\hline
\multicolumn{1}{|l|}{Architecture}	
&\multicolumn{1}{c|}{U-Net}	
&\multicolumn{1}{c|}{U-Net}					
&\multicolumn{1}{c|}{U-Net}
&\multicolumn{1}{c|}{spatial pyramid}
&\multicolumn{1}{c|}{spatial pyramid}  
&\multicolumn{1}{c|}{spatial pyramid} \\

\multicolumn{1}{|l|}{Stacking Multiple Networks}		 			
&\multicolumn{1}{c|}{\xmark}	
&\multicolumn{1}{c|}{\xmark}					
&\multicolumn{1}{c|}{5 networks} 
&\multicolumn{1}{c|}{\xmark}
&\multicolumn{1}{c|}{\xmark}   
&\multicolumn{1}{c|}{\xmark} \\

\multicolumn{1}{|l|}{Multi-Scale Flow Fields$^{1}$}					
&\multicolumn{1}{c|}{levels: 7 -- 3}	
&\multicolumn{1}{c|}{levels: 7 -- 3}					
&\multicolumn{1}{c|}{levels: 7 -- 1$^{2}$} 
&\multicolumn{1}{c|}{levels: 6 or 5 -- 1}  
&\multicolumn{1}{c|}{levels: 7 -- 3} 
&\multicolumn{1}{c|}{levels: 6 -- 2} \\

\multicolumn{1}{|l|}{Cost Volume}		
&\multicolumn{1}{c|}{\xmark}					
&\multicolumn{1}{c|}{single (long range)}						
&\multicolumn{1}{c|}{single (long range)} 
&\multicolumn{1}{c|}{\xmark}   
&\multicolumn{1}{c|}{multiple (short range)}
&\multicolumn{1}{c|}{multiple (short range)} \\

\multicolumn{1}{|l|}{Warping}							
&\multicolumn{1}{c|}{\xmark}			
&\multicolumn{1}{c|}{\xmark}			
&\multicolumn{1}{c|}{image (per network)} 
&\multicolumn{1}{c|}{image (per level)} 
&\multicolumn{1}{c|}{feature (per level)}  
&\multicolumn{1}{c|}{feature (per level)} \\

\multicolumn{1}{|l|}{Flow Inference (per level)}			
&\multicolumn{1}{c|}{direct}				
&\multicolumn{1}{c|}{direct}						
&\multicolumn{1}{c|}{direct} 
&\multicolumn{1}{c|}{residual} 
&\multicolumn{1}{c|}{direct$^{3}$}  
&\multicolumn{1}{c|}{cascaded \& residual} \\

\multicolumn{1}{|l|}{Flow Regularization}			
&\multicolumn{1}{c|}{\xmark}				
&\multicolumn{1}{c|}{\xmark}						
&\multicolumn{1}{c|}{\xmark} 
&\multicolumn{1}{c|}{\xmark}   
&\multicolumn{1}{c|}{\xmark}
&\multicolumn{1}{c|}{per level} \\
\hline
\end{tabular}}
\vspace{-0.5em}
\end{table*}

\vspace{0.1cm}
\noindent
\textbf{Our Design.}
The proposed network, dubbed LiteFlowNet~\cite{Hui18}, consists of a multi-scale feature encoder and a multi-scale flow decoder as shown in Figure~\ref{fig:network structure}. The encoder maps a given image pair, respectively, into two pyramids of multi-scale high-dimensional features~\cite{Dosovitskiy15,Ilg17}. The decoder then estimates optical flow in a coarse-to-fine framework~\cite{Ranjan17}. Specifically, the decoder infers a flow field by selecting and using the features of the same resolution from the encoder at each pyramid level. This design leads to a lighter and a more efficient network compared to FlowNet~\cite{Dosovitskiy15} and FlowNet2~\cite{Ilg17} that adopt U-Net architecture~\cite{Ronneberger15} for flow inference. SPyNet~\cite{Ranjan17} uses a spatial pyramid network to infer a flow field at each pyramid level from the corresponding image pair in the image pyramid. On the contrary, our network separates the processes of feature extraction and flow estimation into encoder and decoder, respectively. This helps us to better pinpoint the bottleneck of accuracy and model size. Particularly, our decoder uses a pair of feature maps from the encoder for flow inference instead of using a pair of images.

At each pyramid level, we introduce a novel cascaded flow inference. Each of them has a f-warp layer to displace the feature maps of the second image towards the first image using the flow estimate from the previous level rather than image warping as practiced in FlowNet2~\cite{Ilg17} and SPyNet~\cite{Ranjan17}. Flow residue is computed to reduce the feature-space distance between the images.
This design is advantageous to the conventional design of using a single network for flow inference. 
First, the cascade progressively improves flow accuracy thus allowing an early correction of the estimate without passing more errors to the next pyramid level.
Second, this design allows seamless integration with descriptor matching. We assign a matching network to the first inference. Consequently, pixel-accuracy flow field can be generated first and then it is refined to sub-pixel accuracy in the subsequent inference network.
Since at each pyramid level the feature-space distance between the images has been reduced by the f-warp, a short searching range rather than a long searching range~\cite{Dosovitskiy15, Ilg17} is used to establish a cost volume. Besides, matching can be performed at sampled positions to aggregate a sparse cost volume. This effectively reduces the computational burden raised by the explicit matching. 
After the cascaded flow inference, the flow field is further regularized by a f-lconv layer.

The effectiveness of the aforementioned designs are depicted in Figure~\ref{fig:overview}. In summary, our contributions are in four aspects: 

\begin{enumerate}
\item We present a study to bridge the correspondences between the well-established principles in conventional methods for optical flow estimation and optical flow CNNs. 

\item More details of our earlier work LiteFlowNet~\cite{Hui18} are presented.

\item LiteFlowNet2, another lightweight convolutional network, is evolved from LiteFlowNet~\cite{Hui18} to better address the problem of optical flow estimation by improving flow accuracy and computation time. 

\item LiteFlowNet2 outperforms the state-of-the-art FlowNet2~\cite{Ilg17} on Sintel and KITTI benchmarks, while being 25.3 times smaller in the model size and 3.1 times faster in the runtime. The optical flow processing frequency of LiteFlowNet2 reaches up to 25 flow fields per second for an image pair in Sintel dataset with size $1024 \times 436$ on a NVIDIA GTX 1080 GPU. Our network protocol and trained models are made publicly available on \url{https://github.com/twhui/LiteFlowNet2}.
\end{enumerate}

%-------------------------------------------------------------------------
\section{Related Work}
%-------------------------------------------------------------------------
The problem of optical flow estimation has been widely studied in the literature since 1980s. A detailed review is beyond the scope of this work. Here, we briefly review some of the major approaches, namely variational, machine learning, and CNN-based methods.

\vspace{0.1cm}
\noindent \textbf{Variational Methods.} Since the pioneering work by Horn and Schunck~\cite{Horn81}, variational methods have dominated in the literature. Brox \etal address illumination changes by combining the brightness and gradient constancy assumptions~\cite{Brox04}. Brox \etal integrate rich descriptors into a variational formulation~\cite{Brox11}. In DeepFlow~\cite{Weinzaepfel13}, Weinzaepfel \etal propose to correlate multi-scale patches and incorporate this as the matching term in a functional. In PatchMatch Filter~\cite{Lu13}, Lu \etal establish dense correspondence using the superpixel-based PatchMatch~\cite{Barnes09}. Revaud \etal propose EpicFlow that uses externally matched flows as the initialization and then performs interpolation~\cite{Revaud15}. Zimmer \etal design the complementary regularization that exploits directional information from the constraints imposed in data term~\cite{Zimmer11}. Our network that infers optical flow and performs flow regularization is inspired by data fidelity and regularization in variational methods. 

\vspace{0.1cm}
\noindent \textbf{Machine Learning Methods.} Black \etal propose to represent complex image motion as a linear combination of the learned basis vectors~\cite{Black97}. Roth \etal formulates the prior probability of flow field as Field-of-Experts model~\cite{Roth05a} that captures higher order spatial statistics~\cite{Roth05b}. Sun \etal study the probabilistic model of brightness inconstancy in a high-order random field framework~\cite{Sun08}. Nir \etal represent image motion using the over-parameterization model~\cite{Nir08}. Rosenbaum \etal model the local statistics of optical flow using Gaussian mixtures~\cite{Rosenbaum13}. Given a set of sparse matches, Wulff \etal propose to regress them to a dense flow field using a set of basis flow fields (PCA-Flow)~\cite{Wulff15}. It can be shown that the parameterized model~\cite{Black97, Nir08, Wulff15} is related to the flow inference in CNNs.

\vspace{0.1cm}
\noindent \textbf{CNN-Based Methods.} A comparison of the major components used in the state-of-the-art optical flow CNNs is summarized in Table~\ref{tab: flow cnn comparison}. 
In FlowNet~\cite{Dosovitskiy15}, Dosovitskiy \etal use an optional post-processing step that involves energy minimization to reduce smoothing effect across flow boundaries. This process is not end-to-end trainable. On the contrary, we present an end-to-end approach that performs in-network flow regularization using a f-lconv layer, which plays a similar role as the regularization term in variational methods. 
In FlowNet2~\cite{Ilg17}, Ilg \etal introduce a huge network cascade (over 160M parameters) that consists of variants of FlowNet (FlowNetS and FlowNetC). The cascade improves flow accuracy with an expense of model size and computational complexity. 
A compact network termed SPyNet~\cite{Ranjan17} from Ranjan \etal uses a spatial pyramid network. It warps the second image toward the first one using the estimated flow field from the previous level. But the accuracy is below FlowNet2 (KITTI 2012~\cite{Geiger12}: 4.1 vs 1.8 measured in AEE, KITTI 2015~\cite{Menze15}: 35.07\% vs 11.48\% measured in Fl-all). On the contrary, LiteFlowNet infers a flow field at each pyramid level from the corresponding feature pair in the encoder and uses feature warping. LiteFlowNetX, a small-sized variant of our network, outperforms SPyNet while being 1.33 times smaller in the model size. Zweig \etal present a network to interpolate third-party sparse flows but requiring off-the-shelf edge detector~\cite{Zweig17}.
DeepFlow~\cite{Weinzaepfel13} that involves convolution and pooling operations is however not a CNN, since the ``filter weights" are non-trainable image patches. It uses correlation according to the terminology used in FlowNet.

A notable concurrent work to LiteFlowNet is PWC-Net~\cite{Sun18}, which is about \textbf{18 times smaller} than FlowNet2~\cite{Ilg17}. LiteFlowNet~\cite{Hui18}, a more lightweight CNN, is about \textbf{30 times smaller} than FlowNet2. Both of the works use the coarse-to-fine flow inference, feature warping, and cost volume for optical flow estimation, and are presented in CVPR 2018. However, there a number of distinctions between them. First, LiteFlowNet incorporates a cascaded flow inference to estimate residual flow at each pyramid level. Specifically, the pixel-level flow estimate that is generated by the cost-volume flow decoder is refined to the sub-pixel level. Second, flow fields resulting from the cascaded flow inference are further regularized by feature-driven local convolutions. Third, densely connected layers and feed-forwarding of feature maps from the previous level are not used in each pyramid level of the decoder. Fourth, LiteFlowNet is also benefited from the use of stage-wise training (more details in Section~\ref{sec:exp liteflownet}) to improve the optical flow accuracy and reduce the training time. These differences make LiteFlowNet to be more efficient in terms of the number of model parameters for solving the optical problem and therefore it attains a smaller model size than PWC-Net. 

An alternative approach for establishing dense correspondence is to match image patches. Zagoruyko \etal introduce to use CNN-feature matching~\cite{Zagoruyko15}. G\"uney \etal use feature representation and formulate optical flow estimation in MRF~\cite{Guney16}. Bailer \etal\cite{Bailer17} use multi-scale features and then perform feature matching as Flow Fields~\cite{Bailer15}. Although pixel-wise matching can establish accurate point correspondence, the computational cost is too high for practical use (several seconds on a GPU). As a tradeoff, Dosovitskiy \etal\cite{Dosovitskiy15} and Ilg \etal\cite{Ilg17} perform feature matching only at a reduced spatial resolution. On the contrary, we reduce the computational burden of feature matching by using a short-ranged matching of warped CNN features and a sub-pixel refinement at every pyramid level. We further reduce the computation cost by constructing sparse cost volumes at high levels. 

Jaderberg \etal propose a spatial transformer that allows spatial manipulation of feature maps within the network~\cite{Jaderberg15}. We use the spatial transformer for the f-warp. Specifically, given a high-dimensional feature map as the input, each feature vector\footnote{We can also use the f-warp layer to displace each channel differently when multiple flow fields are supplied. The usage, however, is beyond the scope of this work.} is individually displaced to a new location by the f-warp layer in accordance with the displacement vector at the corresponding position in the computed flow field. 
In comparison to FlowNet2~\cite{Ilg17} and SPyNet~\cite{Ranjan17}, the spatial transformation is limited to images, LiteFlowNet is a more generic warping network that warps high-level CNN features. 
Brabandere \etal propose a network to predict new frame(s) within a given video~\cite{Brabandere16}. The filters are generated dynamically conditioned on an input. We are inspired by flow regularization in variational methods~\cite{Horn81, Sun08, Werlberger09, Xu12, Zimmer11} and use the feature-driven convolution from Brabandere \etal in our framework to regularize flow fields.

%-------------------------------------------------------------------------
\section{LiteFlowNet}
\label{sec:liteflownet}
%-------------------------------------------------------------------------
Two lightweight sub-networks that are specialized in \textit{pyramidal feature extraction} and \textit{optical flow estimation} constitute LiteFlowNet. Figure~\ref{fig:network structure} shows an overview of its network architecture. Since the spatial dimension of feature maps is contracting in the feature encoder and that of flow fields is expanding in the flow decoder, we name the two sub-networks as NetC and NetE respectively. NetC transforms a given image pair respectively into two pyramids of multi-scale high-dimensional features. NetE consists of cascaded flow inference and regularization modules. It estimate flow fields from low to high spatial resolutions. 

\vspace{0.1cm}
\noindent \textbf{Pyramidal Feature Extraction.} As shown in Figure~\ref{fig:network structure}, NetC is a two-stream sub-network in which the filter weights are shared across the two streams. Each of them functions as a \textit{pyramidal feature descriptor} that transforms a given image $I$ to a pyramid of multi-scale high-dimensional features $\{\mathcal{F}_{k}(I)\}$ from the highest spatial resolution ($k = 1$) to the lowest spatial resolution ($k = L$). The pyramidal features are generated by stride-1 and stride-$s$ convolutions with the reduction of spatial resolution by a factor of $s$ down the inverted pyramid. In the following, we omit the subscript $k$ that indicates the level of pyramid for brevity. We use $\mathcal{F}_{i}$ to represent the extracted CNN features for $I_{i}$. When we discuss the operations in a pyramid level, the same operations are applicable to other levels. 

We use the design principle that high-resolution feature maps require a large receptive field for convolutional processing. For every decrement of two pyramid levels, we assign a smaller receptive field than the previous level. Suppose a 6-level feature encoder is used, the sizes of receptive field are set to 7, 7, 5, 5, 3, and 3 for levels 6 to 1, respectively. Since the size of receptive field across convolution layers can be accumulated, we improve the computational efficiency by replacing a large-kernel convolution layer with multiple small-kernel convolution layers. Except a $7\times7$ kernel is used at the first convolution layer in NetC, $3\times3$ kernels are used for the subsequent layers and the numbers of convolution layers are set to 3, 2, 2, 1, and 1 for levels 5 to 1, respectively. More details about the network architecture can be found in \nameref{sec:appendix}.

\vspace{0.1cm}
\noindent \textbf{Feature Warping.} We denote ${\bf x}$ as a point in the image domain $\Omega \subset \mathbb{R}^{2}$. At each pyramid level, a flow field ${\bf u}$, \ie a function ${\bf u}: \Omega \rightarrow \mathbb{R}^{2}$, is inferred from the features $\mathcal{F}_{1}$ and ${\mathcal F}_{2}$ of images $I_{1}$ and $I_{2}$. Flow inference becomes more challenging if $I_{1}$ and $I_{2}$ are captured far away from each other because a correspondence needs to be searched in a large area. With the motivation of \textit{image warping} used in conventional methods~\cite{Brox04, Papenberg06} and recent CNNs~\cite{Ilg17, Ranjan17} for addressing large-displacement flow, we propose to reduce the feature-space distance between $\mathcal{F}_{1}$ and ${\mathcal F}_{2}$ by \textit{feature warping} (f-warp) prior to recovering the flow field. Specifically, ${\mathcal F}_{2}$ is warped towards ${\mathcal F}_{1}$ by f-warp via a flow estimate ${\bf u}$, \ie $\widetilde {\mathcal F}_{2}({\bf x}) \triangleq {\mathcal F}_{2}({\bf x}+{\bf u}) \sim {\mathcal F}_{1}({\bf x})$. This allows our network to infer residual flow $\Delta {\bf u}$ between ${\mathcal F}_{1}$ and warped ${\mathcal F}_{2}$ (\ie $\widetilde {\mathcal F}_{2}$) that has smaller flow magnitude but not the complete flow field ${\bf u}$ that is more difficult to infer (more details in Section~\ref{sec:cascaded flow inference}).
Unlike conventional methods, f-warp is performed on high-level CNN features but not on images. This makes our network more powerful and efficient in addressing the optical flow problem. To allow end-to-end training, ${\mathcal F}$ is interpolated to ${\widetilde {\mathcal F}}$ for any sub-pixel displacement ${\bf u}$ as follows:
\begin{equation}\label{bi interpolation}
\widetilde{\mathcal F}({\bf x}) = \sum_{{\bf x}_{s}^{i} \in {\mathcal N}({\bf x}_{s})}{\mathcal F}({\bf x}_{s}^{i})\left(1-\left| x_{s} - x_{s}^{i}\right|\right) \left(1-\left| y_{s} - y_{s}^{i}\right|\right),
\end{equation}
where ${\bf x}_{s} = {\bf x}+{\bf u} = (x_{s}, y_{s})^{\top}$ denotes the source coordinates in the input feature map ${\mathcal F}$ that defines the sample point, ${\bf x} = (x, y)^{\top}$ denotes the target coordinates of the regular grid in the interpolated feature map $\widetilde{\mathcal F}$, and ${\mathcal N}({\bf x}_{s})$ denotes the four pixel neighbors of ${\bf x}_{s}$. The above bilinear interpolation allows back-propagation as its gradients can be efficiently computed~\cite{Jaderberg15}. 

\begin{figure}[t]
\centering
   \includegraphics[width=\columnwidth]{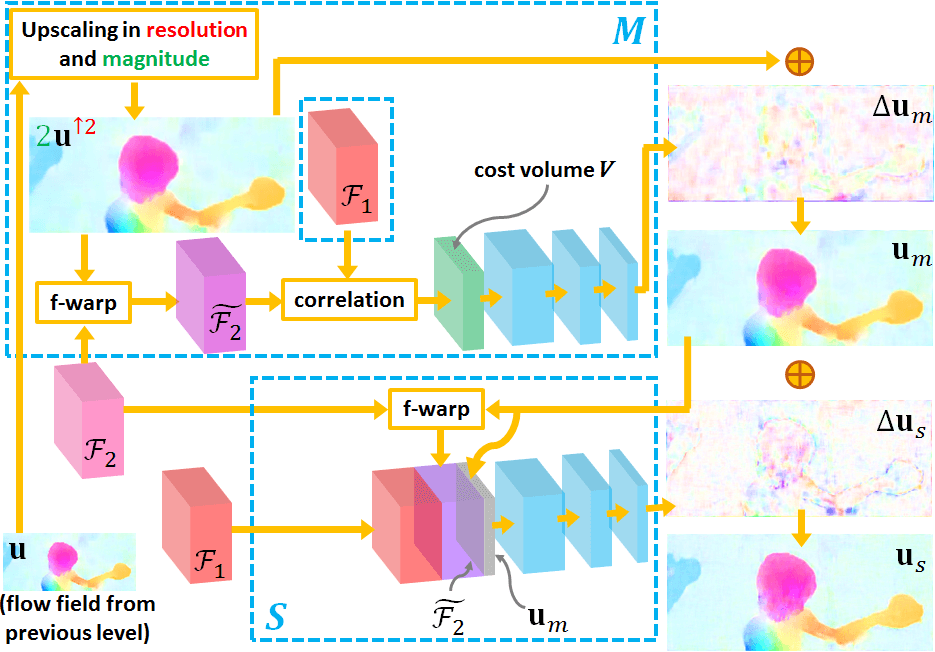}
\caption{A cascaded flow inference module $M$:$S$ in NetE. It consists of a descriptor matching unit $M$ and a sub-pixel refinement unit $S$. In $M$, f-warp transforms the high-level feature ${\mathcal F}_{2}$ to $\widetilde{\mathcal F}_{2}$ using the upscaled (by a factor of 2) flow estimate $2 {\bf u}^{ \uparrow 2}$ from the previous pyramid level. In $S$, $\mathcal F_{2}$ is warped by the flow estimate ${\bf u}_{m}$ resulting from $M$. Residual flow $\Delta {\bf u}_{m}$ is inferred from the cost volume $V$. $\Delta {\bf u}_{s}$ is used to correct ${\bf u}_{m}$ due to the pixel-level cost aggregation. In comparison to the residual flow $\Delta {\bf u}_{m}$, more flow adjustment can be found on flow boundaries in $\Delta {\bf u}_{s}$.}
\label{fig:MS}
\vspace{-0.5em}
\end{figure}

%-------------------------------------------------------------------------
\subsection{Cascaded Flow Inference}
\label{sec:cascaded flow inference}
%-------------------------------------------------------------------------
At each pyramid level of NetE, flow field inference is performed in a two-step procedure. An overview of the working mechanism is illustrated in Figure~\ref{fig:MS}. First, the pixel-by-pixel matching of high-level feature vectors across a given image pair yields a coarse flow estimate. Second, a subsequent refinement on the coarse flow further improves it to sub-pixel accuracy. The use of such a cascaded flow inference is novel in the literature.

\vspace{0.1cm}
\noindent \textbf{First Flow Inference -- Descriptor Matching.} Point correspondence between $I_{1}$ and $I_{2}$ is established through computing the correlation (\ie dot product) of high-level feature vectors in individual pyramidal features ${\mathcal F}_{1}$ and ${\mathcal F}_{2}$ as follows~\cite{Dosovitskiy15}:
\begin{equation}\label{eq:matching cost}
c({\bf x},{\bf d}) = {\mathcal F}_{1}({\bf x}) \cdot {\mathcal F}_{2}({\bf x}+{\bf d}) / N,
\end{equation}
where $c$ is the matching cost between point ${\bf x}$ in ${\mathcal F}_{1}$ and point ${\bf x}+{\bf d}$ in ${\mathcal F}_{2}$, ${\bf d} \in {\mathbb Z^{2}}$ (an 2-D integer set) is the displacement vector from ${\bf x}$, and $N$ is the length of the feature vector. The x- and y-components of ${\bf d}$ are bounded by $\pm D$ and $D \in {\mathbb Z}_{+}$ (an 1-D positive integer set). A cost volume $V$ is built by aggregating all the matching costs $c({\bf x},{\bf d})$ into a 3D grid. At pyramid level $k$, the dimension of $V$ is $\frac{H}{2^{k-1}} \times \frac{W}{2^{k-1}} \times (2D+1)$ for an image pair of size $H \times W$.

Unlike the conventional construction of cost volume~\cite{Dosovitskiy15, Ilg17}, we reduce the computational burden raised in three ways: 
\begin{enumerate}
\item \textit{Multi-Scale Short Searching Range}: Matching of feature vectors between ${\mathcal F}_{1}$ and ${\mathcal F}_{2}$ is performed within a short searching range at every pyramid level instead of using a long searching range only at a high-resolution pyramid level. 

\item \textit{Feature Warping}: We reduce the feature-space distance between ${\mathcal F}_{1}$ and ${\mathcal F}_{2}$ prior to constructing the cost volume. To this end, ${\mathcal F}_{2}$ is warped towards ${\mathcal F}_{1}$ by a f-warp layer using the flow estimate from the previous level. 

\item \textit{Sparse Cost Volume}: We perform feature matching only at the sampled positions in the pyramid levels with high spatial resolution. The sparse cost volume is interpolated in the spatial dimension to fill the missed matching costs for the unsampled positions. 
\end{enumerate}
The first two techniques effectively reduce the searching space needed, while the third technique reduces the frequency of matching per pyramid level. This in turn causes a speed-up in constructing the cost volume. 

In the descriptor matching unit $M$, the residual flow $\Delta{\bf u}_{m}$ between ${\mathcal F}_{1}$ and warped ${\mathcal F}_{2}$, \ie $\widetilde {\mathcal F}_{2}({\bf x}) = {\mathcal F}_{2}({\bf x} + s{\bf u}^{\uparrow s})$, is inferred from the constructed cost volume $V$ as illustrated in Figure~\ref{fig:MS}. A complete flow field ${\bf u}_{m}$ is computed as follows:
\begin{equation}
{\bf u}_{m} = \underbrace{M\big(V({\mathcal F}_{1}, \widetilde {\mathcal F}_{2}; D)\big)}_{\Delta {\bf u}_{m}} + s{\bf u}^{\uparrow s},
\end{equation}
where flow field ${\bf u}$ from a preceding level needs to be upsampled in spatial resolution (denoted by ``$\uparrow$$s$") and magnitude (multiplied by a scalar $s$) to $s{\bf u}^{\uparrow s}$ for matching the resolution of the pyramidal features in the current level. For consecutive levels, we use $s=2$.

\vspace{0.1cm}
\noindent 
\textbf{Second Flow Inference -- Sub-Pixel Refinement.} Since the cost volume in the descriptor matching unit is aggregated by measuring pixel-by-pixel correlation, flow estimate ${\bf u}_{m}$ resulting from the previous inference is only up to pixel-level accuracy. We introduce the second flow inference in the wake of descriptor matching as shown in Figure~\ref{fig:MS}. It aims to refine the pixel-level flow field ${\bf u}_{m}$ resulting from the descriptor matching unit to sub-pixel accuracy. This prevents erroneous flows being amplified by upsampling and passing to the next pyramid level. Specifically, ${\mathcal F}_{2}$ is warped to a new $\widetilde {\mathcal F}_{2}$ using the current flow estimate ${\bf u}_{m}$. For correcting ${\bf u}_{m}$, the sub-pixel refinement unit $S$ yields a more accurate flow field ${\bf u}_{s}$ by minimizing the feature-space distance between ${\mathcal F}_{1}$ and $\widetilde {\mathcal F}_{2}$ through computing a residual flow $\Delta {\bf u}_{s}$ as follows:
\begin{equation}
{\bf u}_{s} = \underbrace{S\big({\mathcal F}_{1}, \widetilde {\mathcal F}_{2}, {\bf u}_{m}\big)}_{\Delta {\bf u}_{s}} + {\bf u}_{m}.
\end{equation}

\begin{figure}[t]
\centering
   \includegraphics[width=\columnwidth]{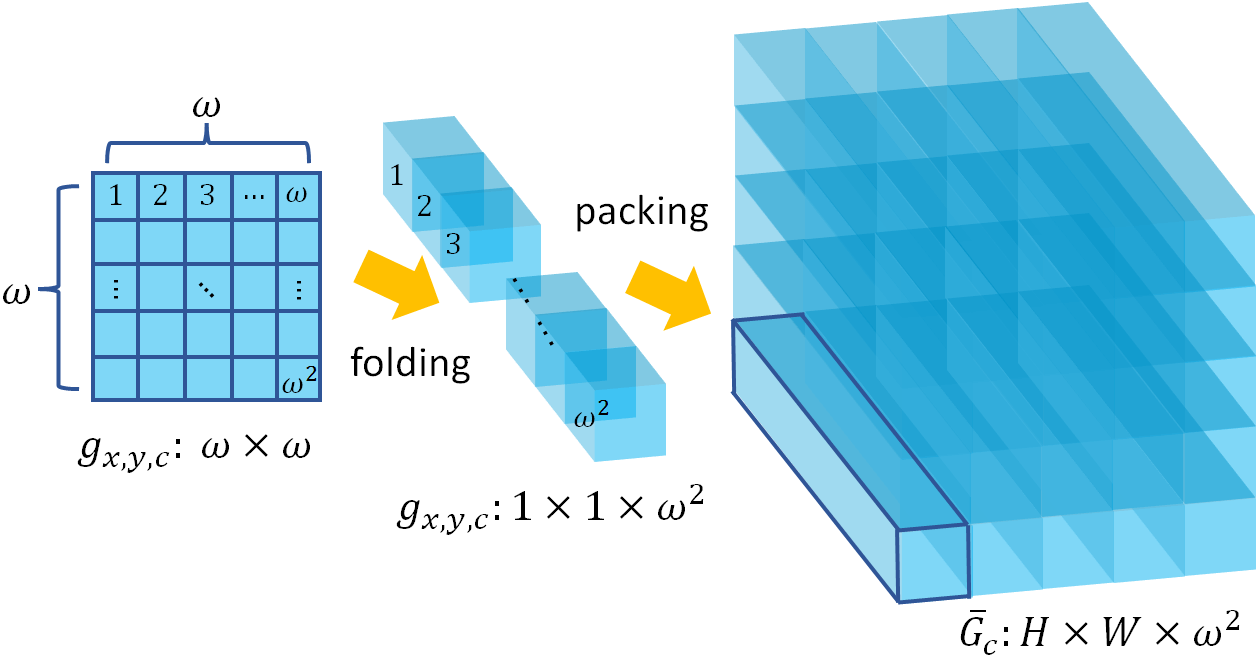}
\caption{Folding and packing of f-lconv filters $\{g\}$. The $(x,y)$-entry of a 3D tensor $\bar G_{c}$ (cube on the right) with size $H \times W \times \omega^{2}$ is a 3D column vector with length $w^{2}$. It corresponds to the unfolded f-lconv filter $g_{x,y,c}$ (plane on the right) with size $\omega \times \omega$ to be applied at position $(x,y)$ and channel $c$ in the vector-valued feature $\mathcal{F}$.}
\label{fig:fold}
\vspace{-0.5em}
\end{figure}

%-------------------------------------------------------------------------
\subsection{Flow Regularization}
\label{sec:flow regularization}
%-------------------------------------------------------------------------
Cascaded flow inference resembles the role of data fidelity in conventional minimization methods. However using data term alone, vague flow boundaries and other undesired artifacts can exist in flow fields~\cite{Werlberger09, Zimmer11}. To tackle this problem, \textit{feature-driven local convolution} (f-lconv) is used to regularize each flow field resulting from the cascaded flow inference. The operation of f-lconv to a flow field is well-governed by the Laplacian formulation of diffusion of pixel values~\cite{Tschumperle05} (see Section~\ref{sec:regularization term} for more details). In contrast to the \textit{local convolution} (lconv) used in conventional CNNs~\cite{Taigman14}, the f-lconv is more generalized. Not only a distinct filter is used for each position of a flow field but the filter is adaptively constructed to regularize each flow vector with a weighted average of flow vectors from nearby pixels.

Consider a general case, a vector-valued feature $\mathcal{F}$ that has to be regularized has a spatial dimension $H \times W$ and $C$ channels. Define $G = \{g\}$ as a set of filters used in a f-lconv layer. The operation of a f-lconv filter $g_{x,y,c}$ with size $\omega \times \omega$ to $\mathcal{F}$ at position $(x, y)$ and channel $c$ is formulated as follow:
\begin{equation}\label{eq:local convolution1}
  {\mathcal F}_{r}(x,y,c) = \sum_{(x_{i}, y_{i}) \in {\mathcal N}(x,y)}g_{x,y,c}(x_{i},y_{i}) \mathcal{F}(x + x_{i},y + y_{i},c),
\end{equation}
where ${\mathcal F}_{r}(x,y,c)$ is the scalar output and ${\mathcal N}(x,y)$ denotes the neighborhood containing $\omega \times \omega$ pixels centered at position $(x,y)$.

To regularize a flow field, f-lconv filters need to be specialized. It should behave as an averaging filter if the variation of flow vectors over a patch is supposed to be smooth. It should also not over-smooth flow vectors across flow boundary. To this end, we design a CNN unit $R_{D}$ to generate a feature-driven variation metric $\mathcal{D}$ with dimension $H \times W \times \omega \times \omega \times C$\footnote{For the case of flow field, the dimension of $\mathcal{D}$ is $H \times W \times \omega \times \omega \times 2$ as a flow field has 2 channels. But for the purpose of a lightweight implementation, both channels of a flow field is regularized equally, \ie $C=1$.}. It predicts the local flow variation over a patch with size $\omega \times \omega$ at all positions in a flow field using pyramidal feature $\mathcal{F}_{1}$, flow field ${\bf u}_{s}$ from the cascaded flow inference, and occlusion probability map\footnote{We use $L_{2}$ brightness error $||I_{2}({\bf x}+{\bf u})-I_{1}({\bf x})||_{2}$ between the warped second image and the first image as the occlusion probability map.} $O$ as follows:  
\begin{equation}\label{eq:distance metric D}
     \mathcal{D} = R_{\mathcal{D}}(\mathcal{F}_{1}, {\bf u}_{s}, O).
\end{equation}
With the introduction of feature-driven variation metric $\mathcal{D}$, each filter $g$ of f-lconv is constructed as follows:
\begin{equation}
     g_{x,y,c}(x_{i},y_{i}) = \frac{\text{exp}(-\mathcal{D}(x,y,x_{i},y_{i},c)^{2})}{\sum_{(x_{j},y_{j}) \in {\mathcal N}(x,y)} \text{exp}(-\mathcal{D}(x,y,x_{j},y_{j},c)^{2})}.
\end{equation} 
We intend to use the negative tail of the exponential function to constrain the values of f-lconv filters in $[0, 1]$ as the rapid-growing positive tail makes the training of the f-lconv more difficult.

Here, we provide a mechanism to perform f-lconv efficiently. For a $C$-channel input $\mathcal{F}$, we use $C$ tensors $\bar G_{1}, ..., \bar G_{C}$ to store f-lconv filter set $G$. As illustrated in Figure~\ref{fig:fold}, each f-lconv filter $g_{x,y,c}$ is folded into a 3D column vector with length $w^{2}$ and then packed into the $(x,y)$-entry of a 3D tensor $\bar G_{c}$ with size $H \times W \times w^{2}$. The same folding and packing operations are also applied to each patch in each channel of $\mathcal{F}$. This results in $C$ tensors $\bar F_{1}, ..., \bar F_{C}$ for $\mathcal{F}$. In this way, Equation~\eqref{eq:local convolution1} is reformulated to:
\begin{equation}\label{eq:local convolution2}
   {\mathcal F}_{r}(c) = \bar G_{c} \odot \bar F_{c},
\end{equation}
where ``$\odot$" denotes element-wise dot product between the corresponding column vectors of the tensors. With the abuse of notation, ${\mathcal F}_{r}(c)$ denotes the $xy$-slice at channel $c$ in the regularized $C$-channel feature $\mathcal{F}_{r}$. The operation of element-wise dot product reduces the dimension of tensors on the right-hand side from $H \times W \times \omega^{2}$ to $H \times W$.

To summarize, ${\bf u}_{s}$ resulting from the cascaded flow inference is adaptively regularized by the flow regularization module $R$ using a set of f-lconv filters $G$ as follows:
\begin{equation}
     {\bf u}_{r} = R({\bf u}_{s}; G).
\end{equation}

%-------------------------------------------------------------------------
\section{Correspondences between Optical Flow CNNs and Variational Methods}
%-------------------------------------------------------------------------
We first provide a brief review for estimating optical flow using variational methods. In the next two sub-sections, we will bridge the correspondences between optical flow CNNs and classical variational methods.

Consider an image sequence $I({\bf x}, t): {\mathbb{R}}^{3} \rightarrow \mathbb{R}$ with ${\bf x} = (x, y)^{\top} \in \Omega$ over a rectangular spatial domain $\Omega \subset \mathbb{R}^{2}$ and a temporal dimension $t$. The optical flow field ${\bf u}: \Omega \rightarrow {\mathbb{R}}^{2}$ that is induced by the spatial motion of the scene and/or the camera itself corresponds to the displacement vector field between images $I_{1}$ (at $t = 1$) and $I_{2}$ (at $t = 2$). The flow field can be estimated by minimizing an energy functional $E$ of the general form~\cite{Zimmer11}:
\begin{equation}\label{eq:functional}
\begin{split}
	E({\bf u}) &=  E_{dat}({\bf u}) + \lambda E_{reg}(\nabla{\bf u}) \\
                             &= \int_{\Omega} \big(e_{data}({\bf u}) +  \lambda e_{reg}(\nabla{\bf u})\big) d{\bf x},
\end{split}
\end{equation}
where $e_{dat}$ and $e_{reg}$ represent the data and regularization costs respectively, and $\lambda > 0$ is the smoothness weight. 

%-------------------------------------------------------------------------
\subsection{Data Term}
%-------------------------------------------------------------------------
Point correspondence across a pair of images is imposed in the data term of Eq.~\eqref{eq:functional} as a combination of several matching quantities $\{D_{i}\}$ as follows~\cite{Zimmer11, Kim13}:
\begin{equation}\label{eq:data term}
	E_{dat}({\bf u}) = \int_{\Omega} \sum \gamma_{i} D_{i} (I_{1}, I_{2}) d{\bf x},
\end{equation}
where $\gamma_{i}$ is the weighting factor for $D_{i}$. Two popular matching quantities are image brightness constancy assumption $\Psi\big(\left| I_{2}({\bf x} + {\bf u} ) - I_{1}({\bf x}) \right|^{2} \big)$~\cite{Horn81} and gradient constancy assumption $\Psi\big(\left| \nabla I_{2}({\bf x} + {\bf u}) - \nabla I_{1}({\bf x}) \right|^{2} \big)$~\cite{Brox04}, where $\Psi$ is a robust penalty function. Other higher-order constancy data terms are also widely used~\cite{Papenberg06}. 
The contributions of different matching quantities need to be compromised by using appropriate weighting factors~\cite{Xu12, Kim13}. It is also necessary to maintain differentiability of both data and regularization (Section~\ref{sec:regularization term}) terms because Eq.~\eqref{eq:functional} needs to be solved using the Euler-Lagrange equation. 

In comparison to conventional methods, state-of-the-art optical flow networks do not explicitly define those matching quantities $\{D_{i}\}$. Back2Basics~\cite{Yu16} uses a photometric loss that is computed as the difference between the first image and the warped second image. SPyNet~\cite{Ranjan17} uses a pair of images from the image pyramids to generate a flow field at the corresponding pyramid level. PWC-Net~\cite{Sun18} and  LiteFlowNet~\cite{Hui18} use a learnable feature encoder instead. In more details, we train NetC of LiteFlowNet as a CNN-based \textit{pyramidal feature descriptor} $\mathcal{F}(I): {\mathbb{R}}^{2}\rightarrow {\mathbb{R}}^{N}$ that transforms a given image pair $(I_{1}, I_{2})$ respectively into two pyramids of multi-scale high-dimensional features. With the introduction of feature descriptor, the \textit{cascaded flow inference} in NetE that has been presented in Section~\ref{sec:cascaded flow inference} is trained to solve for the minimization of the difference between the high-level features $\mathcal{F}_{2}$ of $I_{2}$ and $\mathcal{F}_{1}$ of $I_{1}$ by computing the dense correspondence between them. In other words, feature encoders that are used in LiteFlowNet and other optical flow CNNs~\cite{Ilg17,Ranjan17,Sun18} resemble the role of data term in variational methods. 

%-------------------------------------------------------------------------
\subsection{Regularization Term}
\label{sec:regularization term}
%-------------------------------------------------------------------------
Flow field that is merely computed by data fidelity is fragile to outliers. Energy functional is often augmented to enforce dependency between neighboring flow vectors~\cite{Horn81}. Regularization of a vector field can be viewed as diffusion of pixel values~\cite{Tschumperle05}. By applying the Euler-Lagrange equation to Eq.~\eqref{eq:functional}, the regularization component is given by: 
\begin{equation}\label{eq:divergence form}
\text{div}\left(\partial_{\nabla{\bf u}} E_{reg}(\nabla{\bf u}) \right) = \text{div}({\bf D} \nabla{\bf u}), 
\end{equation}
where ${\bf D}$ is a $2 \times 2$ diffusion tensor. 
The above divergence formulation can also be rewritten into an oriented Laplacian form as follows:
\begin{equation}\label{eq:Laplacian form}
\text{div}({\bf D} \nabla{\bf u}) = \text{trace}({\bf T}{\bf H}_{i}), i = 1, 2,
\end{equation}
where ${\bf H}_{i}$ is the Hessian matrix of the $i$-th vector component of the flow field and ${\bf T}$ is a $2 \times 2$ tensor. The solution of Eq.~\eqref{eq:Laplacian form} is given by:
\begin{equation}\label{eq:flow smoothing}
	{\bf u} = K(\textbf{T}) \ast {\bf u}',
\end{equation}
where ``$\ast$" denotes a convolution and $K$ is a 2D oriented Gaussian kernel (the exact structure of $K$ depends on ${\bf D}$ used in $E_{reg}$) and ${\bf u}'$ is the intermediate flow field generated from the data term~\cite{Xiao06}. In other words, enforcing smoothness constraint on the flow field is equivalent to applying a convolution with a 2D oriented Gaussian kernel to the intermediate flow field generated by the data term. 

Unlike the smoothing kernel in Eq.~\eqref{eq:flow smoothing} that requires engineered regularizing structure, we use a \textit{feature-driven local convolution} (f-lconv) filters $G= \{g\}$ to regularize each flow vector differently in the flow field by adapting f-lconv kernel to the pyramidal feature $\mathcal{F}$ resulting from the encoder, intermediate flow field ${\bf u}'$ from the data term, and occlusion probability map~$O$. Our feature-driven flow regularization is defined as follows:
\begin{equation}\label{eq:feature-driven flow smoothing}
	{\bf u} = g\left(\mathcal{F}_{1}, {\bf u}', O\right) \ast {\bf u}'.
\end{equation}
The flow regularization module $R$ in NetE that performs the above feature-driven flow smoothing operation has been presented in Section~\ref{sec:flow regularization}. By replacing the intermediate flow field ${\bf u}'$ to flow field ${\bf u}_{s}$ generated from the cascaded flow inference, Eq.~\eqref{eq:feature-driven flow smoothing} corresponds to Eq.~\eqref{eq:local convolution1}. This concludes that our feature-driven regularization resembles the role of regularization term in variational methods. In Back2Basics~\cite{Yu16}, flow regularization is enforced by a piecewise smoothness function in the training loss instead.  

\begin{figure}[t]
\centering
  \includegraphics[width=8.0cm]{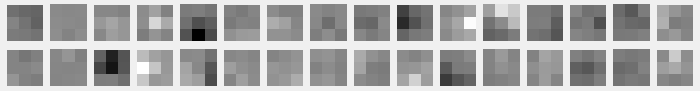} \\ 
  \vspace{0.1cm}
  \includegraphics[width=8.0cm]{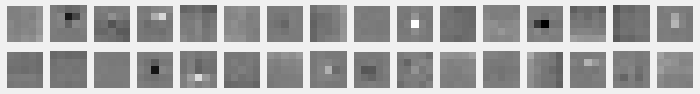} 
\caption{A visualization of the learned filters with sizes $3\times3$ and $5\times5$ for the horizontal flow component at level 6 (top 2 rows) and level 3 (bottom 2 rows) in the sub-pixel refinement unit of LiteFlowNet2, respectively.}
\label{fig:flow bases}
\vspace{-0.5em}
\end{figure}

%-------------------------------------------------------------------------
\section{Relationship between Optical Flow CNNs and Basis Representation}
%-------------------------------------------------------------------------
The parameterized models of image motion~\cite{Black97,Nir08,Wulff15} use a linear combination of basis vectors $\{{\bf m}_{i} \in {\mathbb{R}}^{2hw}\}$ to approximate an image motion ${\bf u}$ within an image patch with size $h \times w$ as follows:
\begin{equation} \label{eq:flow basis equation}
  {\bf u}_{vec} = \textstyle \sum_{i = 1}^{C}a_{i}{\bf m}_{i},  
\end{equation} 
where ${\bf u}_{vec} \in {\mathbb{R}}^{2hw}$ is the vectorized flow field of ${\bf u}$ by packing all the $x$- and $y$-components of ${\bf u}$ into a single vector and $\{a_{i}\}_{i=1, 2, ..., C}$ are the flow coefficients to be estimated. 

The above basis representation is related to flow inferences in LiteFlowNet~\cite{Hui18} and other optical flow CNNs~\cite{Dosovitskiy15,Ilg17,Ranjan17,Sun18}. At a pyramid level, the penultimate and last layers of the descriptor matching and sub-pixel flow refinement units in LiteFlowNet can be represented by the following:
\begin{subequations}\label{eq:S equations}
\renewcommand{\theequation}{\theparentequation.\arabic{equation}}
\begin{align}
     {\mathcal F}^{N-1} &= \sigma\left({\bf W}^{N-1}\ast {\mathcal F}^{N-2} + {\bf b}^{N-1} \right), \label{eq:S equation 1}\\
     \Delta {\bf u} &= {\bf W}^{N} \ast {\mathcal F}^{N-1} + {\bf b}^{N}, \label{eq:S equation 2}
\end{align}
\end{subequations} 
where ``$\ast$'' denotes a convolutional operator and $N$ is the total number of convolution layers used. Furthermore, ${\bf W}^{i}$ and ${\mathcal F}^{i}$ represent the convolution filters and feature maps that are used and generated at the $i$-th layer, respectively. A trainable bias $b^{i}$ is added to each feature map after the convolutional operation. We denote a set of bias scalars as ${\bf b}^{i}$. Each convolution layer is followed by an activation function $(\sigma)$ for non-linear mapping unless otherwise specified.
Suppose ${\mathcal F}^{N-1}$ in Eq.~\eqref{eq:S equation 2} is a $C$-channel vector-valued feature map, the equation can be re-written into the expanded form as follows:
\begin{equation}
   \Delta {\bf u} = \textstyle \sum_{i = 1}^{C}\big({\bf W}_{i}\ast {\mathcal F}_{i} + b_{i}\big), \label{eq:expanded S equation}
\end{equation}
where ${\bf W} = \{{\bf W}_{i}\}$, ${\bf b} = \{b_{i}\}$, and ${\mathcal F}_{i}$ is the $i$-th channel of ${\mathcal F}$  (superscripts $N$ and $N-1$ are removed for brevity). 

\vspace{0.1cm}
\noindent 
\textbf{Similarities.} Suppose the residual flow $\Delta {\bf u}$ in Eq.~\eqref{eq:expanded S equation} is the flow field that we need to estimate even though it is not the full flow, the vectorized ${\bf W}_{i}\ast {\mathcal F}_{i} + b_{i}$ resembles $a_{i}{\bf m}_{i}$ in Eq.~\eqref{eq:flow basis equation}. The number of channels of ${\mathcal F}^{N-1}$ in Eq.~\eqref{eq:S equation 2} corresponds to the number of basis vectors in Eq.~\eqref{eq:flow basis equation}.
In particular, the filters ${\bf W}$ and feature maps ${\mathcal F}$ in Eq.~\eqref{eq:S equation 2} correspond to the basis vectors $\{{\bf m}_{i}\}$ and flow coefficients $\{a_{i}\}$ in Eq.~\eqref{eq:flow basis equation}, respectively. The computation of feature maps (\ie flow coefficients in conventional basis representation) for CNN flow inference is governed by the $N-1$ convolution layers prior to Eq.~\eqref{eq:S equation 2}. Figure~\ref{fig:flow bases} provides an example of the visualization of the learned filters (\ie flow bases in conventional basis representation) at level 6 and level 3 in the sub-pixel refinement unit of LiteFlowNet2. 

\vspace{0.1cm}
\noindent 
\textbf{Differences.} The dimension of CNN filters is usually small (a few pixels width) while the dimension of basis fields (before vectorization) is same as image patches under consideration. The dimension of CNN feature maps is proportional to that of the given images (depending on the pyramid level under consideration) while flow coefficients are scalars. Furthermore, a flow vector is constructed by a convolution between CNN filters and feature patches centered at the corresponding position in the feature maps as the flow vector while each vectorized flow patch is a linear combination of basis vectors.

%-------------------------------------------------------------------------
\section{Experiments}
\label{sec:experiments}
%-------------------------------------------------------------------------
%-------------------------------------------------------------------------
\subsection{LiteFlowNet} 
\label{sec:exp liteflownet}
%-------------------------------------------------------------------------
\noindent
\textbf{Network Details.} In LiteFlowNet, NetC is a 6-level feature encoder and NetE is a flow decoder for generating flow fields from levels 6 to 2 in a coarse-to-fine manner. Flow field at level 2 is upsampled by a bilinear interpolation to the same resolution at level~1 as the given image pair. 
We set the maximum searching radius for constructing cost volumes to 3 and 6 pixels for levels 6 to 4 and levels 3 to 2, respectively. Matching is performed at every position across two pyramidal features to form a cost volume, except for levels 3 to 2 that it is performed at a regularly sampled grid (using a stride of 2) to form a sparse cost volume. 
All convolution layers use $3\times3$ filters, except the first layer in NetC uses $7\times7$ filters, each last layer in descriptor matching $M$, sub-pixel refinement $S$, and flow regularization $R$ uses $5\times5$ filters for levels 4 to 3 and $7\times7$ filters for level 2. 
Each convolution layer is followed by a leaky rectified linear unit layer, except f-lconv and the last layers in $M$, $S$ and $R$ networks.
More network details can be found in \nameref{sec:appendix}.

\vspace{0.1cm}
\noindent
\textbf{Training Details.} In conventional training methods~\cite{Ilg17,Sun18}, all parts of network are trained by the same number of iterations. On the contrary, we pre-train LiteFlowNet on FlyingChairs dataset~\cite{Dosovitskiy15} using \textbf{stage-wise training protocol} as follows: First, NetC and $M_{6}$:$S_{6}$ of NetE are trained for 300K iterations. Second, $R_{6}$ together with the trained network in step 1 are trained for 300K iterations. Third, for levels $k \in [5, 2]$, $M_{k}$:$S_{k}$ followed by $R_{k}$ is added into the trained network each time. The new network cascade is trained for 240K iterations, except the last-level network is trained for 300K iterations. The new filter weights at level $k$ are initialized from the previous level $k-1$. The advantages of stage-wise training over the conventional training are:
\begin{enumerate}
\item \textit{Shorter training time}: Since the network stages are gradually added in the cascade, the stages that are lately added are relatively trained by a smaller number of iterations than the early added stages. Furthermore, the runtime of the cascade consisting lesser network stages is faster than the more complete network. The overall network, therefore, requires lesser training time than the conventional training method. The training of LiteFlowNet2 (including training and validation phases) requires 5.5 days instead of 8 days on an NVIDIA TITAN X. 

\item \textit{Better performance}: Although the network stages that are lately added are relatively trained by a smaller number of iterations, stage-wise training promotes lower training losses on the overall network. This is possible because filter weights in the succeeding stage are well-initialized from the previously trained stage rather than randomly assigned. The average end-point error (AEE) of LiteFlowNet2 is improved from 4.66 to 4.11 on KITTI 2012~\cite{Geiger12} and is significantly improved from 12.42 to 11.31 on KITTI 2015~\cite{Menze15}. For benchmarking on FlyingChairs, AEEs are 1.68 (vs 1.70) and 1.60 (vs 1.61) on the training and validation sets, respectively. The results are significantly improved on KITTI but are similar on FlyingChairs. This indicates that stage-wise training is effective in alleviating the over-fitting issue as well.
\end{enumerate}

Learning rates are initially set to 1e-4, 5e-5, and 4e-5 for levels 6 to 4, 3, and 2 respectively. We reduce it by a factor of 2 starting at 120K, 160K, 200K, and 240K iterations. We use the same batch size of 8, data set resolution (randomly cropped: $448\times320$), loss weights (levels 6 to 2: 0.32, 0.08, 0.02, 0.01, 0.005), training loss ($L_{2}$ flow error), Adam optimization ($\gamma = 0.5$, weight decay = 4e-4), data augmentation (including noise injection), scaled ground-truth flow (by a factor of $\frac{1}{20}$) as FlowNet2~\cite{Ilg17}. Furthermore, we use a training loss for every inferred flow field. 

After pre-training LiteFlowNet on FlyingChairs (Chairs)~\cite{Dosovitskiy15}, it is trained on a more challenging data set,  Things3D\footnote{We excluded a small amount of training data in Things3D undergoing extremely large flow displacement as advised by the authors~(\url{https://github.com/lmb-freiburg/flownet2/issues}).}~\cite{Mayer16} according to the training schedule (Chairs~$\rightarrow$~Things3D) as FlowNet2~\cite{Ilg17}. It is trained for 500K iterations. Batch size is reduced to 4 and dataset resolution is increased to $768\times384$. Learning rate is set to 3e-6 and is reduced by half starting at 200K iterations for every increment of 100K iterations. No stage-wise training is used for subsequent fine-tunings. We denote \textbf{LiteFlowNet-pre} and \textbf{LiteFlowNet} as the networks pre-trained on Chairs and fine-tuned on Things3D, respectively. 

After training on Things3D, we use the generalized Charbonnier function $\rho(x) = (x^{2} + \epsilon^{2})^{q}$ ($\epsilon^{2} = 0.01$ and $q = 0.2$) as the robust training loss for further fine-tuning on subsequent datasets. The flow accuracy of LiteFlowNet2 is improved (KITTI 2012~\cite{Geiger12}: 3.73 vs 3.42 and KITTI 2015~\cite{Menze15}: 9.80 vs 8.97) when the robust loss with a higher learning rate 1e-5 is used for fine-tuning on Things3D. However, the results on the testing sets of Sintel~\cite{Butler12} and KITTI are not much different from the case without using the robust loss. Therefore, we choose to use $L_{2}$ loss when fine-tuning on Things3D. The fine-tuning details on the respective training sets of Sintel and KITTI will be presented in Section~\ref{sec:results}.

%-------------------------------------------------------------------------
\subsection{LiteFlowNet2} 
\label{sec:liteflownet2}
%-------------------------------------------------------------------------
\begin{table}[t]
\small
\centering
\caption{AEE and runtime (for Sintel) measured at different components (NetC: a feature encoder, NetE: a multi-scale flow decoder) and pyramid levels of LiteFlowNet~\cite{Hui18} trained on Things3D. The percentage change is relative to the previous level.} \label{tab:AEE and runtime analysis}
\scalebox{0.85}{
\begin{tabular}{ccccccc}
\hline

\multicolumn{1}{|c|}{}  				&\multicolumn{1}{c|}{NetC}         
											&\multicolumn{5}{c|}{NetE}  \\ 
\hline
\multicolumn{1}{|l|}{Level} 		     &\multicolumn{1}{c|}{-}
											&\multicolumn{1}{c|}{6}
											&\multicolumn{1}{c|}{5}
											&\multicolumn{1}{c|}{4}
											&\multicolumn{1}{c|}{3}
											&\multicolumn{1}{c|}{2} \\
\hline
\multicolumn{1}{|l|}{Sintel Clean} &\multicolumn{1}{c|}{-}
											   &\multicolumn{1}{c|}{~~~5.41~~~}
											   &\multicolumn{1}{c|}{3.85}
											   &\multicolumn{1}{c|}{3.03}
											   &\multicolumn{1}{c|}{2.65}
											   &\multicolumn{1}{c|}{2.48} \\	
\multicolumn{1}{|l|}{} &\multicolumn{1}{c|}{-}
											   &\multicolumn{1}{c|}{-}
											   &\multicolumn{1}{c|}{-28.8\%}
											   &\multicolumn{1}{c|}{-21.3\%}
											   &\multicolumn{1}{c|}{-12.5\%}
											   &\multicolumn{1}{c|}{-6.4\%} \\
\hline	
\multicolumn{1}{|l|}{KITTI 2012} &\multicolumn{1}{c|}{-}
											   &\multicolumn{1}{c|}{~~~8.58~~~}
											   &\multicolumn{1}{c|}{6.04}
											   &\multicolumn{1}{c|}{4.67}
											   &\multicolumn{1}{c|}{4.18}
											   &\multicolumn{1}{c|}{4.00} \\	
\multicolumn{1}{|l|}{} &\multicolumn{1}{c|}{-}
											   &\multicolumn{1}{c|}{-}
											   &\multicolumn{1}{c|}{-29.6\%}
											   &\multicolumn{1}{c|}{-22.7\%}
											   &\multicolumn{1}{c|}{-10.5\%}
											   &\multicolumn{1}{c|}{-4.4\%} \\
\hline				   
\multicolumn{1}{|l|}{Runtime (ms)}  &\multicolumn{1}{c|}{14.36}
											&\multicolumn{1}{c|}{1.69}
											&\multicolumn{1}{c|}{2.03}
											&\multicolumn{1}{c|}{4.88}
											&\multicolumn{1}{c|}{13.06}
											&\multicolumn{1}{c|}{52.51} \\	
\multicolumn{1}{|l|}{} &\multicolumn{1}{c|}{-}
											   &\multicolumn{1}{c|}{-}
											   &\multicolumn{1}{c|}{+20.1\%}
											   &\multicolumn{1}{c|}{+140\%}
											   &\multicolumn{1}{c|}{+168\%}
											   &\multicolumn{1}{c|}{+302\%} \\										
\hline											
\end{tabular}}
\vspace{-0.1cm}
\end{table}

\begin{figure*}[t]
\centering
\captionsetup[subfigure]{labelformat=empty, justification=centering}
\captionsetup[subfloat]{farskip=0pt,captionskip=0pt}
\begin{tabular}{ccccc}
	 \subfloat[~~~~Level 6: $32\times14$, \newline AEE: 2.28]{\includegraphics[width=3.55cm]{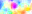}}\hfill
	\subfloat[~~~~Level 5: $64\times28$, \newline AEE: 1.22]{\includegraphics[width=3.55cm]{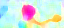}}\hfill
	\subfloat[~~~~Level 4: $128\times56$, \newline AEE: 0.76]{\includegraphics[width=3.55cm]{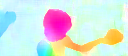}}\hfill
	\subfloat[~~~~Level 3: $256\times112$, \newline AEE: 0.47]{\includegraphics[width=3.55cm]{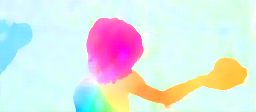}}\hfill
	\subfloat[~~~~Level 2: $512\times224$, \newline AEE: 0.33]{\includegraphics[width=3.55cm]{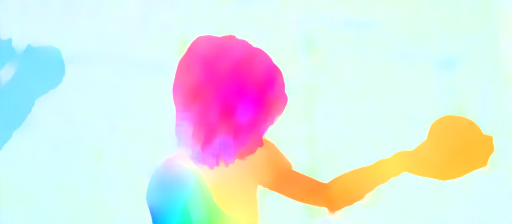}}\hfil
\end{tabular}
\vspace{-0.2cm}
\caption{An example of coarse-to-fine flow fields generated from LiteFlowNet~\cite{Hui18} trained on Chairs~$\rightarrow$~Things3D. Each of the flow fields is upsampled to the same resolution as the ground truth by bilinear interpolation prior to computing AEE.}
\label{fig:a pyramid of flows}
\vspace{-0.5em}
\end{figure*}

We analyze the flow accuracy in terms of AEE and the computation time at each pyramid level of LiteFlowNet~\cite{Hui18} trained on Chairs~$\rightarrow$~Things3D. The results are summarized in Table~\ref{tab:AEE and runtime analysis}. An example of multi-scale flow fields on Sintel Clean training set is also provided in Figure~\ref{fig:a pyramid of flows}. We optimize the network architecture of LiteFlowNet with the following motivations and evolve our earlier model to a faster and more accurate model \-- LiteFlowNet2. 

\vspace{0.1cm}
\noindent
\textbf{Pyramid Level.} As summarized in Table~\ref{tab:AEE and runtime analysis}, the computation time increases exponentially with the resolution of flow field. In particular, the improvement in flow accuracy is not significant when comparing the AEE at level 3 to that at level~2. On the contrary, about 60\% of the total computation time spent on the flow decoder at level~2. In LiteFlowNet2, we improve the computational efficiency by reducing the number of pyramid levels in NetE from five (levels 6 to 2) to four (levels 6 to 3).

\vspace{0.1cm}
\noindent
\textbf{Network Depth.} By limiting the pyramid level of NetE up to level 3 in LiteFlowNet~\cite{Hui18}, the flow accuracy is decreased as revealed in Table~\ref{tab:AEE and runtime analysis}. In order to compensate the loss, we add two convolution layers (with 128 and 96 output channels) between the 128- and 64-channel convolution layers to each flow decoder in the cascaded flow inference of NetE. We will show in Section~\ref{sec:results} that LiteFlowNet2 has a higher flow accuracy than LiteFlowNet.

\vspace{0.1cm}
\noindent
\textbf{Pseudo Flow Inference and Regularization.} 
We also address the inefficient computation at level 2 by introducing a simplified flow inference (without descriptor matching) and regularization at this level for the model fine-tuned on KITTI.
The pseudo network is constructed as follows: First, we remove all the layers before the last layer in the original flow inference and regularization, respectively. Then, we replace the removed layers by feed-forwarding the upsampled features respectively from the layer prior to the last layer in the flow inference and regularization at level~3. 
Using the pseudo network, the runtime at level 2 is greatly reduced from 52.51ms to 8.95ms. We have experienced that the pseudo network can improve the flow accuracy on KITTI testing set (evaluation will be provided in Section~\ref{sec:results}) but there is no significant improvement on Sintel testing set. Since the latter is a synthetic dataset, a flow CNN is more easily to be trained for fitting the non-realistic scene. However, the variability in real-world data such as lighting and object textures is more challenging. Therefore, using one more flow inference and regularization is beneficial to the refinement of preceding flow estimate on KITTI.

\begin{table}[t]
\small
\centering
\caption{AEE of LiteFlowNet2 trained on Chairs using different training protocols against LiteFlowNet~\cite{Hui18}.} \label{tab:results for different training protocols}
\scalebox{0.85}{
\begin{tabular}{l|c|c|c|c|c}
\hline          
\multicolumn{1}{|c|}{} 												
&\multicolumn{1}{c|}{Sintel Clean} 
&\multicolumn{1}{c|}{Sintel Final}  
&\multicolumn{1}{c|}{KITTI12}  
&\multicolumn{1}{c|}{KITTI15} \\
\hline 
\multicolumn{1}{|l|}{LiteFlowNet~\cite{Hui18}} 												&\multicolumn{1}{c|}{} 
&\multicolumn{1}{c|}{}   
&\multicolumn{1}{c|}{}   
&\multicolumn{1}{c|}{} \\
\multicolumn{1}{|l|}{~learning rate: 5e-5} %L3												
&\multicolumn{1}{c|}{2.94} 
&\multicolumn{1}{c|}{4.28}   
&\multicolumn{1}{c|}{4.73}   
&\multicolumn{1}{c|}{11.75} \\
\hline 
\multicolumn{1}{|l|}{LiteFlowNet2} 												
&\multicolumn{1}{c|}{} 
&\multicolumn{1}{c|}{}   
&\multicolumn{1}{c|}{}   
&\multicolumn{1}{c|}{} \\
\multicolumn{1}{|l|}{~learning rate: 5e-5} %L3												
&\multicolumn{1}{c|}{2.84} 
&\multicolumn{1}{c|}{4.16}   
&\multicolumn{1}{c|}{4.29}   
&\multicolumn{1}{c|}{12.01} \\
\multicolumn{1}{|l|}{~learning rate: 6e-5} %L3												
&\multicolumn{1}{c|}{2.80} 
&\multicolumn{1}{c|}{\textbf{4.14}}   
&\multicolumn{1}{c|}{4.25}   
&\multicolumn{1}{c|}{11.76} \\
\multicolumn{1}{|l|}{~+ extra training loss}											
&\multicolumn{1}{c|}{\textbf{2.78}} 
&\multicolumn{1}{c|}{\textbf{4.14}}   
&\multicolumn{1}{c|}{\textbf{4.11}}   
&\multicolumn{1}{c|}{\textbf{11.31}} \\
\hline 
\end{tabular}}
\vspace{-0.5em}
\end{table}

\vspace{0.1cm}
\noindent
\textbf{Training Details.}
We pre-train LiteFlowNet2 (\textbf{LiteFlowNet2-pre}) using the same stage-wise training protocol as LiteFlowNet~\cite{Hui18} except for a few minor differences. Learning rate is set to 6e-5 instead of 5e-5. At the output of last flow regularization in NetE, the flow field is further upsampled to the same resolution as the image pair and we introduce an additional training loss with a loss weight 6.25e-4. Table~\ref{tab:results for different training protocols} summarizes the results of LiteFlowNet and LiteFlowNet2 under different training protocols. Using the same training protocol as LiteFlowNet, LiteFlowNet2 outperforms LiteFlowNet on Sintel and KITTI 2012. If the learning rate of LiteFlowNet2 is increased to 6e-5, it is on par with LiteFlowNet on KITTI 2015. Using the learning rate of 6e-5 and extra training loss, LiteFlowNet2 outperforms LiteFlowNet on KITTI 2012 and 2015. 
The fine-tuning protocol for the respective training set of Sintel and KITTI is the same as LiteFlowNet unless otherwise specified. The improvements due to the better fine-tuning protocol will be presented in Section~\ref{sec:results}.

\begin{table}[t]
\small
\centering
\caption{AEE on the Chairs testing set. Models are trained on the Chairs training set.} \label{tab:flyingchairs results}
\scalebox{0.85}{
\begin{tabular}{c|c|c|c|c|c}
\hline          
\multicolumn{1}{|c|}{FlowNetS} 							
&\multicolumn{1}{c|}{FlowNetC}						
&\multicolumn{1}{c|}{SPyNet} 
&\multicolumn{1}{c|}{LiteFlowNetX-pre}   
&\multicolumn{1}{c|}{LiteFlowNet-pre} \\
\hline 
\multicolumn{1}{|c|}{2.71} 							
&\multicolumn{1}{c|}{2.19}						
&\multicolumn{1}{c|}{2.63} 
&\multicolumn{1}{c|}{2.25}   
&\multicolumn{1}{c|}{\textbf{1.57}} \\
\hline
\end{tabular}}
\vspace{-0.5em}
\end{table}
\begin{table*}[t]
\small
\centering
\caption{A comparison on the performance of the state-of-the-art optical flow methods in terns of AEE. The values in parentheses are the results of the networks on the data they were trained on, and hence are not directly comparable to the others. Out-Noc: Percentage of erroneous pixels defined as end-point error (EPE) $>$3 pixels in non-occluded areas. Fl-all: Percentage of outliers averaged over all pixels. Inliers are defined as EPE $<$3 pixels or $<$5\%. The best number for each category is highlighted in bold and the second best is underlined. (Notes: $^{1}$The values are reported from~\cite{Ilg17}. $^{2,3,4}$The values are computed using the trained models provided by the authors. $^{3}$Large discrepancy exists as the authors mistakenly evaluated the results on the disparity dataset. $^{4}$Up-to-date dataset is used. $^{6}$Trained on Driving and Monkaa~\cite{Mayer16}. $^{7}$Results are reported from the arXiv paper~\cite{Hui18-arxiv}.)} \label{tab:results}

\scalebox{0.85}{
\begin{tabular}{|c|l|c c|c c|c c c|c c c|}
\hline

\multirow{1}{*}{} 
&\multirow{1}{*}{Method}   	                             	
&\multicolumn{2}{c|}{Sintel Clean} 							
&\multicolumn{2}{c|}{Sintel Final}						
&\multicolumn{3}{c|}{KITTI 2012} 
&\multicolumn{3}{c|}{KITTI 2015} \\

\multirow{1}{*}{}
&\multirow{1}{*}{}
&\multicolumn{1}{c}{Train}&\multicolumn{1}{c|}{Test}
&\multicolumn{1}{c}{Train}&\multicolumn{1}{c|}{Test}
&\multicolumn{1}{c}{Train}&\multicolumn{1}{c}{Test}&\multicolumn{1}{c|}{Test (Out-Noc)}
&\multicolumn{1}{c}{Train}&\multicolumn{1}{c}{Train (Fl-all)}&\multicolumn{1}{c|}{Test (Fl-all)} \\	
\hline\
\multirow{6}{*}{\rotatebox[origin=c]{90}{Conventional}}
&\multirow{1}{*}{LDOF$^{1}$~\cite{Brox11}}				
&4.64&\multicolumn{1}{c|}{7.56}	           
&5.96&\multicolumn{1}{c|}{9.12}
&10.94&\multicolumn{1}{c}{12.4}&\multicolumn{1}{c|}{-}
&18.19&\multicolumn{1}{c}{38.11\%}&\multicolumn{1}{c|}{-}\\
           
\multirow{1}{*}{}                   
&\multirow{1}{*}{DeepFlow$^{1}$~\cite{Weinzaepfel13}}				
&2.66&\multicolumn{1}{c|}{5.38}	           
&3.57&\multicolumn{1}{c|}{7.21}
&4.48&\multicolumn{1}{c}{5.8}&\multicolumn{1}{c|}{-}
&10.63&\multicolumn{1}{c}{26.52\%}&\multicolumn{1}{c|}{29.18\%}\\
                                                   
\multirow{1}{*}{}                              
&\multirow{1}{*}{Classic+NLP~\cite{Sun14}}				
&4.49&\multicolumn{1}{c|}{6.73}	           
&7.46&\multicolumn{1}{c|}{8.29}
&-&\multicolumn{1}{c}{7.2}&\multicolumn{1}{c|}{-}
&-&\multicolumn{1}{c}{-}&\multicolumn{1}{c|}{-}\\
  
\multirow{1}{*}{}                                                                     
&\multirow{1}{*}{PCA-Layers$^{1}$~\cite{Wulff15}}				
&3.22&\multicolumn{1}{c|}{5.73}	           
&4.52&\multicolumn{1}{c|}{7.89}
&5.99&\multicolumn{1}{c}{5.2}&\multicolumn{1}{c|}{-}
&12.74&\multicolumn{1}{c}{27.26\%}&\multicolumn{1}{c|}{-}\\                                                                    

\multirow{1}{*}{}
&\multirow{1}{*}{EpicFlow$^{1}$~\cite{Revaud15}}				
&2.27&\multicolumn{1}{c|}{4.12}	           
&3.56&\multicolumn{1}{c|}{6.29}
&\textbf{3.09}&\multicolumn{1}{c}{3.8}&\multicolumn{1}{c|}{-}
&9.27&\multicolumn{1}{c}{27.18\%}&\multicolumn{1}{c|}{\textbf{27.10\%}}\\
  
\multirow{1}{*}{}                                                                                                                                        
&\multirow{1}{*}{FlowFields$^{1}$~\cite{Bailer15}}				
&\textbf{1.86}&\multicolumn{1}{c|}{\textbf{3.75}}	           
&\textbf{3.06}&\multicolumn{1}{c|}{\textbf{5.81}}
&3.33&\multicolumn{1}{c}{\textbf{3.5}}&\multicolumn{1}{c|}{-}	
&\textbf{8.33}&\multicolumn{1}{c}{\textbf{24.43\%}}&\multicolumn{1}{c|}{-}	\\
                                                      
\hline    
\multirow{3}{*}{\rotatebox[origin=c]{90}{Hybrid}}
&\multirow{1}{*}{Deep DiscreteFlow~\cite{Guney16}}				
&-&\multicolumn{1}{c|}{3.86}	           
&-&\multicolumn{1}{c|}{5.73}
&-&\multicolumn{1}{c}{3.4}&\multicolumn{1}{c|}{-}	
&-&\multicolumn{1}{c}{-}&\multicolumn{1}{c|}{21.17\%}	\\  

\multirow{1}{*}{}
&\multirow{1}{*}{Bailer \etal~\cite{Bailer17}}				
&-&\multicolumn{1}{c|}{\textbf{3.78}}	           
&-&\multicolumn{1}{c|}{5.36}
&-&\multicolumn{1}{c}{\textbf{3.0}}&\multicolumn{1}{c|}{-}	
&-&\multicolumn{1}{c}{-}&\multicolumn{1}{c|}{19.44\%}	\\  

\multirow{1}{*}{}
&\multirow{1}{*}{DC Flow~\cite{Xu17}}				
&-&\multicolumn{1}{c|}{-}	           
&-&\multicolumn{1}{c|}{\textbf{5.12}}
&-&\multicolumn{1}{c}{-}&\multicolumn{1}{c|}{-}	
&-&\multicolumn{1}{c}{-}&\multicolumn{1}{c|}{\textbf{14.86\%}}\\                                                                                                                                                                
                                                                                                                                                                                                                        
\hline
\multirow{9}{*}{\rotatebox[origin=c]{90}{Heavyweight CNN}}
&\multirow{1}{*}{FlowNetS~\cite{Dosovitskiy15}}				
&4.50&\multicolumn{1}{c|}{7.42}	           
&5.45&\multicolumn{1}{c|}{8.43}
&8.26&\multicolumn{1}{c}{-}&\multicolumn{1}{c|}{-}
&-&\multicolumn{1}{c}{-}&\multicolumn{1}{c|}{-}\\       

\multirow{1}{*}{}
&\multirow{1}{*}{FlowNetS-ft~\cite{Dosovitskiy15}}				
&(3.66)&\multicolumn{1}{c|}{6.96}	           
&(4.44)&\multicolumn{1}{c|}{7.76}
&7.52&\multicolumn{1}{c}{9.1}&\multicolumn{1}{c|}{-}
&-&\multicolumn{1}{c}{-}&\multicolumn{1}{c|}{-} \\ 
                                 
\multirow{1}{*}{}
&\multirow{1}{*}{FlowNetC~\cite{Dosovitskiy15}}				
&4.31&\multicolumn{1}{c|}{7.28}	           
&5.87&\multicolumn{1}{c|}{8.81}
&9.35&\multicolumn{1}{c}{-}&\multicolumn{1}{c|}{-}	
&-&\multicolumn{1}{c}{-}&\multicolumn{1}{c|}{-}\\

\multirow{1}{*}{}
&\multirow{1}{*}{FlowNetC-ft~\cite{Dosovitskiy15}}				
&(3.78)&\multicolumn{1}{c|}{6.85}	           
&(5.28)&\multicolumn{1}{c|}{8.51}
&8.79&\multicolumn{1}{c}{-}&\multicolumn{1}{c|}{-}	
&-&\multicolumn{1}{c}{-}&\multicolumn{1}{c|}{-}\\
  
\multirow{1}{*}{}                                                                  
&\multirow{1}{*}{FlowNet2-S$^{2}$~\cite{Ilg17}}				
&3.79&\multicolumn{1}{c|}{-}	           
&4.99&\multicolumn{1}{c|}{-}
&7.26&\multicolumn{1}{c}{-}&\multicolumn{1}{c|}{-}
&14.28&\multicolumn{1}{c}{51.06\%}&\multicolumn{1}{c|}{-}\\                                                                                                                                                                                                                                                                               

\multirow{1}{*}{}
&\multirow{1}{*}{FlowNet2-C$^{2}$~\cite{Ilg17}}				
&3.04&\multicolumn{1}{c|}{-}	           
&4.60&\multicolumn{1}{c|}{-}
&5.79&\multicolumn{1}{c}{-}&\multicolumn{1}{c|}{-}
&11.49&\multicolumn{1}{c}{44.09\%}&\multicolumn{1}{c|}{-} \\                                                                     

\multirow{1}{*}{}
&\multirow{1}{*}{FlowNet2~\cite{Ilg17}}				
&\textbf{2.02}&\multicolumn{1}{c|}{\textbf{3.96}}	           
&\textbf{3.54}$^{3}$&\multicolumn{1}{c|}{6.02}
&4.01$^{4}$&\multicolumn{1}{c}{-}&\multicolumn{1}{c|}{-}
&10.08$^{4}$&\multicolumn{1}{c}{29.99\%$^{4}$}&\multicolumn{1}{c|}{-}\\ 

\multirow{1}{*}{}                                                                                            
&\multirow{1}{*}{FlowNet2-ft-sintel~\cite{Ilg17}}				
&(1.45)&\multicolumn{1}{c|}{4.16}	           
&(2.19$^{3}$)&\multicolumn{1}{c|}{\textbf{5.74}}
&\textbf{3.54}$^{4}$&\multicolumn{1}{c}{-}&\multicolumn{1}{c|}{-}
&\textbf{9.94}$^{4}$&\multicolumn{1}{c}{\textbf{28.02\%}$^{4}$}&\multicolumn{1}{c|}{-}\\ 

\multirow{1}{*}{}
&\multirow{1}{*}{FlowNet2-ft-kitti~\cite{Ilg17}}				
&3.43&\multicolumn{1}{c|}{-}	           
&4.83$^{3}$&\multicolumn{1}{c|}{-}
&(1.43$^{4}$)&\multicolumn{1}{c}{\textbf{1.8}}&\multicolumn{1}{c|}{4.82\%}
&(2.36$^{4}$)&\multicolumn{1}{c}{(8.88\%$^{4}$)}&\multicolumn{1}{c|}{\textbf{11.48\%}}\\ 

\hline
\multirow{14}{*}{\rotatebox[origin=c]{90}{Lightweight CNN}}
\multirow{1}{*}{}
&\multirow{1}{*}{SPyNet~\cite{Ranjan17}}				
&4.12&\multicolumn{1}{c|}{6.69}	           
&5.57&\multicolumn{1}{c|}{8.43}
&9.12&\multicolumn{1}{c}{-}&\multicolumn{1}{c|}{-}
&-&\multicolumn{1}{c}{-}&\multicolumn{1}{c|}{-}\\  

\multirow{1}{*}{}
&\multirow{1}{*}{SPyNet-ft~\cite{Ranjan17}}				
&(3.17)&\multicolumn{1}{c|}{6.64}	           
&(4.32)&\multicolumn{1}{c|}{8.36}
&\textit{3.36}$^{6}$&\multicolumn{1}{c}{4.1}&\multicolumn{1}{c|}{12.31\%}	
&-&\multicolumn{1}{c}{-}&\multicolumn{1}{c|}{35.07\%}	\\  

\multirow{1}{*}{}
&\multirow{1}{*}{PWC-Net~\cite{Sun18}}				
&2.55&\multicolumn{1}{c|}{-}	           
&3.93&\multicolumn{1}{c|}{-}
&4.14&\multicolumn{1}{c}{-}&\multicolumn{1}{c|}{-}
&10.35&\multicolumn{1}{c}{33.67\%}&\multicolumn{1}{c|}{-}\\  

\multirow{1}{*}{}
&\multirow{1}{*}{PWC-Net-ft~\cite{Sun18}}				
&(2.02)&\multicolumn{1}{c|}{4.39}	           
&(2.08)&\multicolumn{1}{c|}{5.04}
&(1.45)&\multicolumn{1}{c}{1.7}&\multicolumn{1}{c|}{4.22\%}
&(2.16)&\multicolumn{1}{c}{(9.80\%)}&\multicolumn{1}{c|}{9.60\%}\\  

\multirow{1}{*}{}
&\multirow{1}{*}{PWC-Net\_ROB~\cite{Sun19}}				
&(1.81)&\multicolumn{1}{c|}{3.90}	           
&(2.29)&\multicolumn{1}{c|}{4.90}
&-&\multicolumn{1}{c}{-}&\multicolumn{1}{c|}{-}
&-&\multicolumn{1}{c}{-}&\multicolumn{1}{c|}{11.63\%}\\

\multirow{1}{*}{}
&\multirow{1}{*}{PWC-Net-ft+~\cite{Sun19}}				
&(1.71)&\multicolumn{1}{c|}{\textbf{3.45}}	           
&(2.34)&\multicolumn{1}{c|}{\textbf{4.60}}
&(0.99)&\multicolumn{1}{c}{\textbf{1.4}}&\multicolumn{1}{c|}{3.36\%}	
&(1.47)&\multicolumn{1}{c}{(7.59\%)}&\multicolumn{1}{c|}{\underline{7.72\%}}\\ 

\multirow{1}{*}{}
&\multirow{1}{*}{LiteFlowNetX-pre~\cite{Hui18}}				
&3.70&\multicolumn{1}{c|}{-}	           
&4.82&\multicolumn{1}{c|}{-}
&6.81&\multicolumn{1}{c}{-}&\multicolumn{1}{c|}{-}		
&16.64&\multicolumn{1}{c}{36.64\%}&\multicolumn{1}{c|}{-} \\                                                       

\multirow{1}{*}{}
&\multirow{1}{*}{LiteFlowNetX~\cite{Hui18}}				
&3.58&\multicolumn{1}{c|}{-}	           
&4.79&\multicolumn{1}{c|}{-}
&6.38&\multicolumn{1}{c}{-}&\multicolumn{1}{c|}{-}		
&15.81&\multicolumn{1}{c}{34.90\%}&\multicolumn{1}{c|}{-}\\                                                                                                                                                                                                                                                
                                                                
\multirow{1}{*}{}                                                                    
&\multirow{1}{*}{LiteFlowNet-pre~\cite{Hui18}}				
&2.78&\multicolumn{1}{c|}{-}	           
&4.17&\multicolumn{1}{c|}{-}
&4.56&\multicolumn{1}{c}{-}&\multicolumn{1}{c|}{-}		
&11.58&\multicolumn{1}{c}{32.59\%}&\multicolumn{1}{c|}{-}\\                                                                                       

\multirow{1}{*}{}
&\multirow{1}{*}{LiteFlowNet$^{7}$~\cite{Hui18}}				
&2.48&\multicolumn{1}{c|}{-}	           
&4.04&\multicolumn{1}{c|}{-}
&4.00&\multicolumn{1}{c}{-}&\multicolumn{1}{c|}{-}		
&10.39&\multicolumn{1}{c}{28.50\%}&\multicolumn{1}{c|}{-}\\                                                                                      

\multirow{1}{*}{}
&\multirow{1}{*}{LiteFlowNet-ft$^{7}$~\cite{Hui18}}				
&(1.35)&\multicolumn{1}{c|}{4.54}           
&(1.78)&\multicolumn{1}{c|}{5.38}
&(1.05)&\multicolumn{1}{c}{\underline{1.6}}&\multicolumn{1}{c|}{\underline{3.27\%}}	
&(1.62)&\multicolumn{1}{c}{(5.58\%)}&\multicolumn{1}{c|}{9.38\%}\\    

\multirow{1}{*}{}                                                                    
&\multirow{1}{*}{LiteFlowNet2-pre}				
&2.78&\multicolumn{1}{c|}{-}	           
&4.14&\multicolumn{1}{c|}{-}
&4.11&\multicolumn{1}{c}{-}&\multicolumn{1}{c|}{-}		
&11.31&\multicolumn{1}{c}{32.12\%}&\multicolumn{1}{c|}{-}\\                                                                                       

\multirow{1}{*}{}
&\multirow{1}{*}{LiteFlowNet2}				
&\textbf{2.32}&\multicolumn{1}{c|}{-}           
&\textbf{3.85}&\multicolumn{1}{c|}{-}
&\textbf{3.77}&\multicolumn{1}{c}{-}&\multicolumn{1}{c|}{-}		
&\textbf{9.83}&\multicolumn{1}{c}{\textbf{28.45\%}}&\multicolumn{1}{c|}{-}\\  

\multirow{1}{*}{}
&\multirow{1}{*}{LiteFlowNet2-ft}				
&(1.41)&\multicolumn{1}{c|}{\underline{3.48}}           
&(1.83)&\multicolumn{1}{c|}{\underline{4.69}}
&(0.95)&\multicolumn{1}{c}{\textbf{1.4}}&\multicolumn{1}{c|}{\textbf{2.63\%}}		
&(1.33)&\multicolumn{1}{c}{(4.32\%)}&\multicolumn{1}{c|}{\textbf{7.62\%}}\\ 
\hline
\end{tabular}}
\vspace{-0.5em}
\end{table*}

%-------------------------------------------------------------------------
\subsection{Results} 
\label{sec:results}
%-------------------------------------------------------------------------
We evaluate LiteFlowNet and LiteFlowNet2 against the state-of-the-art methods on the public optical flow benchmarks including FlyingChairs (Chairs)~\cite{Dosovitskiy15}, Sintel Clean and Final~\cite{Butler12}, KITTI~2012~\cite{Geiger12}, and KITTI 2015~\cite{Menze15}. Average end-point error (AEE) and specialized percentage error are reported.

\vspace{0.1cm}
\noindent
\textbf{FlyingChairs.} We first compare the intermediate results of several well-performing networks trained on Chairs alone in Table~\ref{tab:flyingchairs results}. LiteFlowNet-pre outperforms the compared networks. No intermediate result is available for FlowNet2~\cite{Ilg17} as each stacking network is trained on the Chairs~$\rightarrow$~Things3D schedule individually. Since FlowNetC, FlowNetS (variants of FlowNet~\cite{Dosovitskiy15}), and SPyNet~\cite{Ranjan17} have fewer parameters than FlowNet2 and the latter two models do not perform feature matching, we construct a small-size counterpart \textbf{LiteFlowNetX-pre} for a fair comparison by removing the matching part and shrinking the model sizes of NetC and NetE by about 4 and 5 times, respectively. Despite LiteFlowNetX-pre is 43 and 1.33 times smaller than FlowNetC and SPyNet, respectively, it outperforms these networks and is on par with FlowNetC that uses explicit feature matching. As shown in Table~\ref{tab:results}, LiteFlowNet2-pre which is also trained on Chairs is on par with LiteFlowNet on Sintel and outperforms LiteFlowNet on KITTI.

\begin{figure*}[t]
\begin{center}
\captionsetup[subfigure]{labelformat=empty, justification=centering}
\captionsetup[subfloat]{farskip=0pt,captionskip=0pt}
\begin{tabular}{cccccc}
   \includegraphics[width=3.0cm]{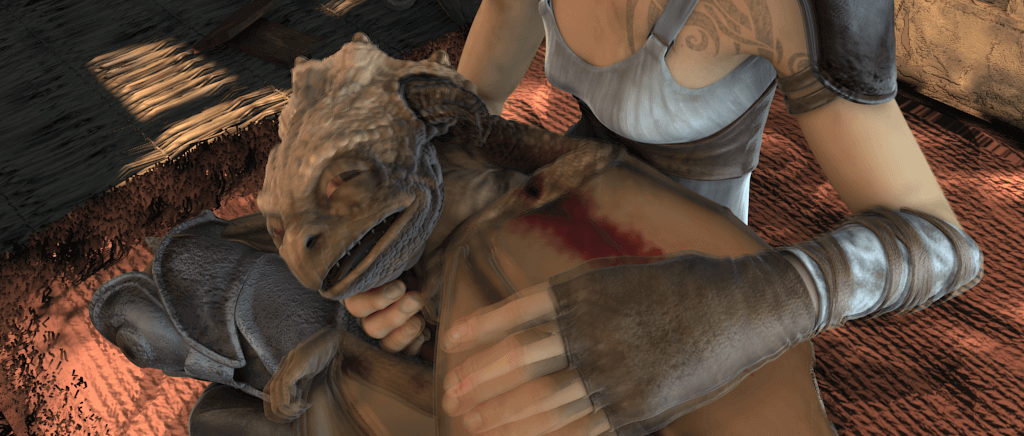}\hfill
   \includegraphics[width=3.0cm]{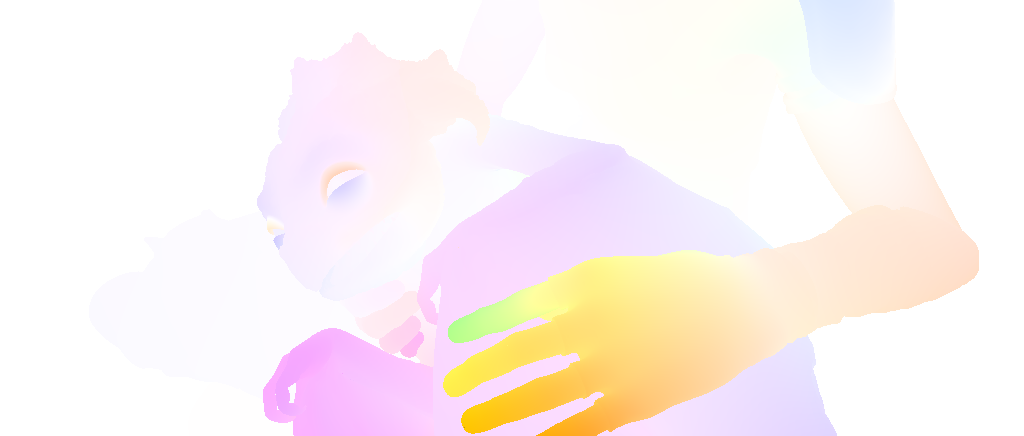}\hfill
   \includegraphics[width=3.0cm]{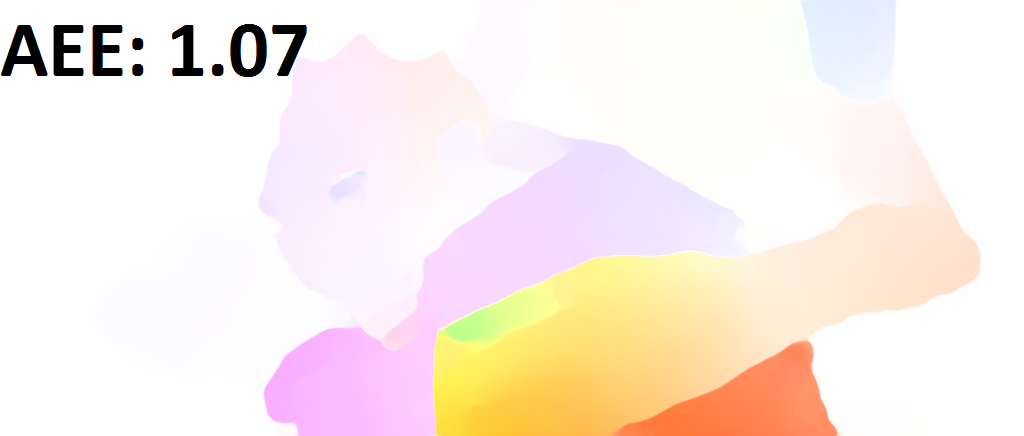}\hfill
   \includegraphics[width=3.0cm]{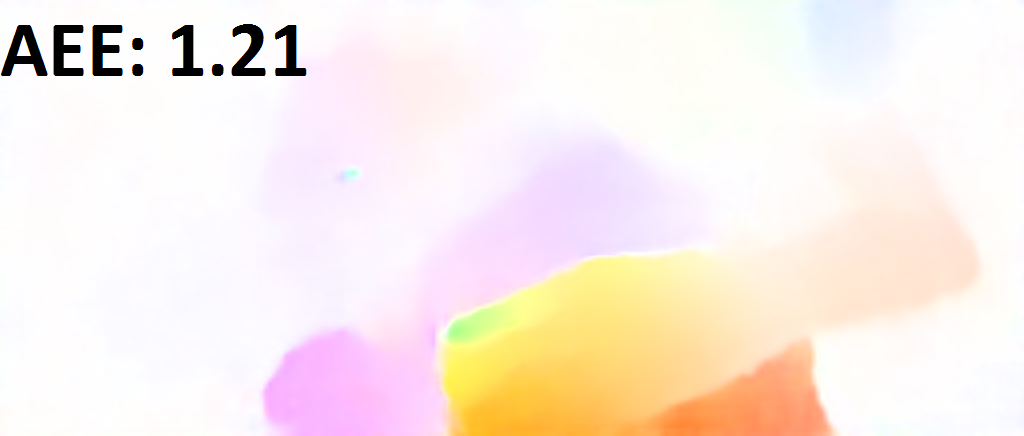}\hfill
   \includegraphics[width=3.0cm]{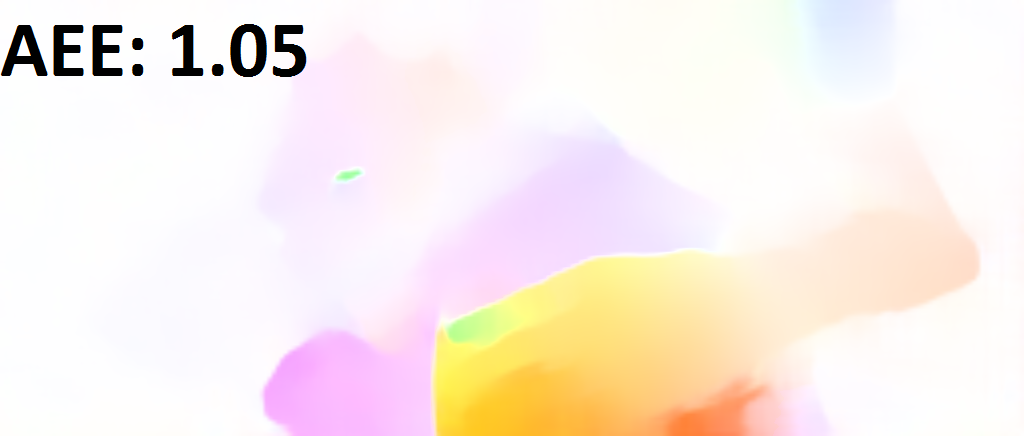}\hfill
   \includegraphics[width=3.0cm]{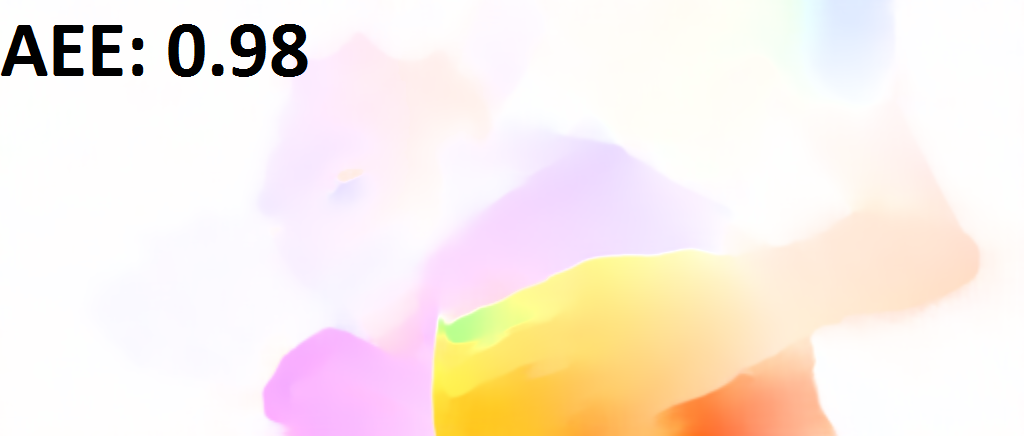}\\
   
   \includegraphics[width=3.0cm]{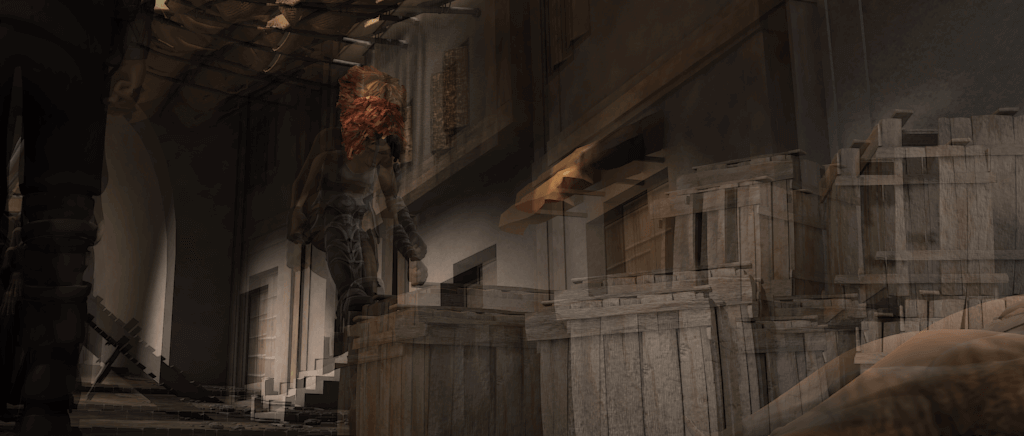}\hfill
   \includegraphics[width=3.0cm]{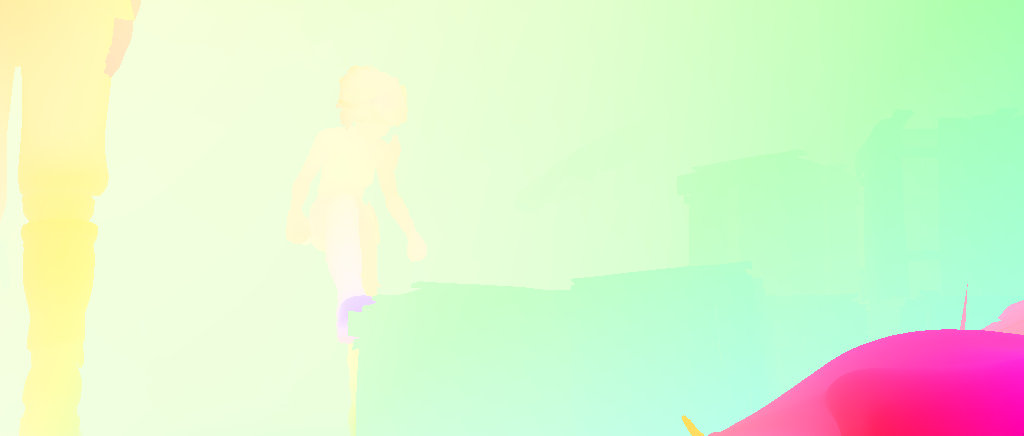}\hfill
   \includegraphics[width=3.0cm]{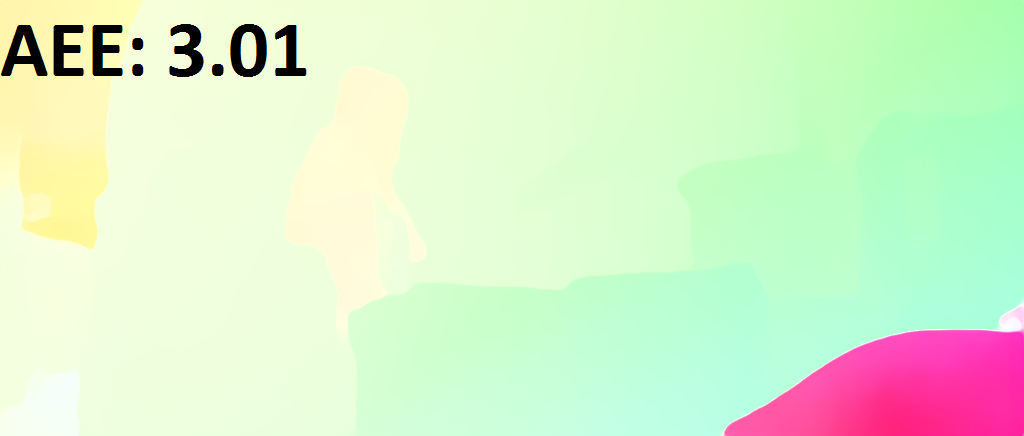}\hfill
   \includegraphics[width=3.0cm]{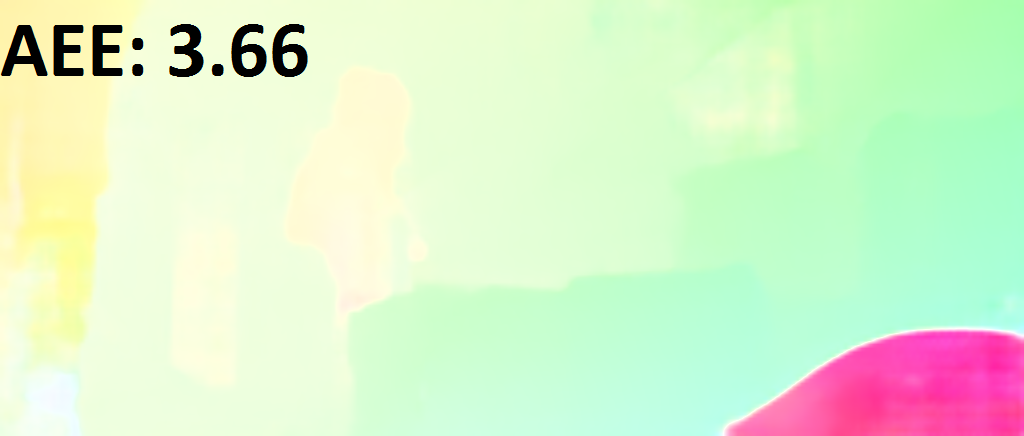}\hfill
   \includegraphics[width=3.0cm]{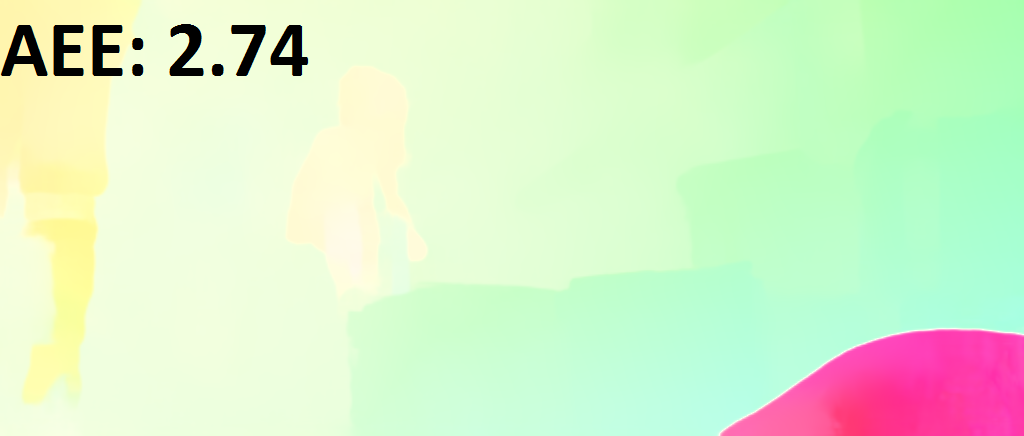}\hfill
   \includegraphics[width=3.0cm]{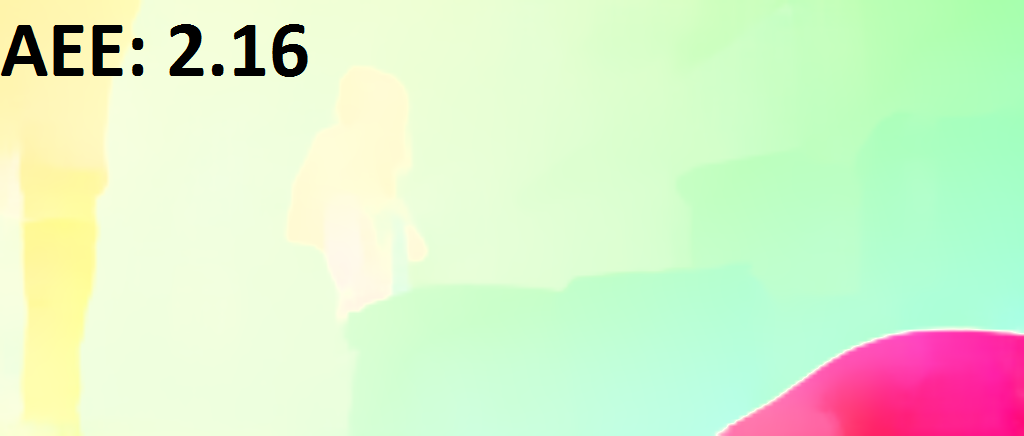} \\
   
   \includegraphics[width=3.0cm]{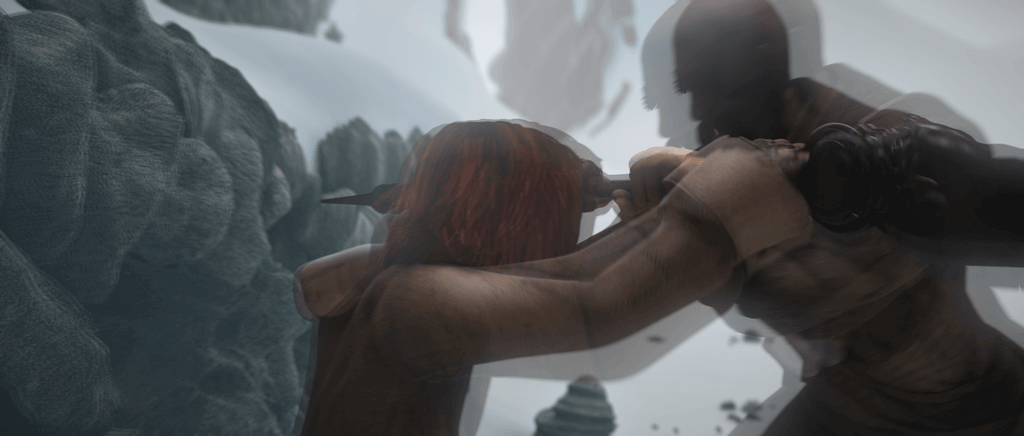}\hfill
   \includegraphics[width=3.0cm]{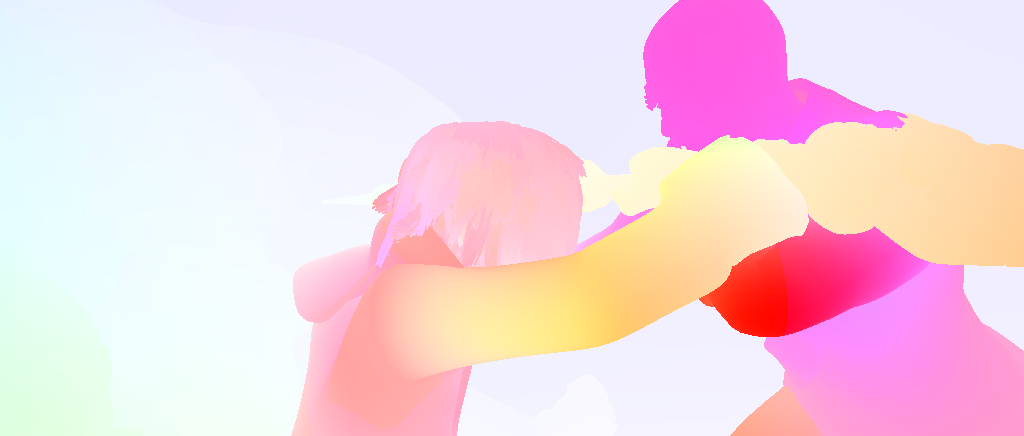}\hfill
   \includegraphics[width=3.0cm]{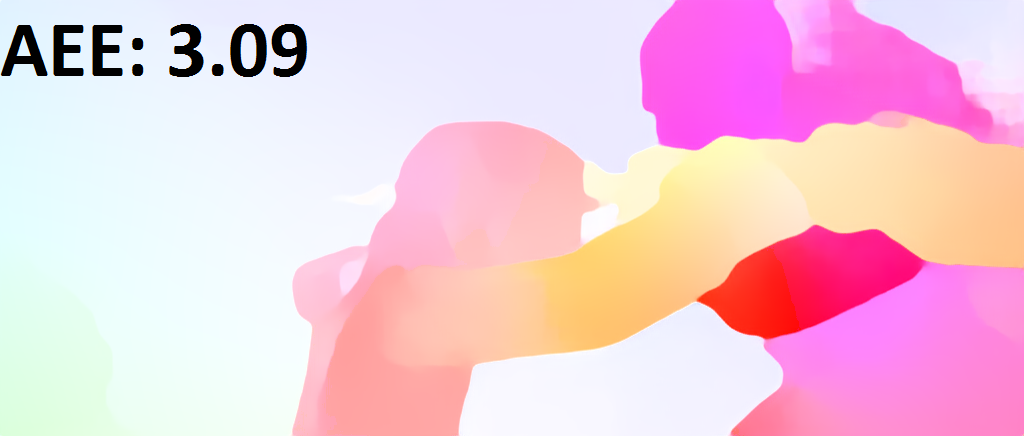}\hfill
   \includegraphics[width=3.0cm]{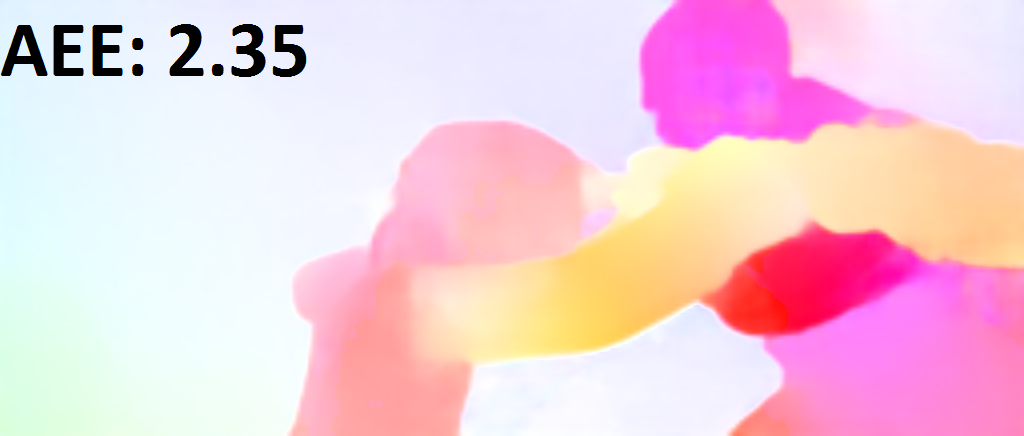}\hfill
   \includegraphics[width=3.0cm]{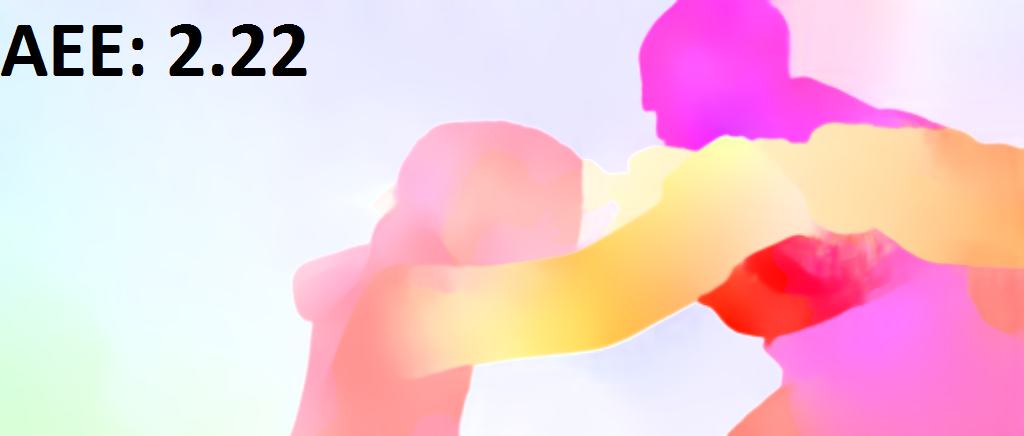}\hfill
   \includegraphics[width=3.0cm]{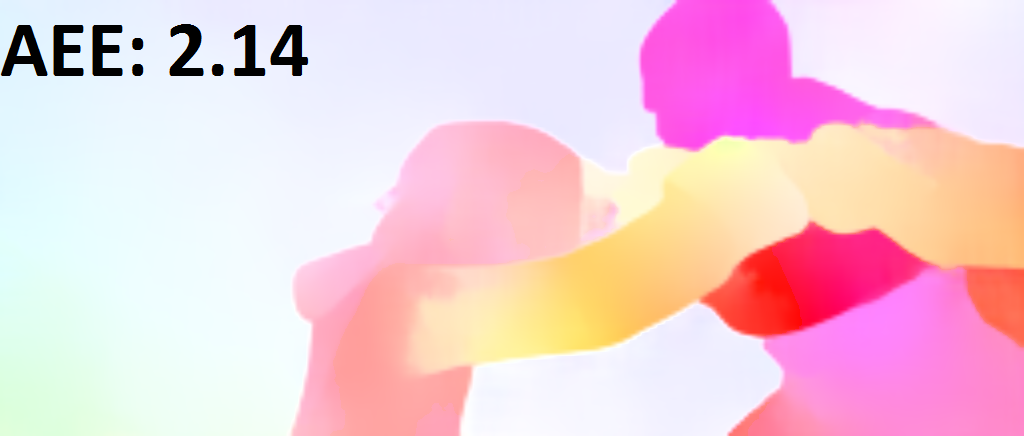}\\
   
   \subfloat[Image overlay]{\includegraphics[width=3.0cm]{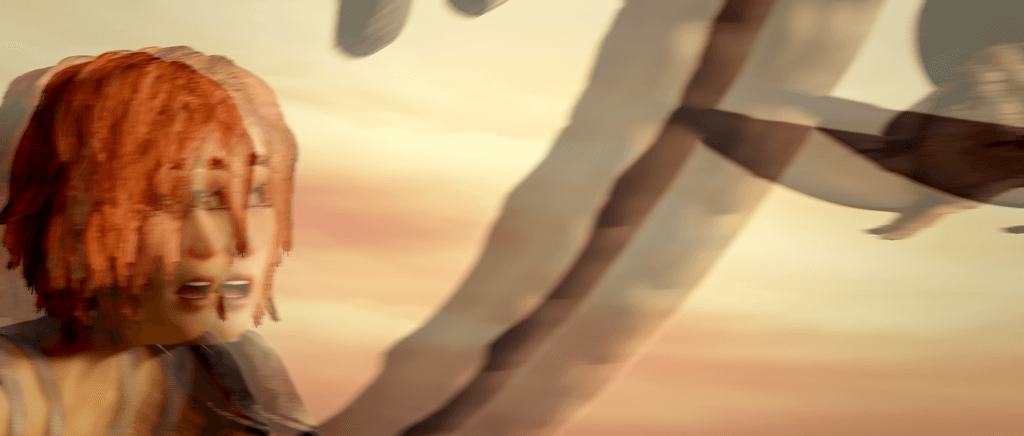}}\hfill
   \subfloat[Ground truth]{\includegraphics[width=3.0cm]{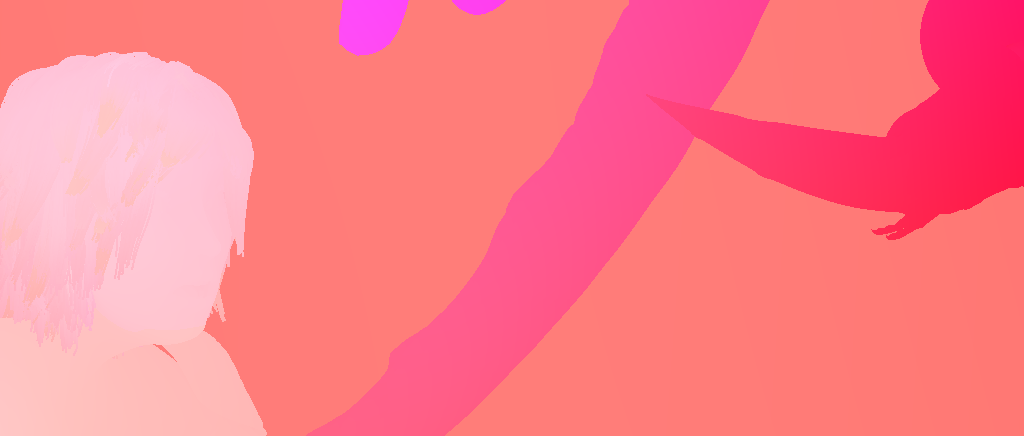}}\hfill
   \subfloat[FlowNet2~\cite{Ilg17}]{\includegraphics[width=3.0cm]{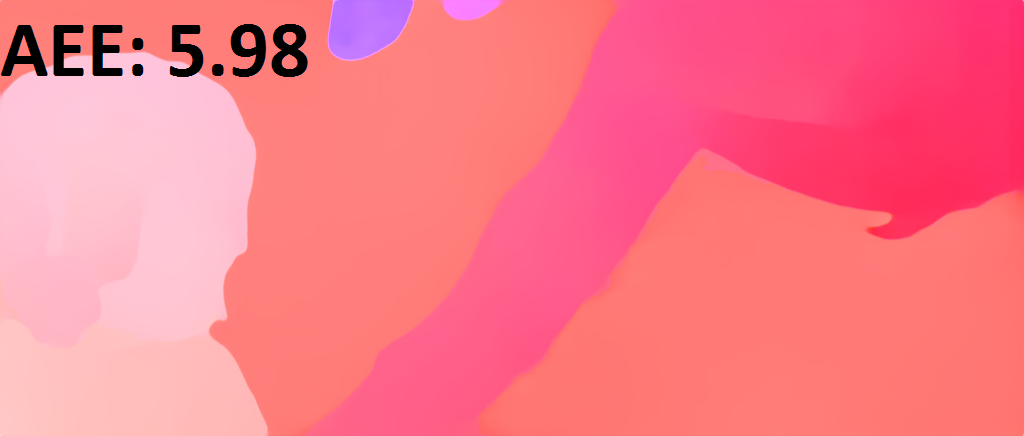}}\hfill
   \subfloat[PWC-Net$^{1}$~\cite{Sun18}]{\includegraphics[width=3.0cm]{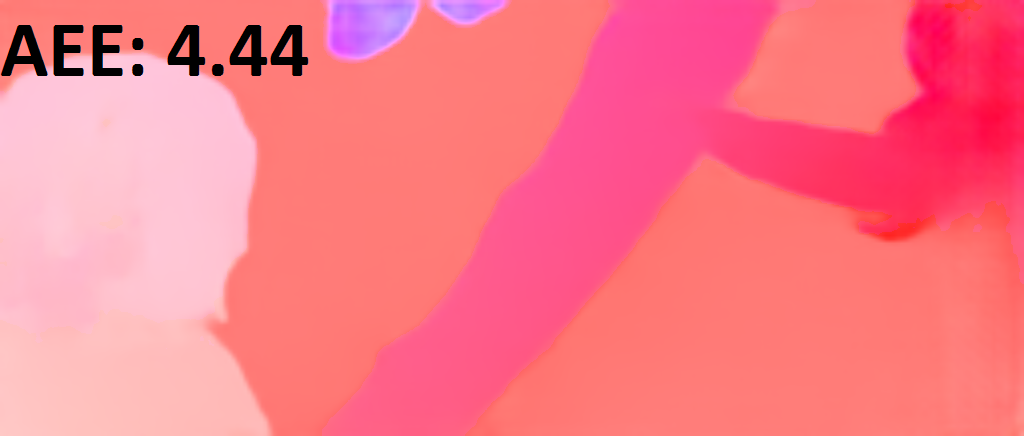}}\hfill
   \subfloat[LiteFlowNet~\cite{Hui18}]{\includegraphics[width=3.0cm]{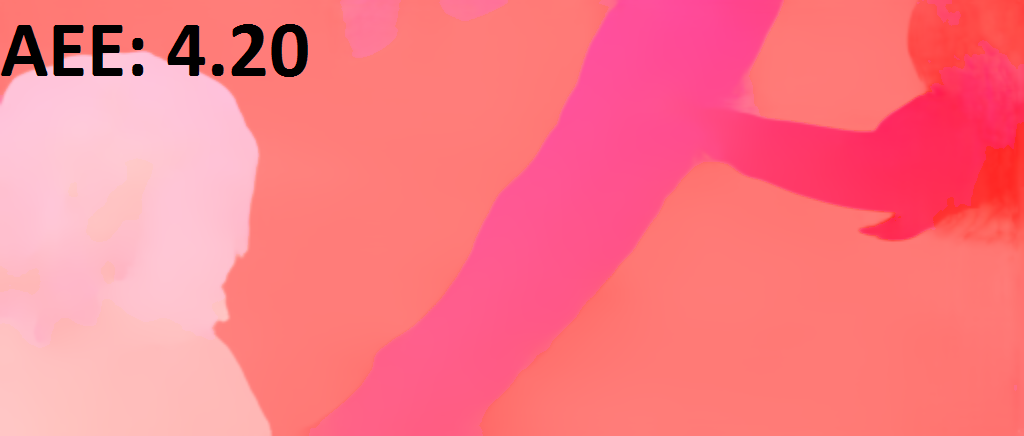}}\hfill
   \subfloat[LiteFlowNet2]{\includegraphics[width=3.0cm]{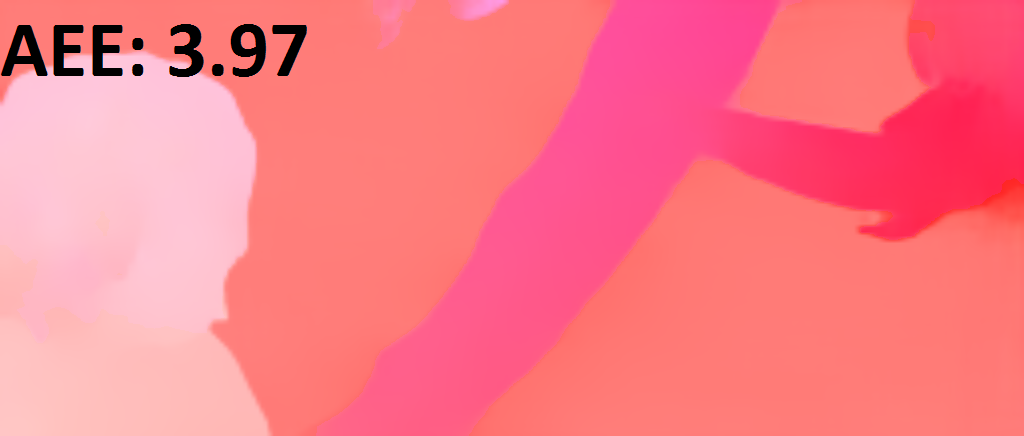}}\\  
   
   \includegraphics[width=3.0cm]{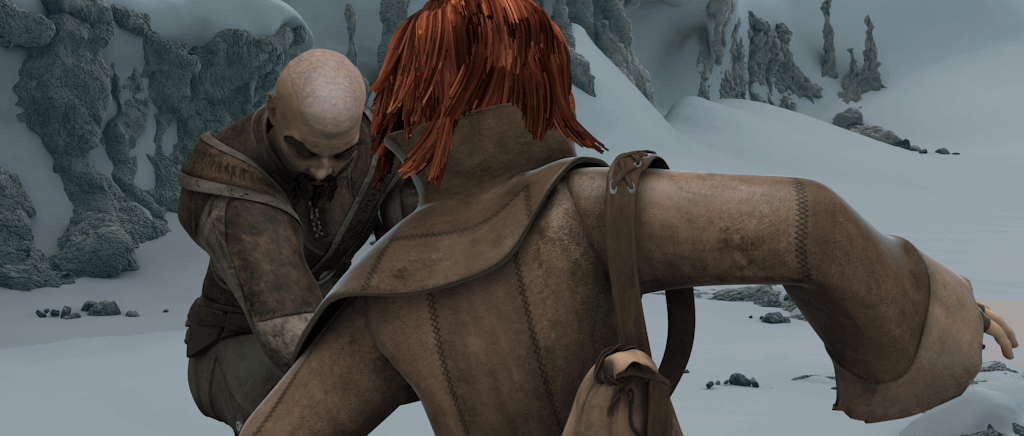}\hfill
   \includegraphics[width=3.0cm]{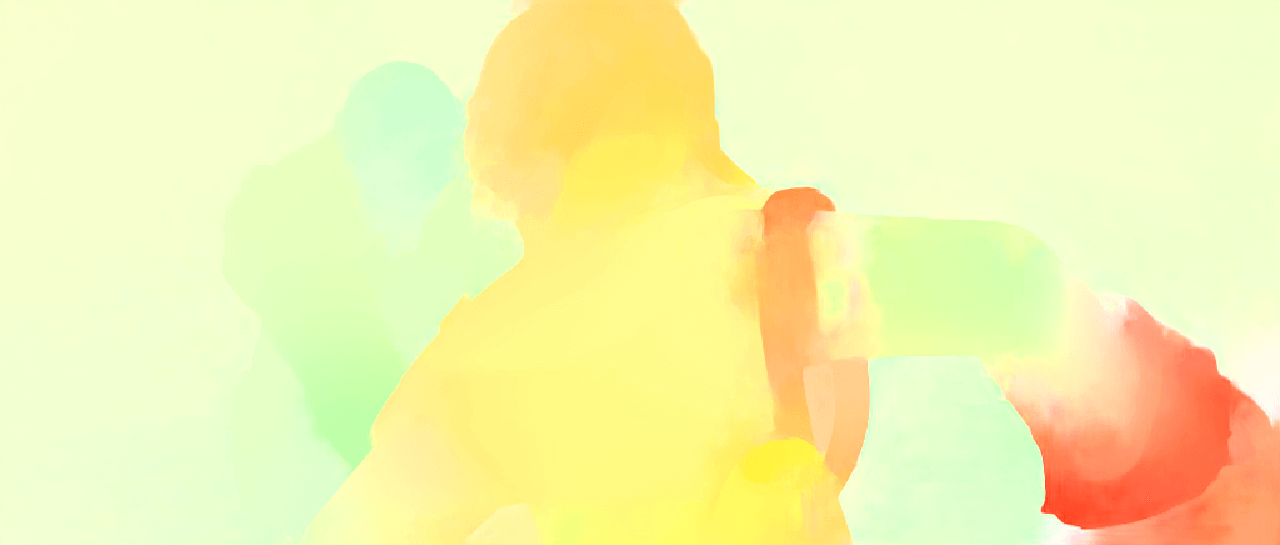}\hfill
   \includegraphics[width=3.0cm]{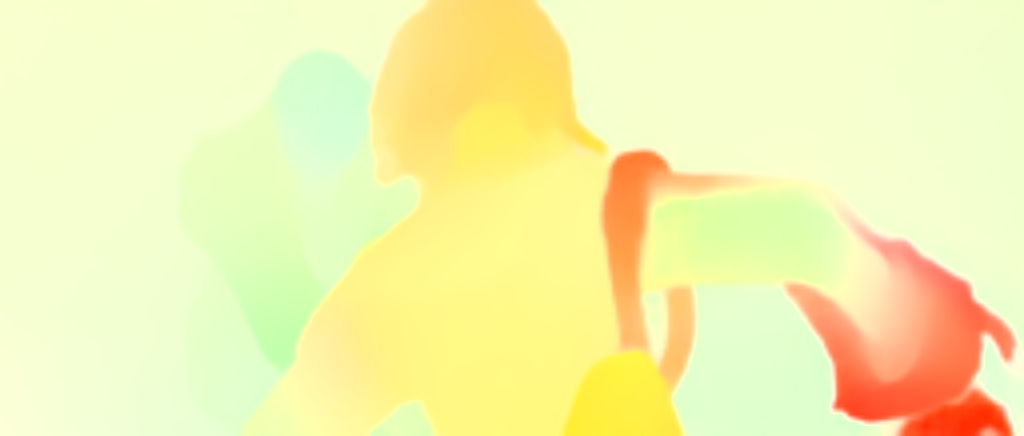}\hfill
   \includegraphics[width=3.0cm]{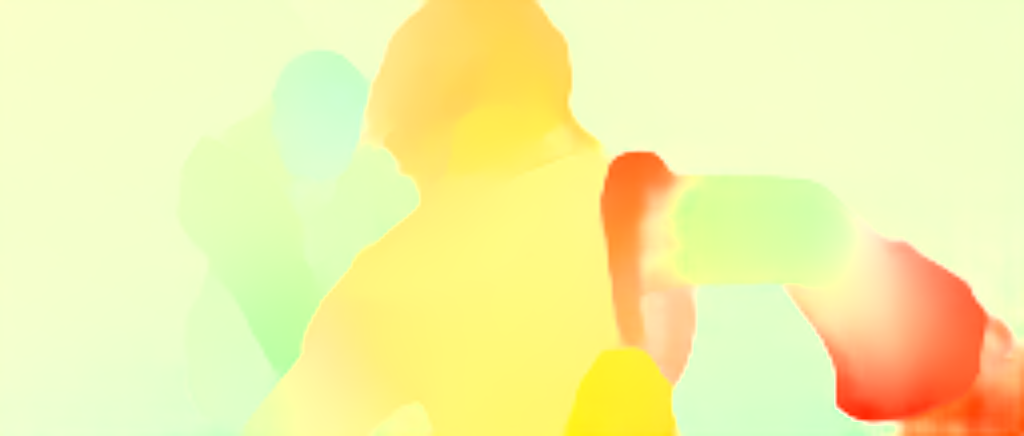}\hfill
   \includegraphics[width=3.0cm]{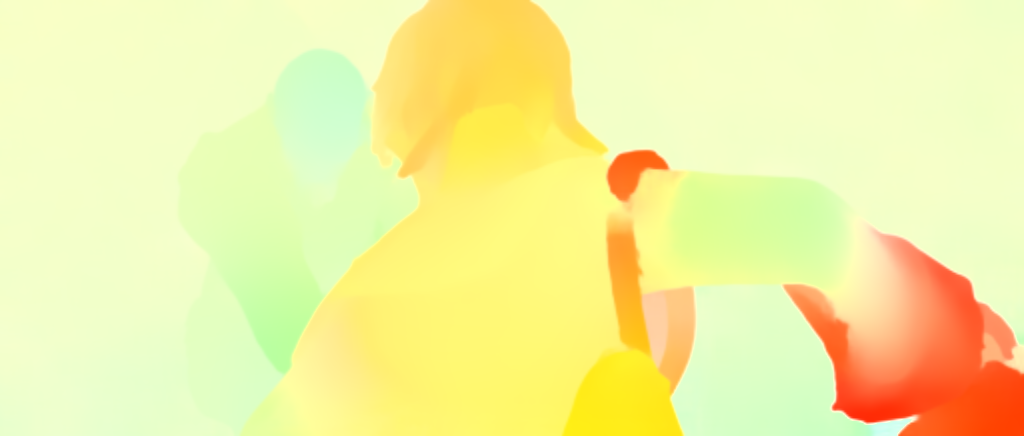}\hfill
   \includegraphics[width=3.0cm]{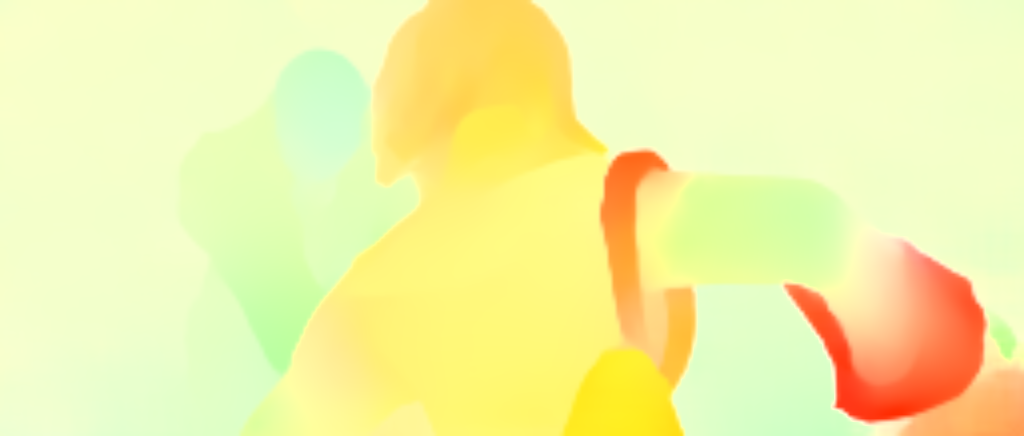}\\
   
   \subfloat[First image]{\includegraphics[width=3.0cm]{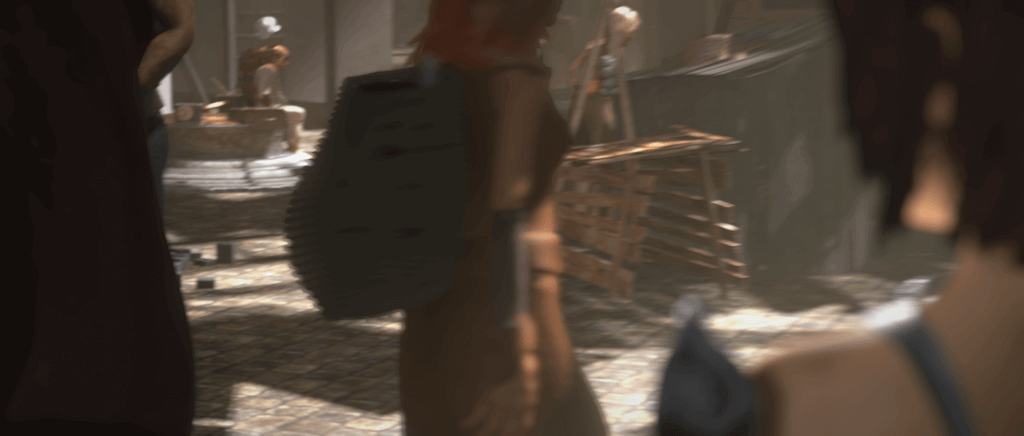}}\hfill
   \subfloat[SPyNet-ft~\cite{Ranjan17}]{\includegraphics[width=3.0cm]{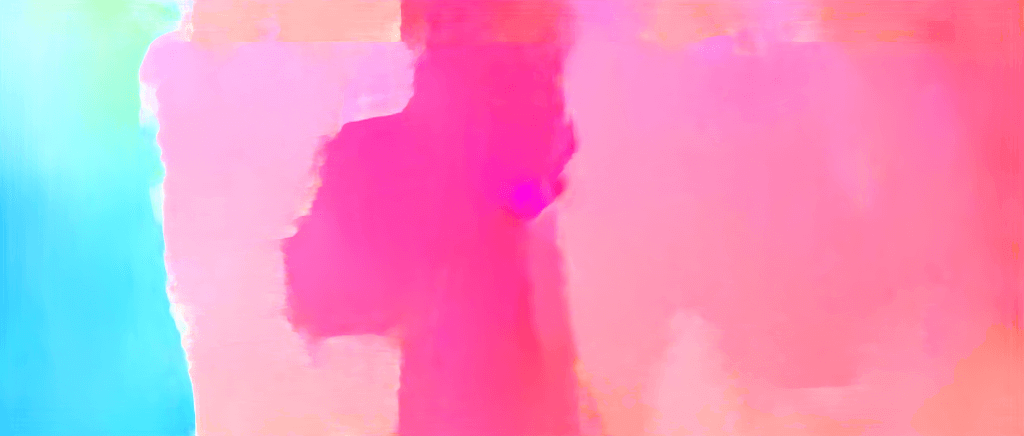}}\hfill
   \subfloat[FlowNet2-ft~\cite{Ilg17}]{\includegraphics[width=3.0cm]{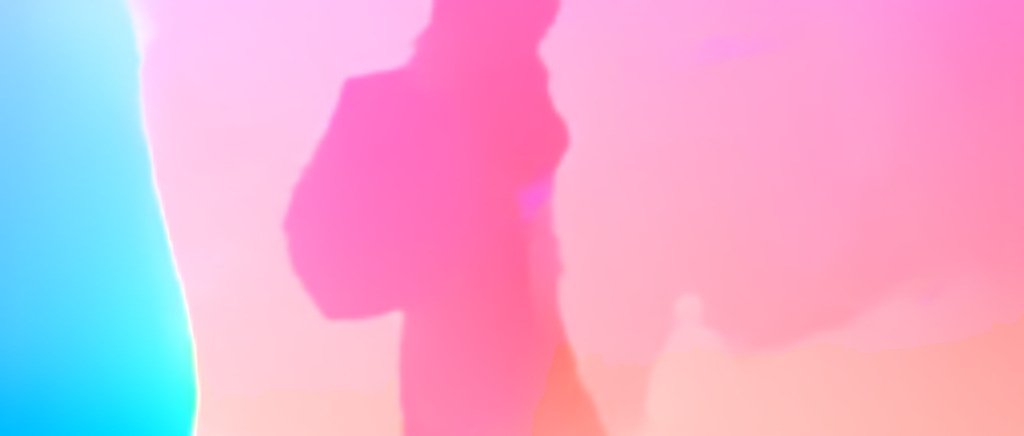}}\hfill
   \subfloat[PWC-Net+~\cite{Sun19}]{\includegraphics[width=3.0cm]{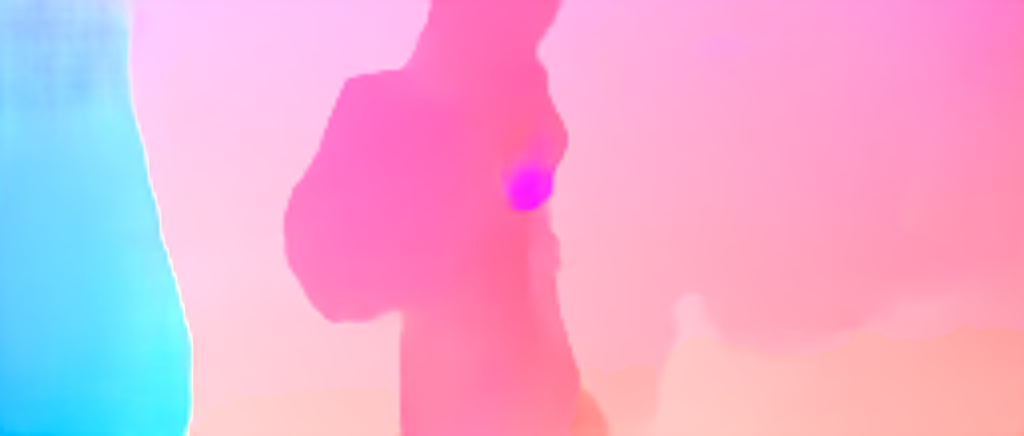}}\hfill
   \subfloat[LiteFlowNet-ft~\cite{Hui18}]{\includegraphics[width=3.0cm]{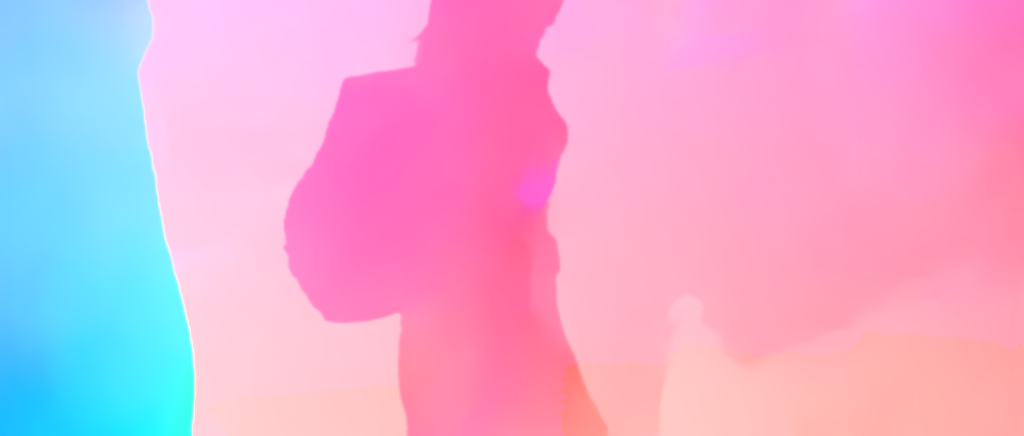}}\hfill
   \subfloat[LiteFlowNet2-ft]{\includegraphics[width=3.0cm]{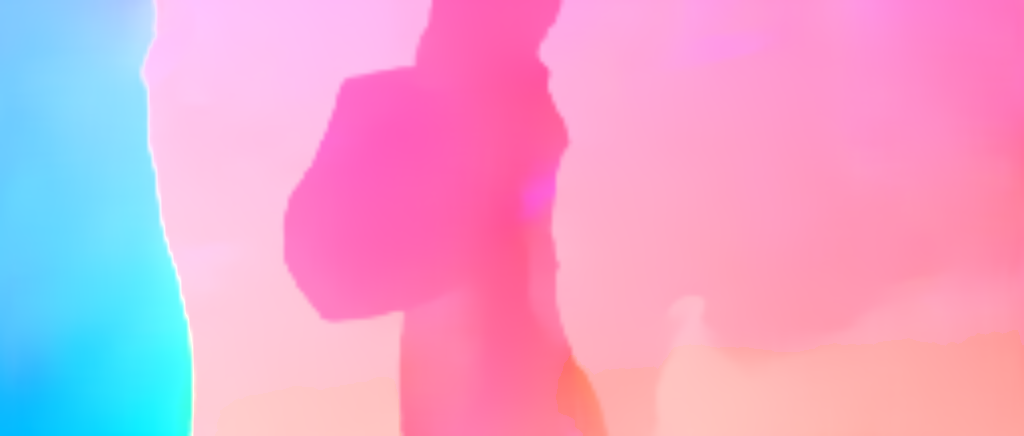}}\\
\end{tabular}
\end{center}
\vspace{-0.3cm}
\caption{Examples of flow fields from different methods on Sintel training sets (clean pass: first to second rows, final pass: third to fourth rows) and testing sets (clean pass: fifth row, final pass: last row). Fine details are well preserved and less artifacts can be observed in the flow fields of LiteFlowNet2 and LiteFlowNet2-ft. For the best visual comparison, it is recommended to enlarge the figure electronically. (Note: $^{1}$At the time of submission, the authors~\cite{Sun18} only release the trained model of PWC-Net that uses a larger feature encoder (overall footprint: 9.37M vs 8.75M) and has a slower runtime (41.12ms vs 39.63ms) trained on Chairs~$\rightarrow$~Things3D.)}
\label{fig:Sintel flows}
\vspace{-0.5em}
\end{figure*}

\vspace{0.1cm}
\noindent
\textbf{MPI Sintel.} The results are summarized in Table~\ref{tab:results}. LiteFlowNetX-pre outperforms FlowNetS~\cite{Dosovitskiy15}, FlowNetC~\cite{Dosovitskiy15}, and SPyNet~\cite{Ranjan17} that are trained on Chairs on all cases. LiteFlowNet, trained on the Chairs~$\rightarrow$~Things3D schedule, performs better than LiteFlowNet-pre as expected. LiteFlowNet also outperforms SPyNet, FlowNet2-S~\cite{Ilg17}, and FlowNet2-C~\cite{Ilg17}. It is on par with PWC-Net~\cite{Sun18}.
With the improved architecture and training protocol, LiteFlowNet2 outperforms its predecessor LiteFlowNet and PWC-Net.
We also fine-tuned LiteFlowNet (\textbf{LiteFlowNet-ft}) on a mixture of Sintel Clean and Final training data using the generalized Charbonnier loss with the settings $\epsilon^{2} = 0.01$ and $q = 0.2$. We randomly crop $768\times384$ patches and use a batch size of 4. No noise augmentation is performed but we introduce image mirroring~\cite{Sun18} to improve the diversity of the training set. Learning rate is set to 5e-5 and the training schedule is similar to the training on Things3D except it is trained for 600K and is re-trained with a reduced learning rate for a reduced number of iterations. LiteFlowNet-ft outperforms FlowNet2-ft-sintel~\cite{Ilg17} and EpicFlow~\cite{Revaud15} on Sintel Final testing set. It is on par with PWC-Net-ft~\cite{Sun18}. Despite DC Flow~\cite{Xu17} (a hybrid method consists of CNN and post-processing) performs better than LiteFlowNet, its GPU runtime requires several seconds that makes it formidable in many applications. 
For fine-tuning LiteFlowNet2 (\textbf{LiteFlowNet2-ft}), we further improve the diversity of the training set by using a mixture of Sintel and KITTI data  for a batch size of 4 containing two image pairs from each of the training sets. Unlike PWC-Net+~\cite{Sun19}, our mixture does not contains HD1K dataset~\cite{Kondermann16} as we have experienced that there is no significant improvement after including it. LiteFlowNet2-ft outperforms all the compared methods and is on par with PWC-Net-ft+ on Sintel Clean and Final testing sets. On the other hand, when LiteFlowNet2 is fine-tuned on the same training set as LiteFlowNet (\ie containing Sintel training set only), AEEs are increased from 3.48 to 3.83 and 4.69 to 5.06 on Sintel Clean and Final testing sets, respectively. Nevertheless, it still outperforms LiteFlowNet.
We also train LiteFlowNet using the new fine-tuning protocol. AEE is decreased from 4.54 to 4.01 and 5.38 to 5.21 on the testing sets of Sintel Clean and Final, respectively.
Some examples of flow fields on the training and testing sets of Sintel are provided in Figure~\ref{fig:Sintel flows}. Since LiteFlowNet(-ft) and LiteFlowNet2(-ft) have flow regularization, sharper flow boundaries and lesser artifacts can be observed in the resulting flow fields. 

\begin{figure*}[t]
\begin{center}
\captionsetup[subfigure]{labelformat=empty, justification=centering}
\captionsetup[subfloat]{farskip=0pt,captionskip=0pt}
\begin{tabular}{cccccc}
	\includegraphics[width=3.0cm,trim={2cm 0 2cm 0},clip]{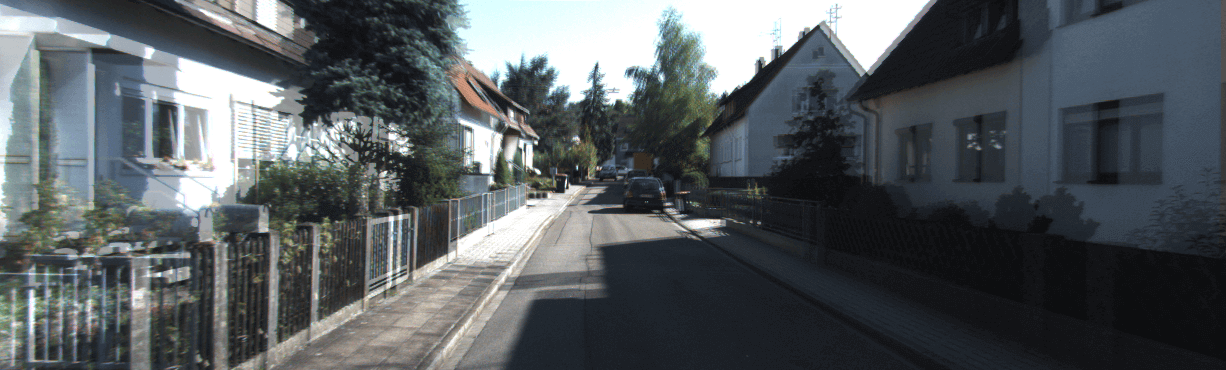}\hfill
	\includegraphics[width=3.0cm,trim={2cm 0 2cm 0},clip]{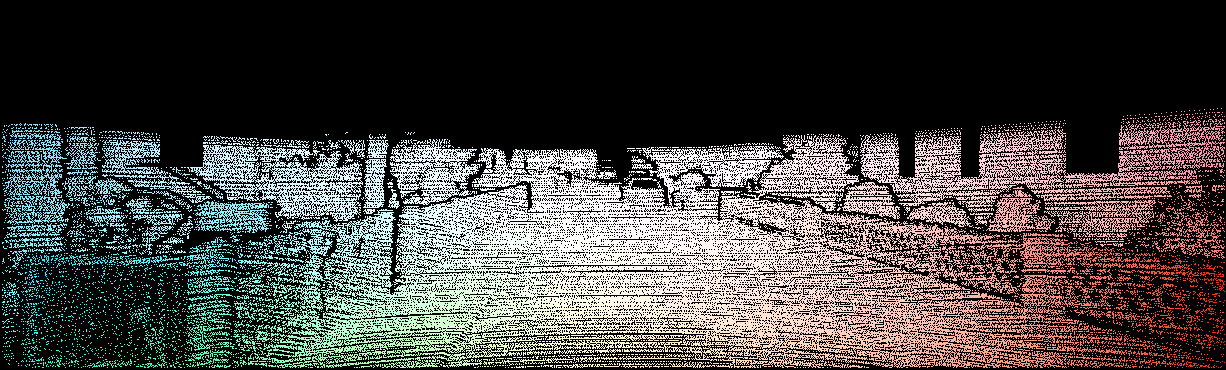}\hfill
	\includegraphics[width=3.0cm,trim={2cm 0 2cm 0},clip]{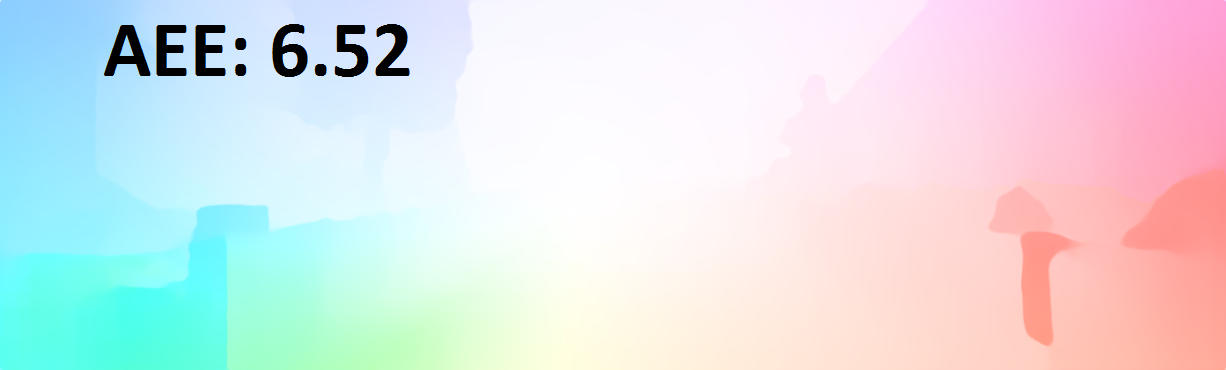}\hfill
	\includegraphics[width=3.0cm,trim={2cm 0 2cm 0},clip]{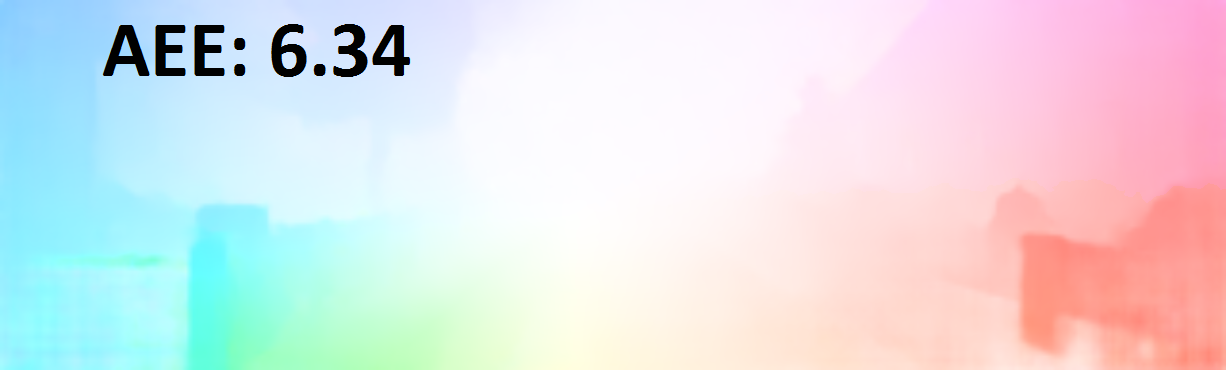}\hfill
	\includegraphics[width=3.0cm,trim={2cm 0 2cm 0},clip]{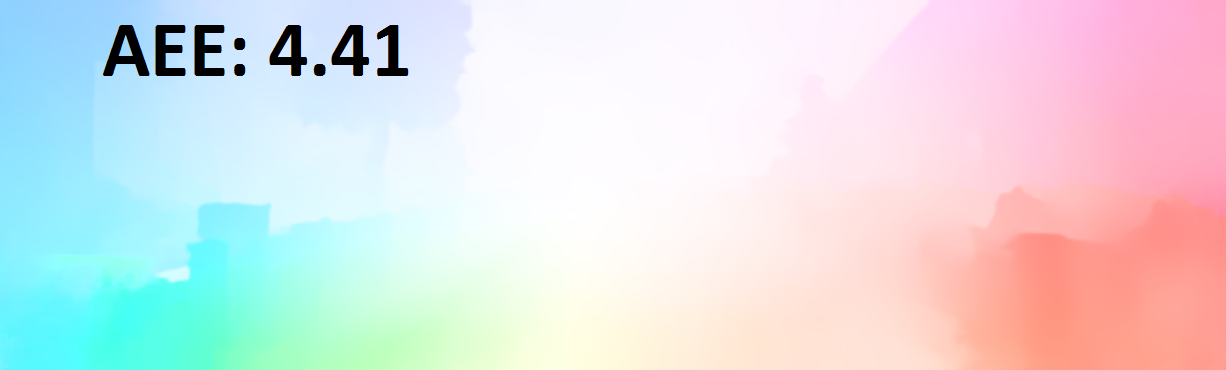}\hfill
	\includegraphics[width=3.0cm,trim={2cm 0 2cm 0},clip]{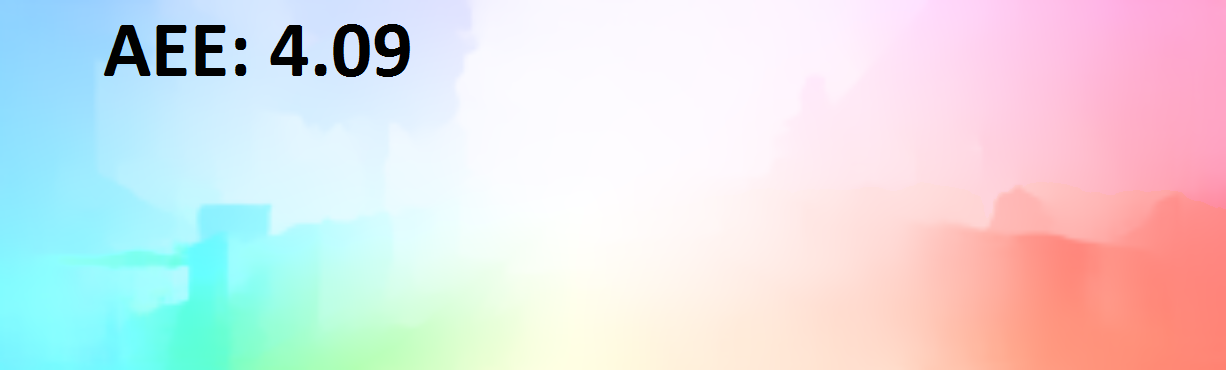}\\
	
	\includegraphics[width=3.0cm,trim={2cm 0 2cm 0},clip]{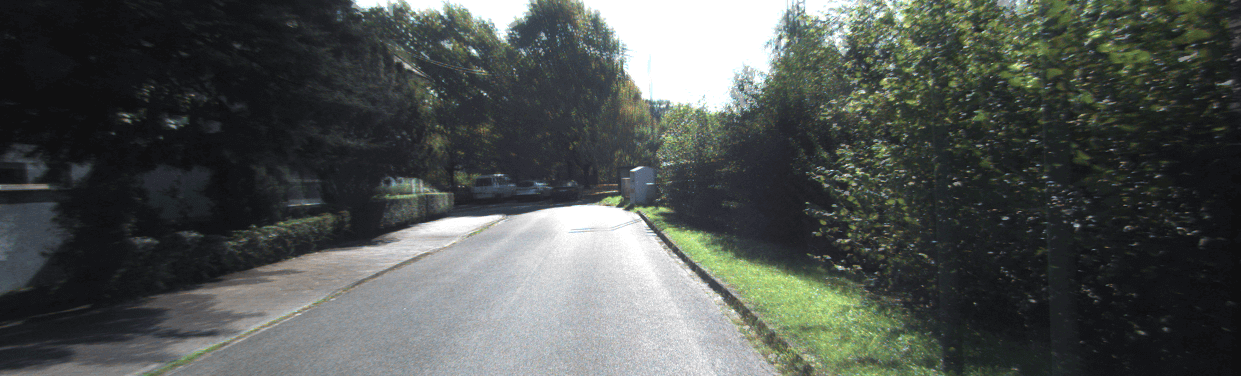}\hfill
	\includegraphics[width=3.0cm,trim={2cm 0 2cm 0},clip]{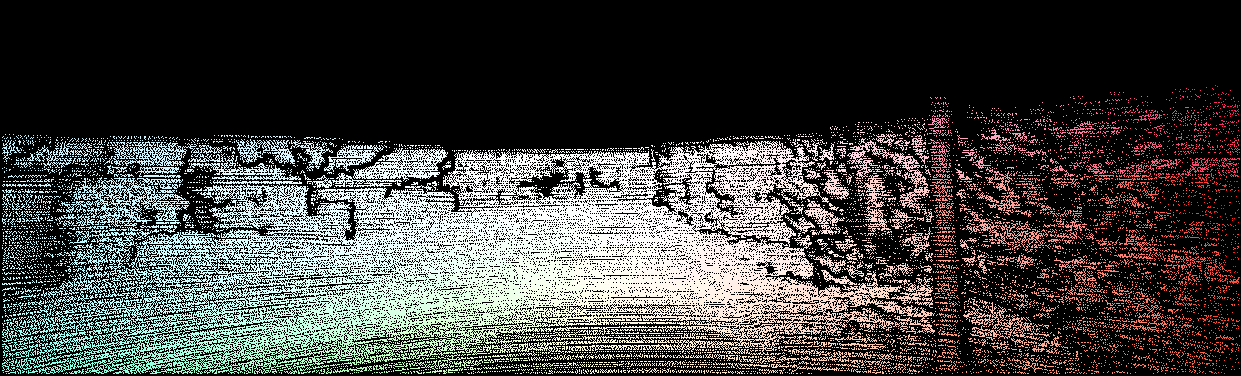}\hfill
	\includegraphics[width=3.0cm,trim={2cm 0 2cm 0},clip]{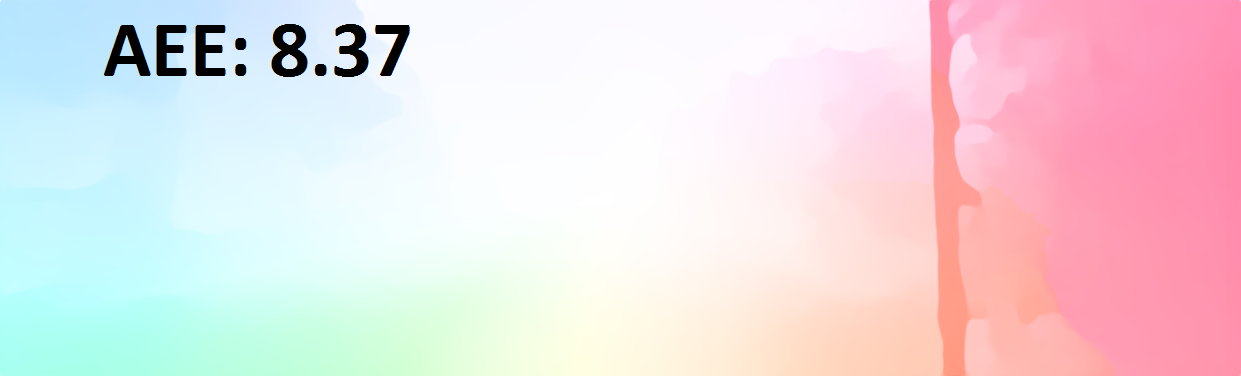}\hfill
	\includegraphics[width=3.0cm,trim={2cm 0 2cm 0},clip]{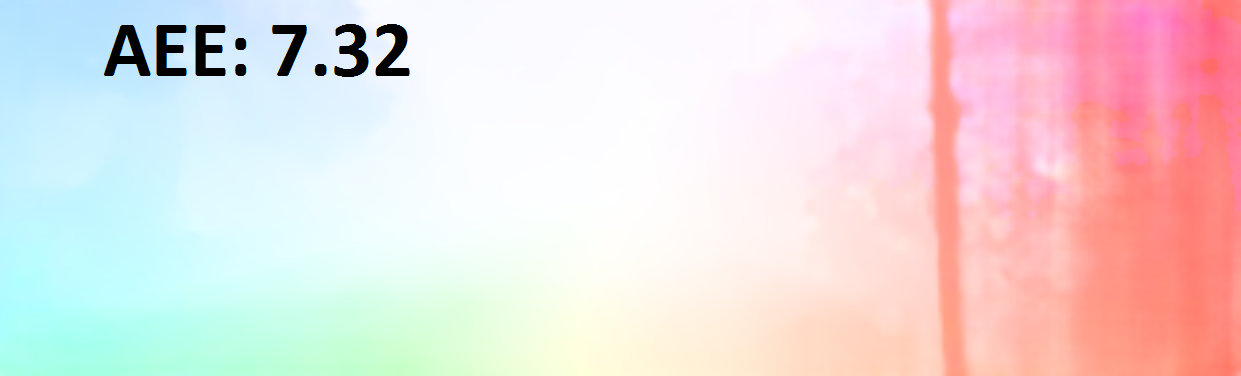}\hfill
	\includegraphics[width=3.0cm,trim={2cm 0 2cm 0},clip]{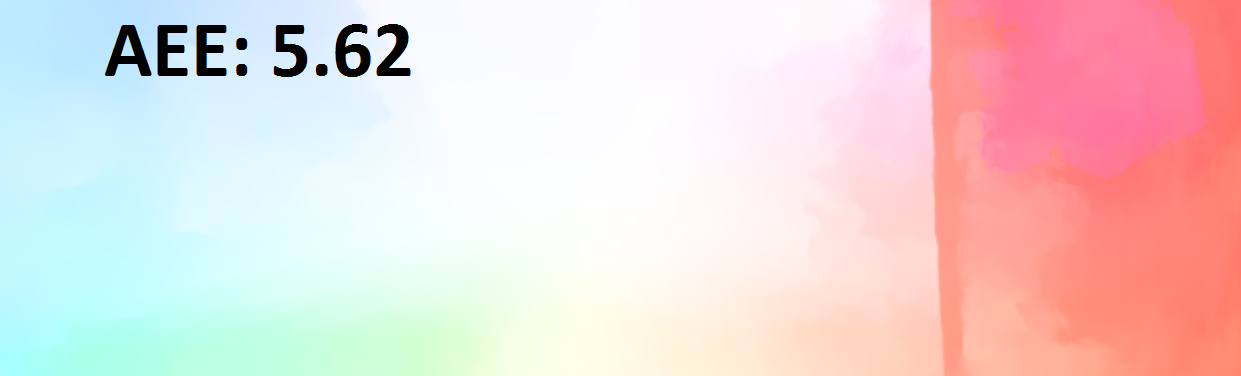}\hfill
	\includegraphics[width=3.0cm,trim={2cm 0 2cm 0},clip]{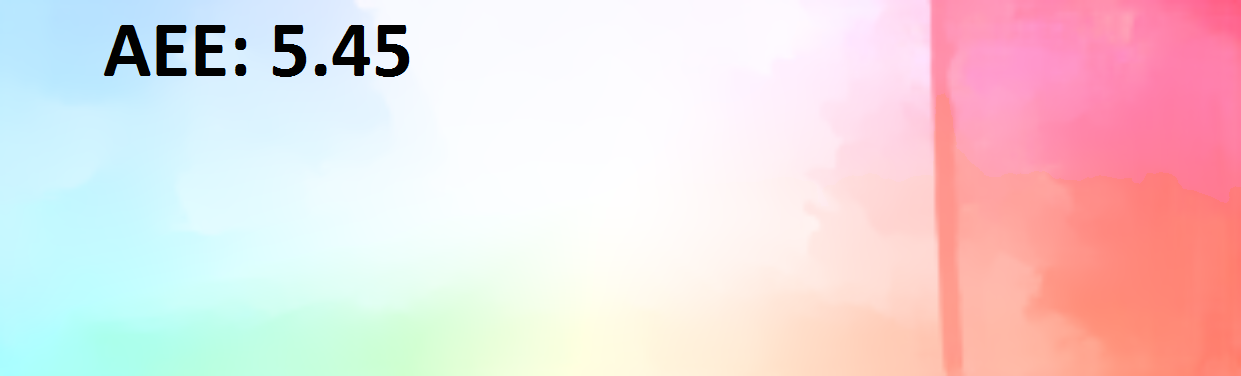}\\
	
	\includegraphics[width=3.0cm,trim={2cm 0 2cm 0},clip]{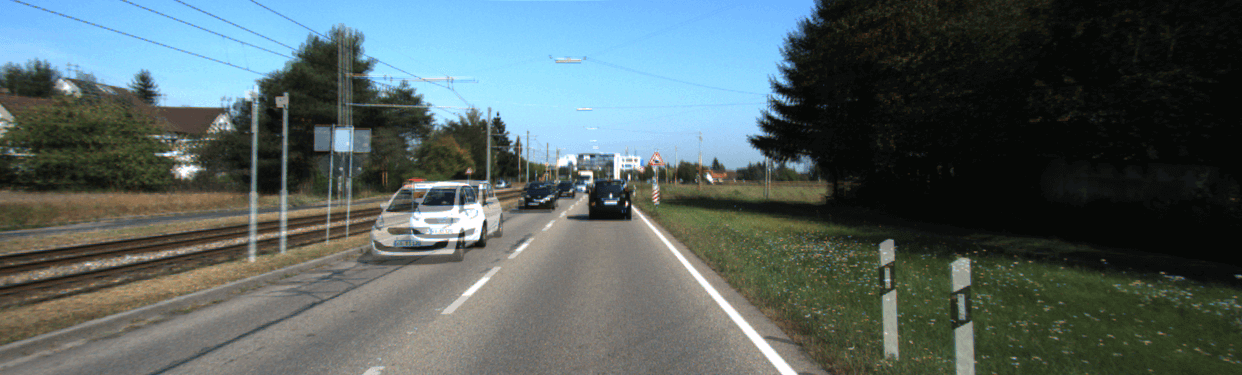}\hfill
	\includegraphics[width=3.0cm,trim={2cm 0 2cm 0},clip]{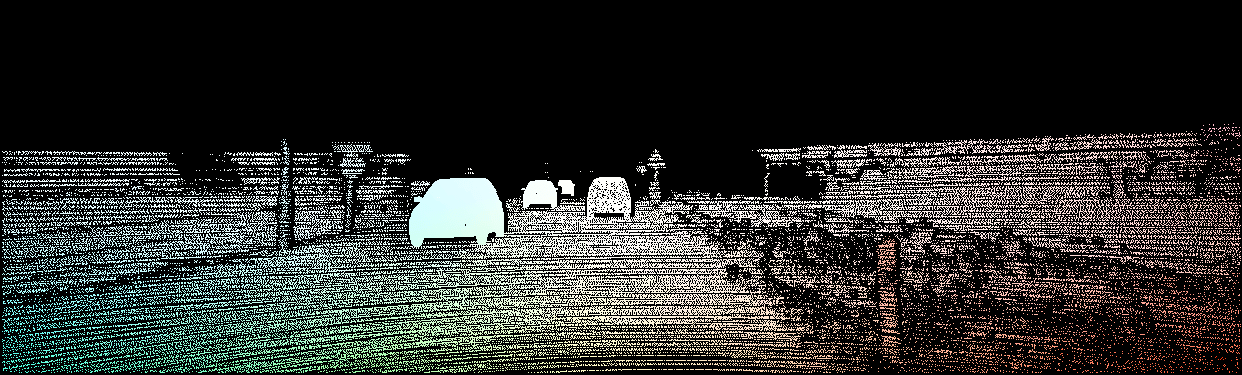}\hfill
	\includegraphics[width=3.0cm,trim={2cm 0 2cm 0},clip]{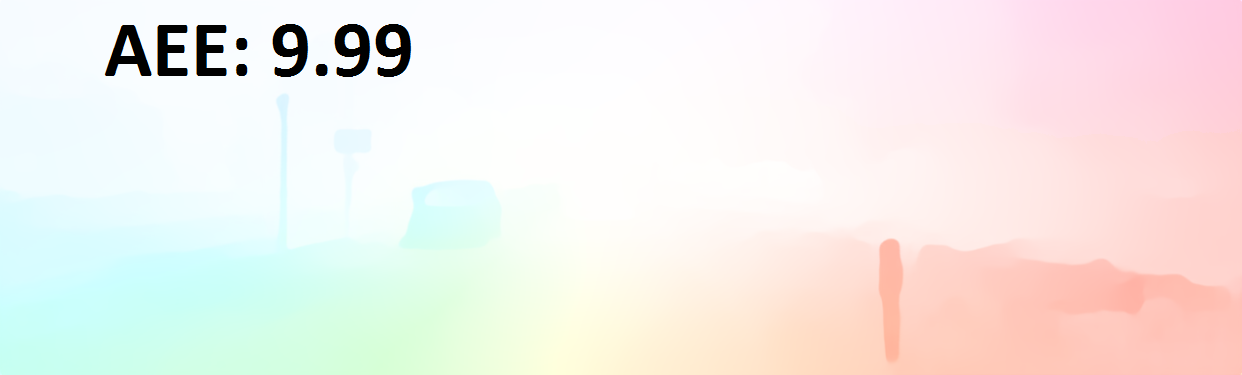}\hfill
	\includegraphics[width=3.0cm,trim={2cm 0 2cm 0},clip]{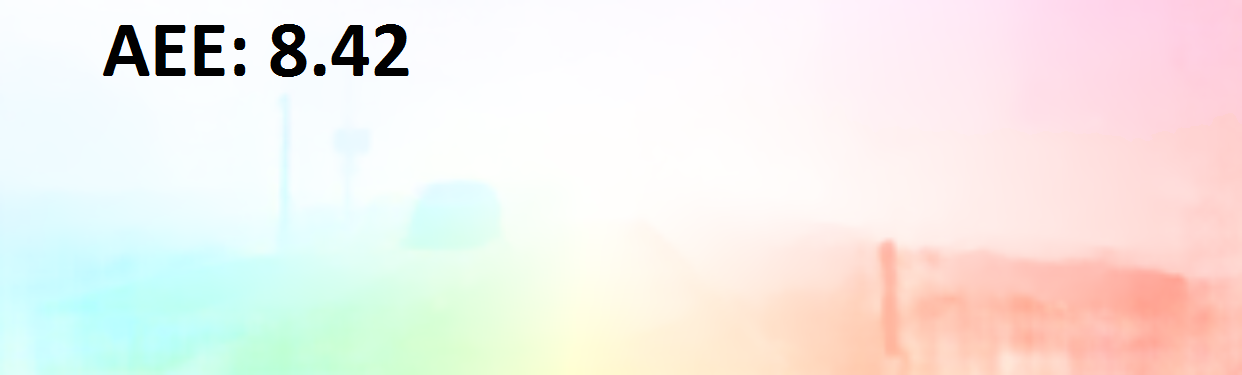}\hfill
	\includegraphics[width=3.0cm,trim={2cm 0 2cm 0},clip]{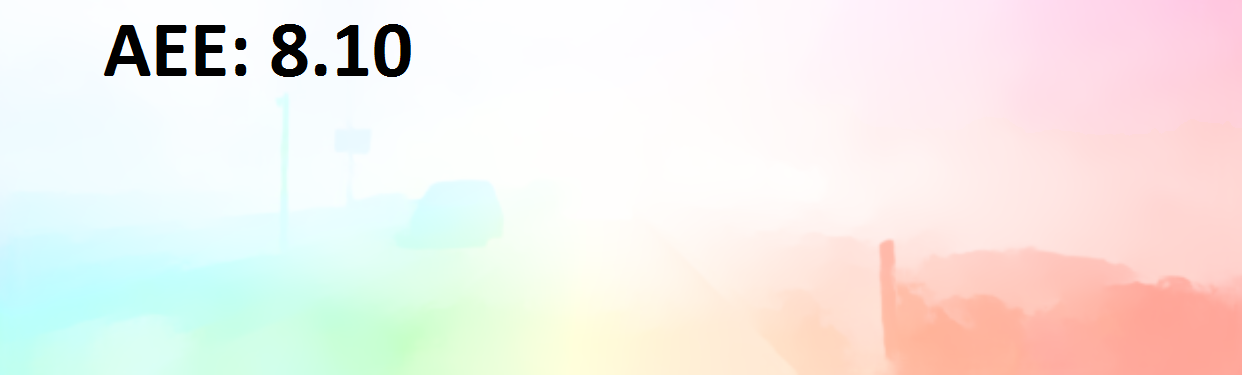}\hfill
	\includegraphics[width=3.0cm,trim={2cm 0 2cm 0},clip]{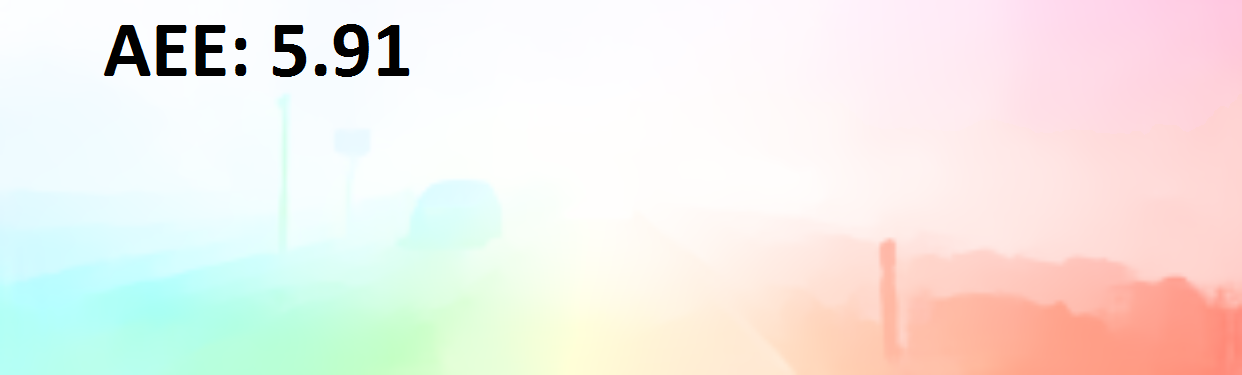}\\
		
	\subfloat[Image overlay]{\includegraphics[width=3.0cm,trim={2cm 0 2cm 0},clip]{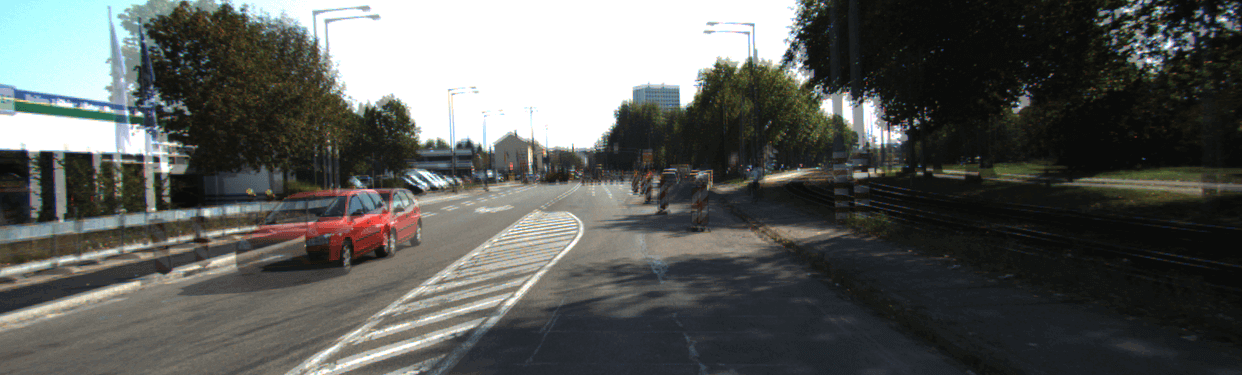}}\hfill
	\subfloat[Ground truth]{\includegraphics[width=3.0cm,trim={2cm 0 2cm 0},clip]{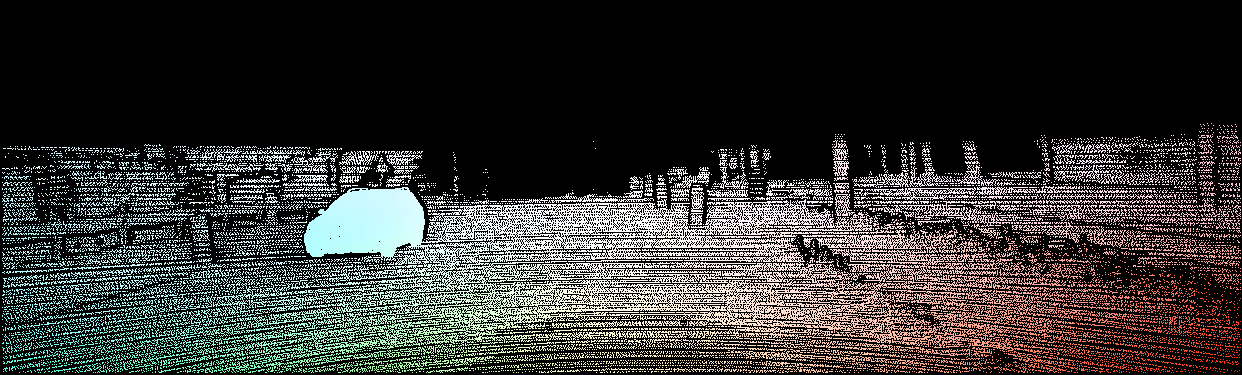}}\hfill
	\subfloat[FlowNet2~\cite{Ilg17}]{\includegraphics[width=3.0cm,trim={2cm 0 2cm 0},clip]{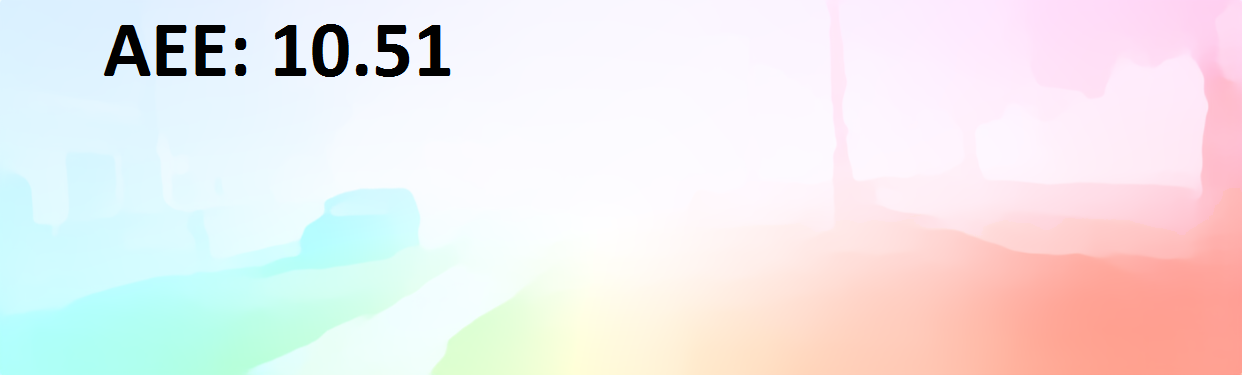}}\hfill
	\subfloat[PWC-Net$^{1}$~\cite{Sun18}]{\includegraphics[width=3.0cm,trim={2cm 0 2cm 0},clip]{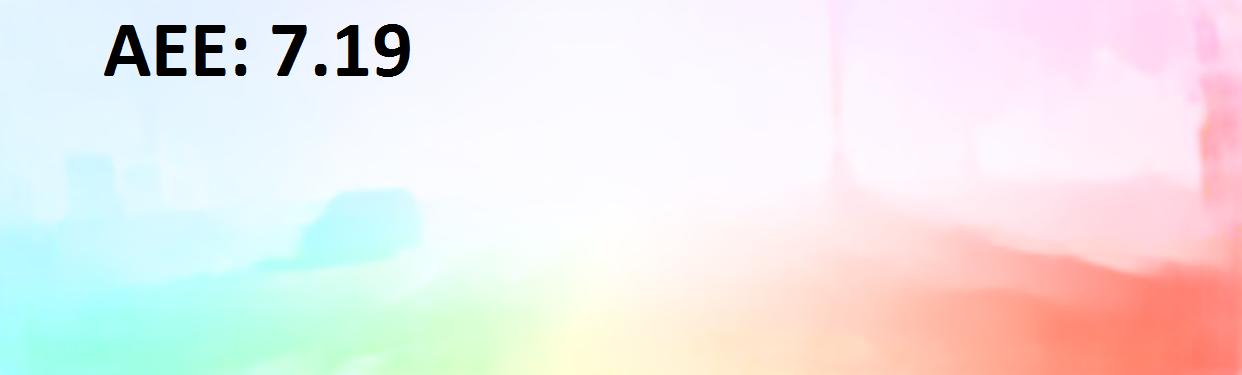}}\hfill
	\subfloat[LiteFlowNet~\cite{Hui18}]{\includegraphics[width=3.0cm,trim={2cm 0 2cm 0},clip]{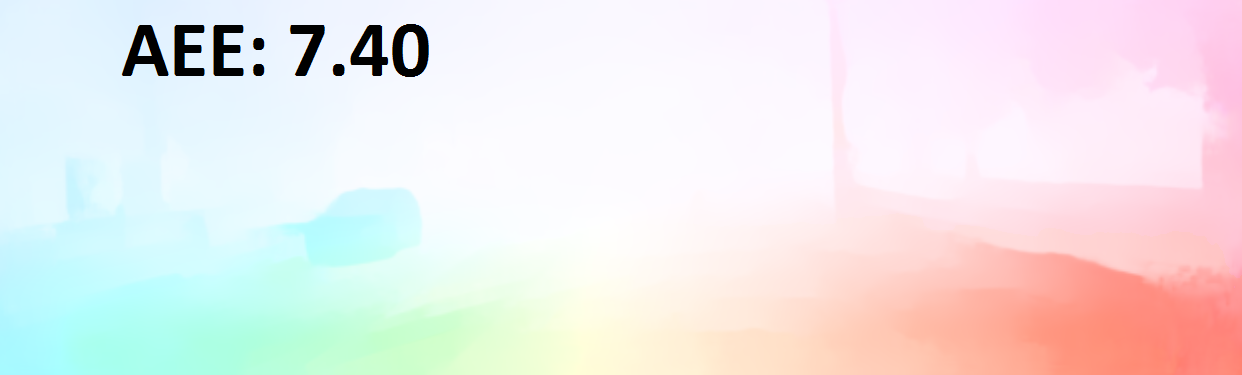}}\hfill
	\subfloat[LiteFlowNet2]{\includegraphics[width=3.0cm,trim={2cm 0 2cm 0},clip]{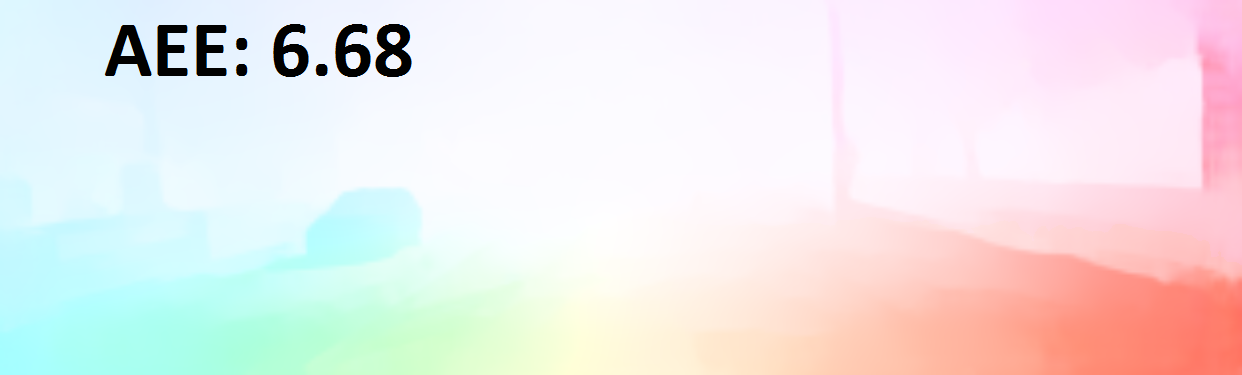}}\\
  
     \includegraphics[width=3.0cm,trim={2cm 0 2cm 0},clip]{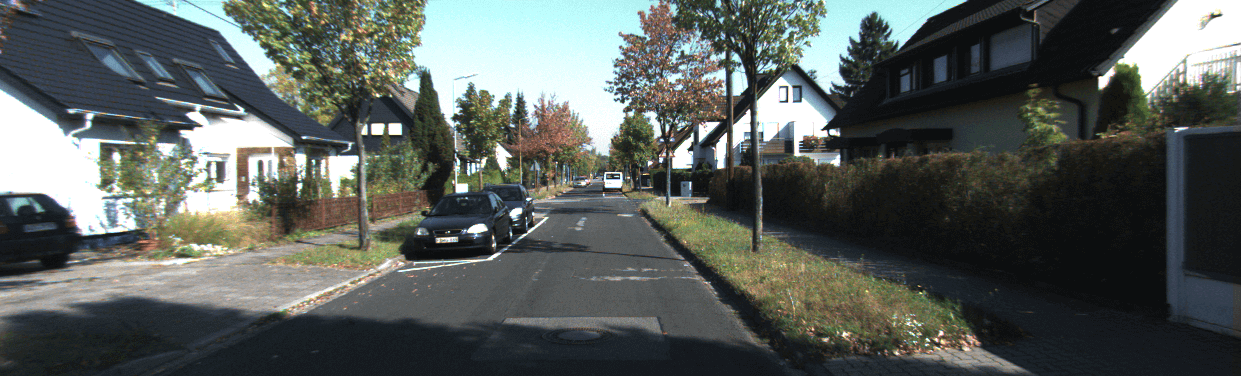}\hfill
     \includegraphics[width=3.0cm,trim={2cm 0 2cm 0},clip]{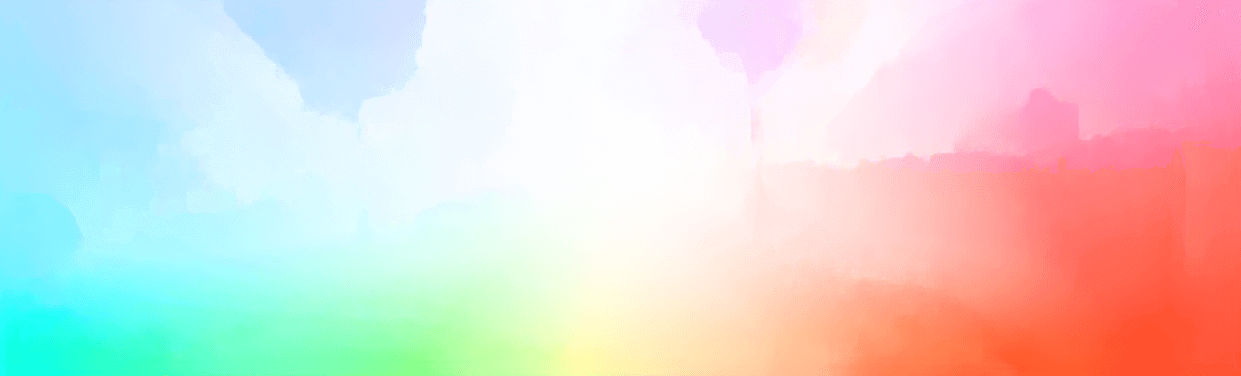}\hfill
     \includegraphics[width=3.0cm,trim={2cm 0 2cm 0},clip]{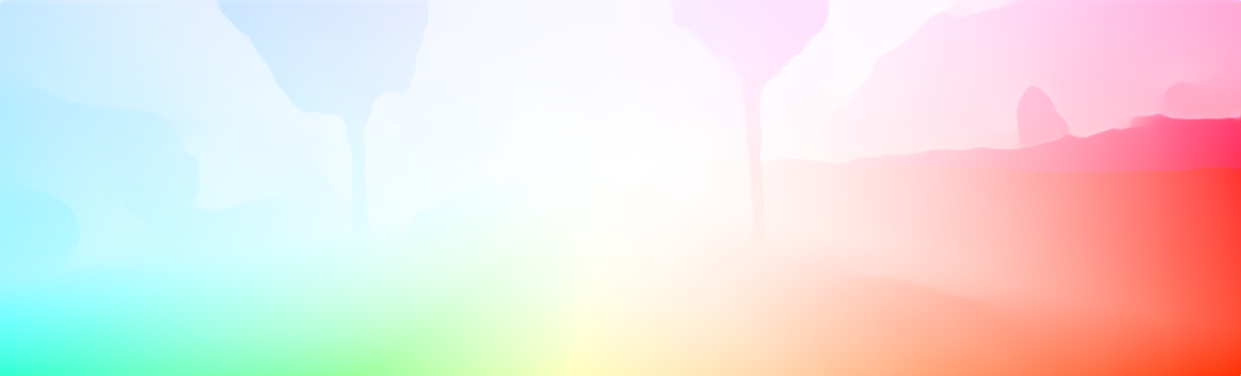}\hfill
     \includegraphics[width=3.0cm,trim={2cm 0 2cm 0},clip]{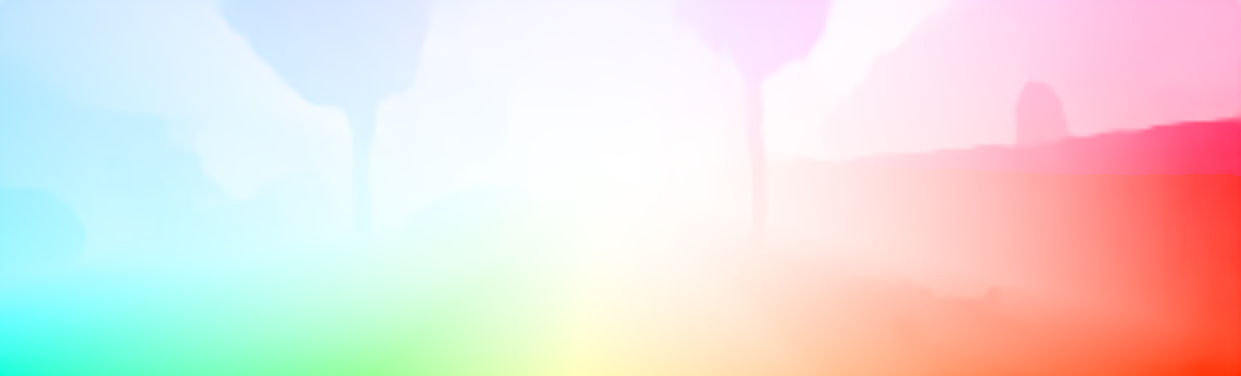}\hfill
     \includegraphics[width=3.0cm,trim={2cm 0 2cm 0},clip]{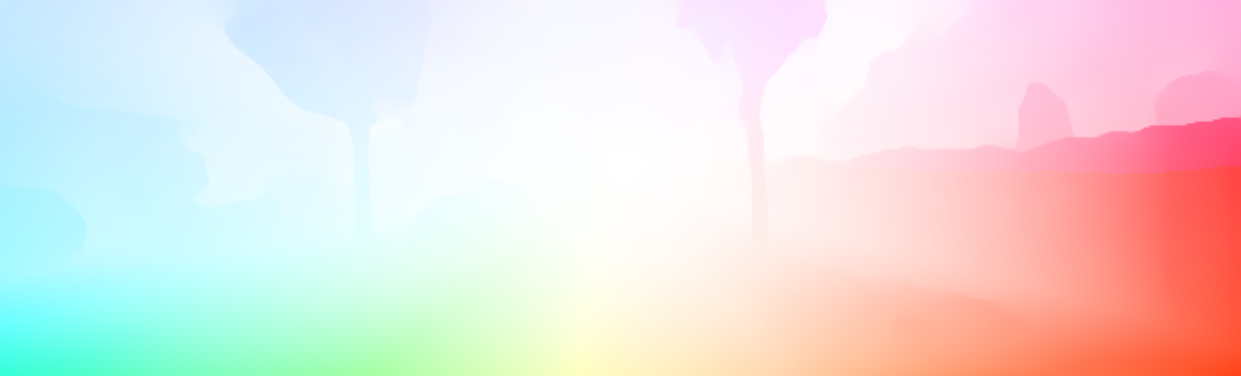}\hfill
     \includegraphics[width=3.0cm,trim={2cm 0 2cm 0},clip]{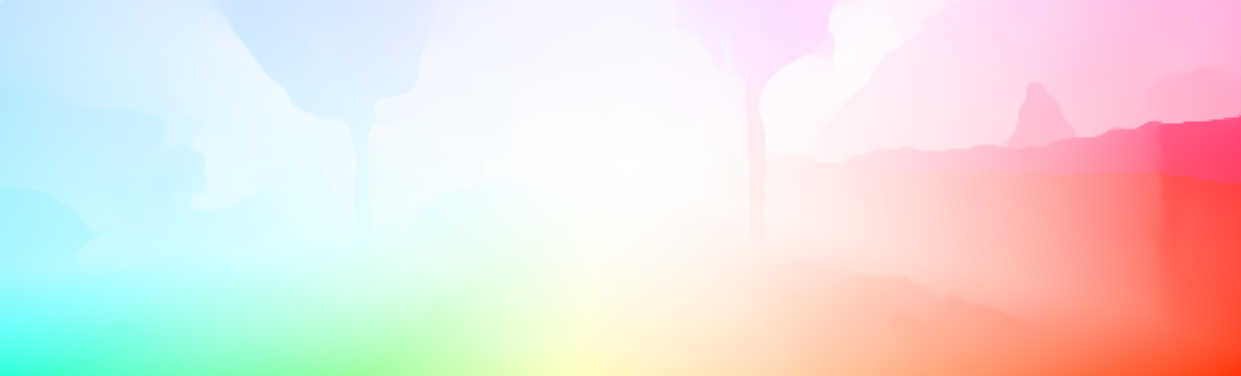}\\
   
     \subfloat[First image]{\includegraphics[width=3.0cm,trim={2cm 0 2cm 0},clip]{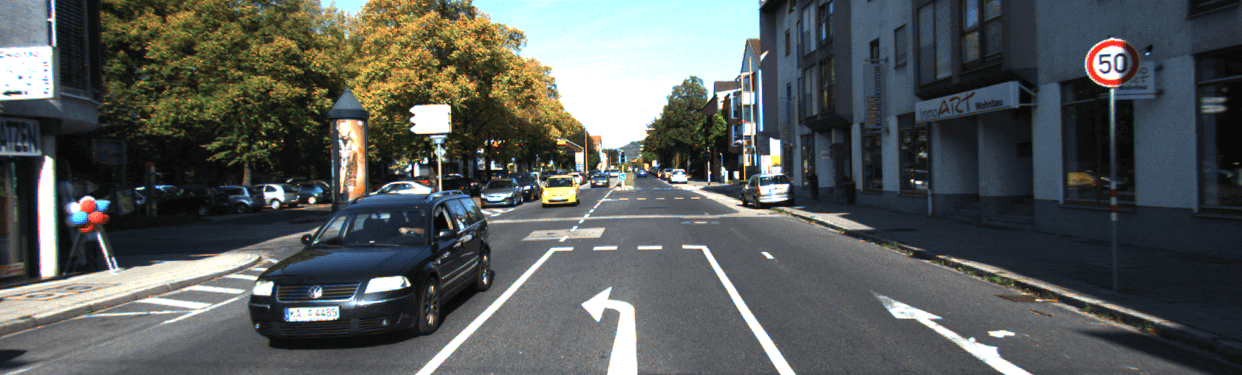}}\hfill
     \subfloat[SPyNet-ft~\cite{Ranjan17}]{\includegraphics[width=3.0cm,trim={2cm 0 2cm 0},clip]{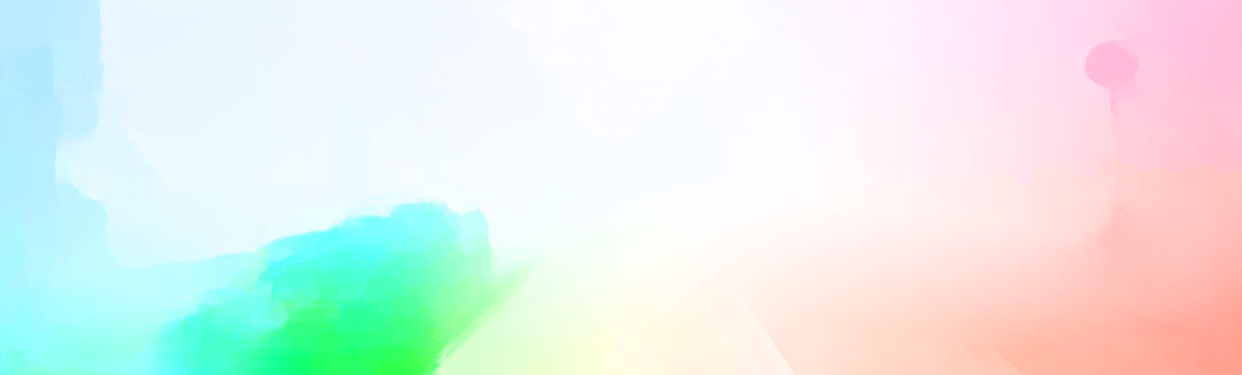}}\hfill
     \subfloat[FlowNet2-ft~\cite{Ilg17}]{\includegraphics[width=3.0cm,trim={2cm 0 2cm 0},clip]{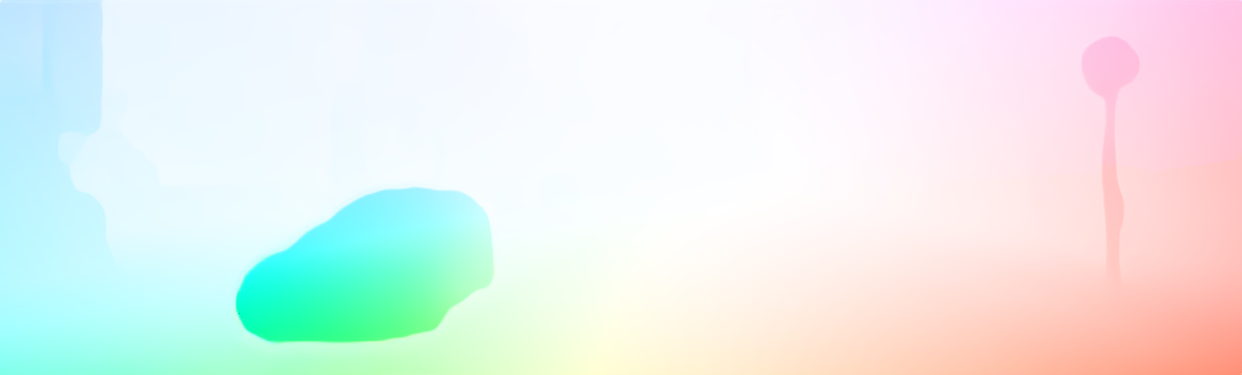}}\hfill
     \subfloat[PWC-Net+~\cite{Sun19}]{\includegraphics[width=3.0cm,trim={2cm 0 2cm 0},clip]{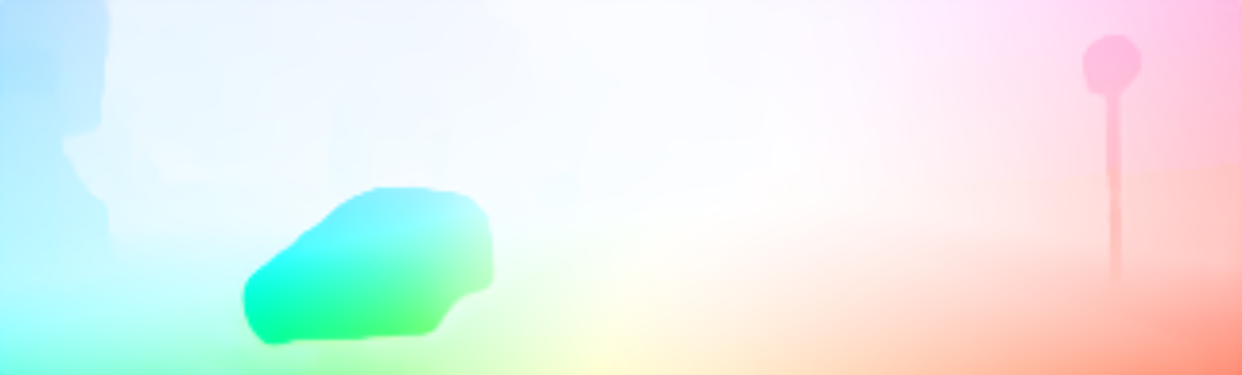}}\hfill
     \subfloat[LiteFlowNet-ft~\cite{Hui18}]{\includegraphics[width=3.0cm,trim={2cm 0 2cm 0},clip]{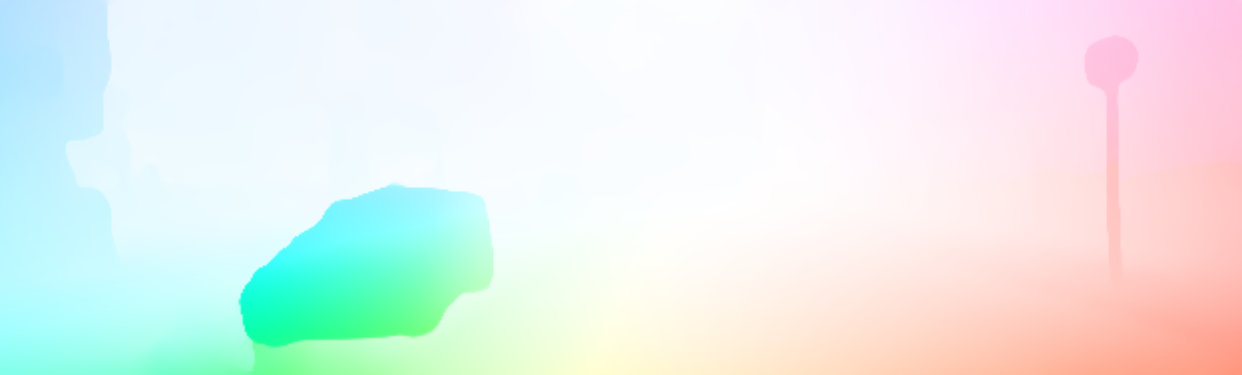}}\hfill
     \subfloat[LiteFlowNet2-ft]{\includegraphics[width=3.0cm,trim={2cm 0 2cm 0},clip]{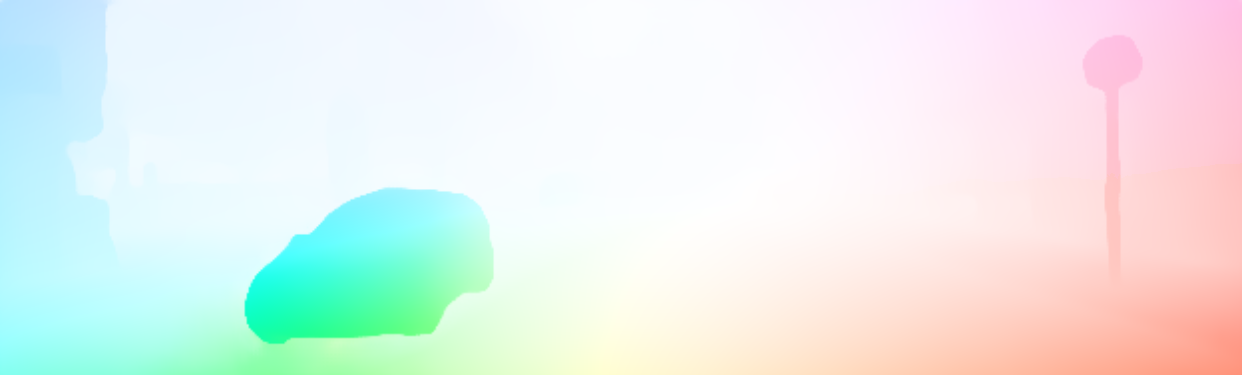}}\\
\end{tabular}
\end{center}
\vspace{-0.3cm}
\caption{Examples of flow fields from different methods on the KITTI 2012 and 2015 training sets (2012: first to second rows, 2015: third to fourth row) and testing sets (2012: fifth row, 2015: last row). For the best visual comparison, it is recommended to enlarge the figure electronically. (Note: $^{1}$At the time of submission, the authors~\cite{Sun18} only release the trained model of PWC-Net that uses a larger feature encoder (overall footprint: 9.37M vs 8.75M) and has a slower runtime (41.12ms vs 39.63ms) trained on Chairs~$\rightarrow$~Things3D.)}
\label{fig:KITTI flows}
\vspace{-0.5em}
\end{figure*}

\vspace{0.1cm}
\noindent
\textbf{KITTI.} LiteFlowNet consistently performs better than LiteFlowNet-pre especially on KITTI 2015 as shown in Table~\ref{tab:results}. It also outperforms SPyNet~\cite{Ranjan17}, FlowNet2-S~\cite{Ilg17}, and FlowNet2-C~\cite{Ilg17}. LiteFlowNet2, the successor of LiteFlowNet, outperforms FlowNet2~\cite{Ilg17}, LiteFlowNet, and PWC-Net~\cite{Sun18} as well.
We also fine-tuned LiteFlowNet (\textbf{LiteFlowNet-ft}) and LiteFlowNet2 (\textbf{LiteFlowNet2-ft}) on a mixture of KITTI 2012 and KITTI 2015 training data using the same augmentation and training schedule as the case of Sintel except that we reduced the amount of augmentation for spatial motion~\cite{Sun18} to fit the driving scene. The height of each image in KITTI dataset is less than that of Sintel about 100 pixels. We randomly crop 896$\times$320 patches to maintain a similar patch area as Sintel and use a batch size of 4. We have experienced that training on KITTI is more challenging than Sintel not only because the training set of KITTI 2012 and KITTI 2015 contains just less than 400 image pairs but also the flow labels are sparse. The insufficient number of per-pixel flow labels greatly affect the performance of the flow network. When fine-tuning LiteFlowNet2 on KITTI, we upsample the constructed flow fields by a factor of 2 in each pyramid level. This effectively increases the number of per-pixel flow labels available. Table~\ref{table:improvements on KITTI} summarizes the improvements in terms of AEE under different network and training configurations.
After fine-tuning, LiteFlowNet and LiteFlowNet2 generalize well to real-world data. LiteFlowNet-ft outperforms all the compared conventional and hybrid methods by a large extent. It also outperforms FlowNet2-ft-kitti~\cite{Ilg17} and PWC-Net-ft~\cite{Sun18}. With the improved architecture and training protocol, LiteFlowNet2-ft outperforms LiteFlowNet, PWC-Net-ft,  and PWC-Net+~\cite{Sun19}. 
Figure~\ref{fig:KITTI flows} shows some examples of flow fields on the training and testing sets KITTI 2012 and KITTI 2012. As in the case for Sintel, LiteFlowNet(-ft), and LiteFlowNet2(-ft) perform the best among the compared methods. Even though LiteFlowNet and LiteFlowNet2 perform pyramidal descriptor matching in a limited searching range, it yields reliable large-displacement flow fields for real-world data due to the feature warping (f-warp) layer introduced. An ablation study of different components in LiteFlowNet will be presented in Section~\ref{sec:ablation study}.

\begin{table}[t]
\small
\centering
\caption{AEE of LiteFlowNet2 fine-tuned on KITTI under different configurations. Out-Noc (or Out-All): Percentage of erroneous pixels in non-occluded areas (or in total). Fl-bg (or Fl-fg): Percentage of optical flow outliers averaged only over background (or foreground) regions. (Note: $^{1}$Comparing to LiteFlowNet~\cite{Hui18}, LiteFlowNet2 uses a simplified (pseudo) network structure for flow inference and regularization at level 2 on KITTI.)} \label{table:improvements on KITTI}
\scalebox{0.80}{
\begin{tabular}{|c|c|c|c|c|}
\hline
\multicolumn{1}{|c|}{}
&\multicolumn{1}{c|}{\cite{Hui18}} 
&\multicolumn{3}{c|}{LiteFlowNet2} \\
\hline
\multicolumn{1}{|l|}{Flow levels up to level 3}
&\multicolumn{1}{c|}{\xmark}
&\multicolumn{1}{c|}{\cmark}
&\multicolumn{1}{c|}{\xmark}
&\multicolumn{1}{c|}{\xmark} \\

\multicolumn{1}{|l|}{Flow levels up to (pseudo)$^{1}$ level 2}
&\multicolumn{1}{c|}{\cmark}
&\multicolumn{1}{c|}{\xmark}
&\multicolumn{1}{c|}{\cmark}
&\multicolumn{1}{c|}{\cmark} \\

\multicolumn{1}{|l|}{Double GT resolution at each level}
&\multicolumn{1}{c|}{\xmark}
&\multicolumn{1}{c|}{\xmark}
&\multicolumn{1}{c|}{\xmark}
&\multicolumn{1}{c|}{\cmark} \\
\hline
\multicolumn{1}{|l|}{KITTI 2012: Train}
&\multicolumn{1}{c|}{(1.05)}
&\multicolumn{1}{c|}{(1.07)}
&\multicolumn{1}{c|}{(1.00)}
&\multicolumn{1}{c|}{(\textbf{0.95})} \\

\multicolumn{1}{|l|}{~~~~~~~~~~~~~~~~~~~~Test (Out-Noc)}
&\multicolumn{1}{c|}{3.27\%}
&\multicolumn{1}{c|}{3.07\%}
&\multicolumn{1}{c|}{2.72\%}
&\multicolumn{1}{c|}{\textbf{2.63\%}} \\

\multicolumn{1}{|l|}{~~~~~~~~~~~~~~~~~~~~Test (Out-all)}
&\multicolumn{1}{c|}{7.27\%}
&\multicolumn{1}{c|}{6.92\%}
&\multicolumn{1}{c|}{6.30\%}
&\multicolumn{1}{c|}{\textbf{6.16\%}} \\

\multicolumn{1}{|l|}{~~~~~~~~~~~~~~~~~~~~Test (Avg-all)}
&\multicolumn{1}{c|}{1.6}
&\multicolumn{1}{c|}{1.5}
&\multicolumn{1}{c|}{\textbf{1.4}}
&\multicolumn{1}{c|}{\textbf{1.4}} \\
\hline
\multicolumn{1}{|l|}{KITTI 2015: Train}
&\multicolumn{1}{c|}{(1.62)}
&\multicolumn{1}{c|}{(1.61)}
&\multicolumn{1}{c|}{(1.47)}
&\multicolumn{1}{c|}{(\textbf{1.33})} \\

\multicolumn{1}{|l|}{~~~~~~~~~~~~~~~~~~~~Train (Fl-all)}
&\multicolumn{1}{c|}{(5.58\%)}
&\multicolumn{1}{c|}{(5.57\%)}
&\multicolumn{1}{c|}{(4.80\%)}
&\multicolumn{1}{c|}{(\textbf{4.32\%})} \\
 
\multicolumn{1}{|l|}{~~~~~~~~~~~~~~~~~~~~Test (Fl-bg)}
&\multicolumn{1}{c|}{9.66\%}
&\multicolumn{1}{c|}{8.72\%}
&\multicolumn{1}{c|}{7.85\%}
&\multicolumn{1}{c|}{\textbf{7.62\%}} \\

\multicolumn{1}{|l|}{~~~~~~~~~~~~~~~~~~~~Test (Fl-fg)}
&\multicolumn{1}{c|}{7.99\%}
&\multicolumn{1}{c|}{8.20\%}
&\multicolumn{1}{c|}{\textbf{7.20\%}}
&\multicolumn{1}{c|}{7.64\%} \\

\multicolumn{1}{|l|}{~~~~~~~~~~~~~~~~~~~~Test (Fl-all)}
&\multicolumn{1}{c|}{9.38\%}
&\multicolumn{1}{c|}{8.63\%}
&\multicolumn{1}{c|}{7.74\%}
&\multicolumn{1}{c|}{\textbf{7.62\%}} \\
\hline 
\end{tabular}}
\vspace{-0.5em}
\end{table}

\vspace{0.1cm}
\noindent
\textbf{LiteFlowNet-CVPR18~\cite{Hui18} vs LiteFlowNet-arXiv~\cite{Hui18-arxiv}.}
In the arXiv version of LiteFlowNet, we excluded a small amount of training data in Things3D undergoing extremely large flow displacement as it is rare to exist in real-world data. On the respective training set, AEE can be improved from 2.52 to 2.48 on Sintel Clean, 4.05 to 4.04 on Sintel Final, 4.25 to 4.00 on KITTI 2012, and 10.46 to 10.39 (Fl-all: 29.30\% to 28.50\%) on KITTI 2015.
For fine-tuning on Sintel, we removed additive noise but introduced image mirroring during data augmentation as~\cite{Sun18}. AEE on the testing set can be improved from 4.86 to 4.54 for the Clean pass and 6.09 to 5.38 for the Final pass. 
For fine-tuning on KITTI, we further reduced the amount of augmentation for spatial motion as\cite{Sun18}. On the respective testing set, AEE can be improved from 1.7 to 1.6 on KITTI 2012 and Fl-all can be improved from 10.24\% to 9.38\% on KITTI 2015. 

\vspace{0.1cm}
\noindent
\textbf{An I/O requirement on LMDB generation.} We use the modified Caffe package~\cite{Dosovitskiy15} to train and test our optical flow networks. The LMDB script requires all image pairs and flow fields to have the same spatial dimension. We knew the I/O requirement as early as our previous CVPR 2018 work (LiteFlowNet)~\cite{Hui18}. There are five types of spatial dimensions in the combined training sets of KITTI 2012 and KITTI 2015, namely $1224\times370$, $1226 \times370$, $1238\times374$, $1241\times376$, and $1242\times375$. In order to fulfill the requirement, all the images and flow fields are cropped to $1224\times370$ before generating LMDB files. A recent work~\cite{Sun19} reports the I/O requirement using Caffe and regards the previous improper usage~\cite{Sun18} as an I/O bug.

\begin{table*}[ht]
\small
\centering
\caption{Number of training parameters and runtime. The model for which the runtime is in parentheses is measured using Torch, and hence are not directly comparable to the others using Caffe. (Note: $^{1}$The runtime is longer when comparing to the value provided by the authors~\cite{Sun19} because it was measured by a faster NVIDIA TITAN Xp GPU than ours.)} \label{tab:model size and runtime}
\scalebox{0.85}{
\begin{tabular}{cccccccc}
\hline

\multicolumn{1}{|c|}{}  				&\multicolumn{2}{c|}{Shallow}        
											&\multicolumn{5}{c|}{Deep} \\ 
\hline
\multicolumn{1}{|l|}{Model} 		     &\multicolumn{1}{c|}{FlowNetC~\cite{Dosovitskiy15}}
											&\multicolumn{1}{c|}{SPyNet~\cite{Ranjan17}}
											&\multicolumn{1}{c|}{FlowNet2~\cite{Ilg17}}
											&\multicolumn{1}{c|}{PWC-Net+~\cite{Sun19}}
											&\multicolumn{1}{c|}{LiteFlowNetX~\cite{Hui18}}
											&\multicolumn{1}{c|}{LiteFlowNet~\cite{Hui18}}
											&\multicolumn{1}{c|}{LiteFlowNet2} \\
\hline
\multicolumn{1}{|l|}{Number of learnable layers} 	&\multicolumn{1}{c|}{26}
														&\multicolumn{1}{c|}{35}
														&\multicolumn{1}{c|}{115}
														&\multicolumn{1}{c|}{59}
														&\multicolumn{1}{c|}{69}
														&\multicolumn{1}{c|}{94} 
														&\multicolumn{1}{c|}{91} \\
																		
\multicolumn{1}{|l|}{Number of parameters (M)}	&\multicolumn{1}{c|}{39.16}
														&\multicolumn{1}{c|}{1.20}
														&\multicolumn{1}{c|}{162.49}
														&\multicolumn{1}{c|}{8.75}
														&\multicolumn{1}{c|}{0.90}
														&\multicolumn{1}{c|}{5.37}
														&\multicolumn{1}{c|}{6.42}\\
											 
\multicolumn{1}{|l|}{Runtime (ms)} 	&\multicolumn{1}{c|}{31.51}
											&\multicolumn{1}{c|}{(129.83)}
											&\multicolumn{1}{c|}{121.49}
											&\multicolumn{1}{c|}{39.63$^{1}$}
											&\multicolumn{1}{c|}{35.10}
											&\multicolumn{1}{c|}{88.53}
											&\multicolumn{1}{c|}{39.69} \\
											
\multicolumn{1}{|l|}{Frame/second (fps)} 	&\multicolumn{1}{c|}{31}
													&\multicolumn{1}{c|}{(8)}
													&\multicolumn{1}{c|}{8}
													&\multicolumn{1}{c|}{25}
													&\multicolumn{1}{c|}{28}
													&\multicolumn{1}{c|}{12}
													&\multicolumn{1}{c|}{25} \\		
																				
\hline
\end{tabular}}
\end{table*}

%-------------------------------------------------------------------------
\subsection{Runtime and Number of Parameters}
\label{sec:runtime}
%-------------------------------------------------------------------------
We measure runtime on a machine equipped with an Intel Xeon E5 2.2GHz and an NVIDIA GTX 1080. Timings are averaged over 100 runs for a Sintel image pair with size $1024\times436$. For a fair comparison, we also exclude the reading and writing time as PWC-Net(+)~\cite{Sun18, Sun19}.
As summarized in Table~\ref{tab:model size and runtime}, 
\begin{itemize}
   \item LiteFlowNet requires \textbf{30.3x fewer} parameters than FlowNet2~\cite{Ilg17} and is \textbf{1.4x faster} in the runtime. It requires \textbf{1.6x fewer} parameters ($\downarrow$~3.4M) than PWC-Net+.
   \item LiteFlowNetX, a small-model variant of LiteFlowNet, which has no descriptor matching requires \textbf{43.5x fewer} parameters than FlowNetC~\cite{Dosovitskiy15} and has a comparable runtime. It has \textbf{1.3x fewer} parameters than SPyNet~\cite{Ranjan17}.    
   \item LiteFlowNet2 requires \textbf{25.3x fewer} parameters than FlowNet2 while being \textbf{3.1x faster}. It is \textbf{2.2 times faster} than LiteFlowNet. In comparison to PWC-Net+, LiteFlowNet2 requires \textbf{1.4x fewer} parameters ($\downarrow$~2.3M). Its processing frequency can reach up to 25 flow fields per second and is similar to PWC-Net+. 
\end{itemize}

%-------------------------------------------------------------------------
\subsection{Ablation Study}
\label{sec:ablation study}
%-------------------------------------------------------------------------
\begin{figure*}[ht]
\centering
   \includegraphics[width=4.4cm]{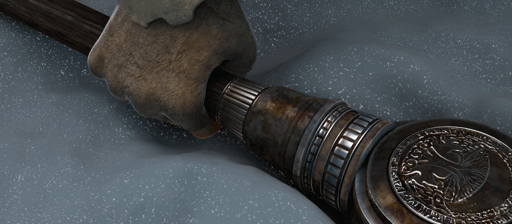}
   \includegraphics[width=4.4cm]{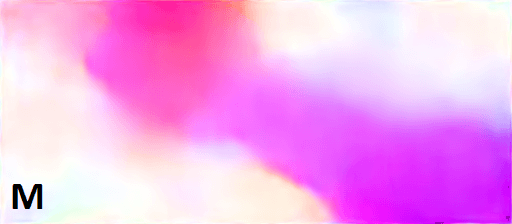} 
   \includegraphics[width=4.4cm]{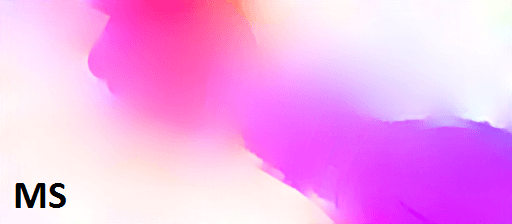} 
   \includegraphics[width=4.4cm]{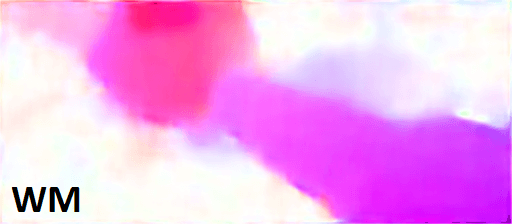}\\
   \includegraphics[width=4.4cm]{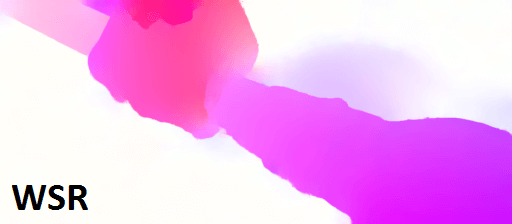} 
   \includegraphics[width=4.4cm]{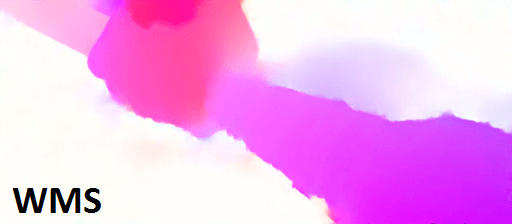}
   \includegraphics[width=4.4cm]{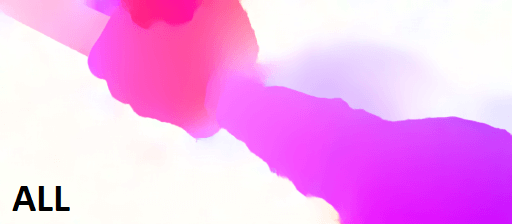}
   \includegraphics[width=4.4cm]{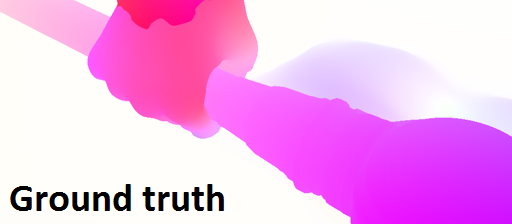}
\caption{Examples of flow fields from different variants of LiteFlowNet trained on Chairs with some of the components disabled. LiteFlowNet is denoted as ``All''. W $=$ Feature \textbf{W}arping, M $=$ Descriptor \textbf{M}atching, S $=$ \textbf{S}ub-Pixel Refinement, R $=$ \textbf{R}egularization.}
\label{fig:ablation flows}
\vspace{-0.5em}
\end{figure*}

\begin{table}[t]
\small
\centering
\caption{AEE of different variants of LiteFlowNet trained on Chairs dataset with some of the components disabled.} \label{tab:ablation study}
\scalebox{0.85}{
\begin{tabular}{|c|c|c|c|c|c|c|}
\hline
\multicolumn{1}{|l|}{Variants}
&\multicolumn{1}{c|}{M}
&\multicolumn{1}{c|}{MS}
&\multicolumn{1}{c|}{WM}
&\multicolumn{1}{c|}{WSR}
&\multicolumn{1}{c|}{WMS}
&\multicolumn{1}{c|}{ALL}  \\
\hline
\multicolumn{1}{|l|}{Feature \textbf{W}arping}
&\multicolumn{1}{c|}{\xmark}
&\multicolumn{1}{c|}{\xmark}
&\multicolumn{1}{c|}{\cmark}
&\multicolumn{1}{c|}{\cmark}
&\multicolumn{1}{c|}{\cmark}
&\multicolumn{1}{c|}{\cmark} \\

\multicolumn{1}{|l|}{Descriptor \textbf{M}atching}
&\multicolumn{1}{c|}{\cmark}
&\multicolumn{1}{c|}{\cmark}
&\multicolumn{1}{c|}{\cmark}
&\multicolumn{1}{c|}{\xmark}
&\multicolumn{1}{c|}{\cmark}
&\multicolumn{1}{c|}{\cmark} \\

\multicolumn{1}{|l|}{\textbf{S}ub-pixel Refinement}
&\multicolumn{1}{c|}{\xmark}
&\multicolumn{1}{c|}{\cmark}
&\multicolumn{1}{c|}{\xmark}
&\multicolumn{1}{c|}{\cmark}
&\multicolumn{1}{c|}{\cmark}
&\multicolumn{1}{c|}{\cmark} \\

\multicolumn{1}{|l|}{\textbf{R}egularization}
&\multicolumn{1}{c|}{\xmark}
&\multicolumn{1}{c|}{\xmark}
&\multicolumn{1}{c|}{\xmark}
&\multicolumn{1}{c|}{\cmark}
&\multicolumn{1}{c|}{\xmark}
&\multicolumn{1}{c|}{\cmark} \\
\hline
\multicolumn{1}{|l|}{FlyingChairs (train)}
&\multicolumn{1}{c|}{3.75}
&\multicolumn{1}{c|}{2.70}
&\multicolumn{1}{c|}{2.98}
&\multicolumn{1}{c|}{1.63}
&\multicolumn{1}{c|}{1.82}
&\multicolumn{1}{c|}{\textbf{1.57}} \\

\multicolumn{1}{|l|}{Sintel clean (train)}
&\multicolumn{1}{c|}{4.70}
&\multicolumn{1}{c|}{4.17}
&\multicolumn{1}{c|}{3.54}
&\multicolumn{1}{c|}{3.19}
&\multicolumn{1}{c|}{2.90}
&\multicolumn{1}{c|}{\textbf{2.78}} \\

\multicolumn{1}{|l|}{Sintel final (train)}
&\multicolumn{1}{c|}{5.69}
&\multicolumn{1}{c|}{5.30}
&\multicolumn{1}{c|}{4.81}
&\multicolumn{1}{c|}{4.63}
&\multicolumn{1}{c|}{4.45}
&\multicolumn{1}{c|}{\textbf{4.17}} \\

\multicolumn{1}{|l|}{KITTI 2012 (train)}
&\multicolumn{1}{c|}{9.22}
&\multicolumn{1}{c|}{8.01}
&\multicolumn{1}{c|}{6.17}
&\multicolumn{1}{c|}{5.03}
&\multicolumn{1}{c|}{4.83}
&\multicolumn{1}{c|}{\textbf{4.56}} \\

\multicolumn{1}{|l|}{KITTI 2015 (train)}
&\multicolumn{1}{c|}{18.24}
&\multicolumn{1}{c|}{16.19}
&\multicolumn{1}{c|}{14.52}
&\multicolumn{1}{c|}{13.20}
&\multicolumn{1}{c|}{12.32}
&\multicolumn{1}{c|}{\textbf{11.58}} \\
\hline 
\end{tabular}}
\end{table}
We investigate the role of each component in LiteFlowNet trained on Chairs (\ie LiteFlowNet-pre) by evaluating the performance of different variants with some of the components disabled unless otherwise stated. The AEE results are summarized in Table~\ref{tab:ablation study} and examples of flow fields are illustrated in Figure~\ref{fig:ablation flows}. 

\vspace{0.1cm}
\noindent
\textbf{Feature Warping.} We consider two variants of LiteFlowNet-pre (WM and WMS) and compare them to the counterparts with feature warping disabled (M and MS). Flow fields from M and MS are more vague. Large degradation in AEE is noticed especially for KITTI 2012 ($\downarrow$~33$\%$) and KITTI 2015 ($\downarrow$~25$\%$). With feature warping (f-warp), pyramidal features that are used as inputs to flow inference are closer in appearance to each other. This facilitates flow estimation in subsequent levels by computing residual flows.

\vspace{0.1cm}
\noindent
\textbf{Descriptor Matching.} We evaluate WSR without descriptor matching for which the flow inference part is made as deep as that in the unamended LiteFlowNet-pre (ALL). No noticeable difference between the flow fields from WSR and ALL. Since the maximum displacement of the example flow field is not very large (only 14.7 pixels), accurate flow field can still be yielded from WSR. For evaluation covering a wide range of flow displacement (especially large-displacement benchmark, KITTI), degradation in AEE is noticed for WSR. This suggests that descriptor matching is useful in addressing large-displacement flow. 

\vspace{0.1cm}
\noindent
\textbf{Sub-Pixel Refinement.} The flow field generated from WMS is more crisp and contains more fine details than that generated from WM with sub-pixel refinement disabled. Less small-magnitude flow artifacts (represented by light color on the background) are observed. Besides, WMS achieves smaller AEE. Since descriptor matching establishes pixel-by-pixel correspondence, sub-pixel refinement is necessary to yield detail-preserving flow fields.

\vspace{0.1cm}
\noindent
\textbf{Regularization.} In comparison WMS with regularization disabled to ALL, undesired artifacts exist in homogeneous regions (represented by very dim color on the background) of the flow field generated from WMS. Flow bleeding and vague flow boundaries are observed. Degradation in AEE is also noticed. This suggests that the proposed feature-driven local convolution (f-lconv) plays the vital role to smooth flow field and maintain crisp flow boundaries as regularization term in conventional variational methods.

\begin{table}[t]
\small
\centering
\caption{AEE and runtime of LiteFlowNet2 trained on Chairs under different cost-volume settings. The value in parentheses represents the setting at level 3.}
 \label{tab:cost volume study}
\scalebox{0.85}{
\begin{tabular}{|c|c|c|c|c|}
\hline                          
\multicolumn{1}{|l|}{Searching Range (pixels)}
&\multicolumn{1}{c|}{3}
&\multicolumn{1}{c|}{3 (6)}
&\multicolumn{1}{c|}{4}\\

\multicolumn{1}{|l|}{Stride}
&\multicolumn{1}{c|}{1}
&\multicolumn{1}{c|}{1 (2)}
&\multicolumn{1}{c|}{1}\\

\multicolumn{1}{|l|}{Levels}
&\multicolumn{1}{c|}{6 to 3}
&\multicolumn{1}{c|}{6 to 4 (3)}
&\multicolumn{1}{c|}{6 to 3}\\
\hline
\multicolumn{1}{|l|}{Sintel Clean (train)}
&\multicolumn{1}{c|}{2.73}
&\multicolumn{1}{c|}{2.78}
&\multicolumn{1}{c|}{\textbf{2.71}}\\

\multicolumn{1}{|l|}{Sintel Final (train)}
&\multicolumn{1}{c|}{\textbf{4.14}}
&\multicolumn{1}{c|}{\textbf{4.14}}
&\multicolumn{1}{c|}{\textbf{4.14}}\\

\multicolumn{1}{|l|}{KITTI 2012 (train)}
&\multicolumn{1}{c|}{4.26}
&\multicolumn{1}{c|}{\textbf{4.11}}
&\multicolumn{1}{c|}{4.20}\\

\multicolumn{1}{|l|}{KITTI 2015 (train)}
&\multicolumn{1}{c|}{11.72}
&\multicolumn{1}{c|}{11.31}
&\multicolumn{1}{c|}{\textbf{11.12}}\\
\hline
\multicolumn{1}{|l|}{Runtime (ms)}
&\multicolumn{1}{c|}{41.33}
&\multicolumn{1}{c|}{\textbf{39.69}}
&\multicolumn{1}{c|}{44.33}\\
\hline 
\end{tabular}}
\vspace{-0.5em}
\end{table}

\vspace{0.1cm}
\noindent
\textbf{Searching Range.} We compare three variants of LiteFlowNet2 trained on Chairs using different cost-volume settings as shown in Table~\ref{tab:cost volume study}. On the whole, a larger searching range leads to a lower AEE. The improvement is more significant on large-displacement benchmark, KITTI. Our design that uses a larger searching range together with a sparse cost volume in a high-resolution pyramid level not only improves flow accuracy but also promotes a more efficient computation. We choose the second cost-volume setting for our final models due to the fastest computation time. 

%-------------------------------------------------------------------------
\section{Conclusion}
\label{sec:conclusion}
%-------------------------------------------------------------------------
We have developed a lightweight and effective CNN for addressing the classical problem of optical flow estimation through adopting data fidelity and regularization from variational methods.
LiteFlowNet uses pyramidal feature extraction, feature warping, multi-scale cascaded flow inference, and flow regularization to break the de facto rule of accurate flow network requiring large model size. To address large-displacement and detail-preserving flows, it exploits a multi-scale short-range matching to generate a pixel-level flow field and further improves the estimate to sub-pixel accuracy in each cascaded flow inference. To result crisp flow boundaries, each flow field is adaptively regularized through the feature-driven local convolution. 
The evolution of LiteFlowNet creates LiteFlowNet2, which runs 2.2 times faster and attains a better flow accuracy. LiteFlowNet2 outperforms the state-of-the-art FlowNet2~\cite{Ilg17} on Sintel and KITTI benchmarks while being 3.1 times faster in the runtime and 25.3 times smaller in the model size. It also outperforms PWC-Net+~\cite{Sun19} on KITTI 2012 and 2015, and is on par with PWC-Net+ on Sintel Clean and Final while being 1.4 times smaller in the model size.
With its lightweight, accurate, and fast flow computation, LiteFlowNet2 can be deployed to many real-time applications such as video processing, motion segmentation, action recognition, SLAM, 3D reconstruction, and more.

%-------------------------------------------------------------------------
\appendix
\section{Appendix -- Network Details of LiteFlowNet2}
\label{sec:appendix}
%-------------------------------------------------------------------------
%
LiteFlowNet2 consists of two compact sub-networks, namely NetC and NetE. NetC is a two-steam sub-network in which the two streams share the same set of filters. The input to NetC is an image pair ($I_{1}$, $I_{2}$). The network architectures of the 6-level NetC and NetE at pyramid level 5 are provided in Table \ref{tab:NetC} and Tables \ref{tab:NetE-C-L5} to \ref{tab:NetE-R-L5}, respectively. We use suffixes ``M'', ``S'' and ``R'' to highlight the layers that are used in descriptor matching, sub-pixel refinement, and flow regularization modules in NetE, respectively. The name of convolution layer is replaced from ``\texttt{conv}'' to ``\texttt{flow}'' to highlight when the output is a flow field. 

\begin{table*}[ht]
\small
\centering
\caption{The network details of NetC. ``\# Ch. In / Out'' means the number of input or output channels of the feature maps. ``\texttt{conv}'' denotes convolution.} \label{tab:NetC}
\scalebox{1}{
\begin{tabular}{cccccc}
\hline

\multicolumn{1}{|c|}{Layer name} 			
&\multicolumn{1}{c|}{Kernel}
&\multicolumn{1}{c|}{Stride}									
&\multicolumn{1}{c|}{\# Ch. In / Out}
&\multicolumn{1}{c|}{Input} \\
\hline
\multicolumn{1}{|c|}{\texttt{conv1}}			
&\multicolumn{1}{c|}{7~$\times$~7}
&\multicolumn{1}{c|}{1}
&\multicolumn{1}{c|}{3 / 32} 
&\multicolumn{1}{c|}{$I_{1}$ or $I_{2}$} \\
																		
\multicolumn{1}{|c|}{\texttt{conv2\_1}}			
&\multicolumn{1}{c|}{3~$\times$~3}
&\multicolumn{1}{c|}{2}
&\multicolumn{1}{c|}{32 / 32}
&\multicolumn{1}{c|}{\texttt{conv1}} \\
											 
\multicolumn{1}{|c|}{\texttt{conv2\_2}	}		
&\multicolumn{1}{c|}{3~$\times$~3}
&\multicolumn{1}{c|}{1}
&\multicolumn{1}{c|}{32 / 32}
&\multicolumn{1}{c|}{\texttt{conv2\_1}} \\
		
\multicolumn{1}{|c|}{\texttt{conv2\_3}}			
&\multicolumn{1}{c|}{3~$\times$~3}
&\multicolumn{1}{c|}{1}
&\multicolumn{1}{c|}{32 / 32}
&\multicolumn{1}{c|}{\texttt{conv2\_2}} \\

\multicolumn{1}{|c|}{\texttt{conv3\_1}}			
&\multicolumn{1}{c|}{3~$\times$~3}
&\multicolumn{1}{c|}{2}
&\multicolumn{1}{c|}{32 / 64}
&\multicolumn{1}{c|}{\texttt{conv2\_3}} \\																							
\multicolumn{1}{|c|}{\texttt{conv3\_2}}			
&\multicolumn{1}{c|}{3~$\times$~3}
&\multicolumn{1}{c|}{1}
&\multicolumn{1}{c|}{64 / 64}
&\multicolumn{1}{c|}{\texttt{conv3\_1}} \\

\multicolumn{1}{|c|}{\texttt{conv4\_1}}			
&\multicolumn{1}{c|}{3~$\times$~3}
&\multicolumn{1}{c|}{2}
&\multicolumn{1}{c|}{64 / 96}
&\multicolumn{1}{c|}{\texttt{conv3\_2}} \\																							
\multicolumn{1}{|c|}{\texttt{conv4\_2}}			
&\multicolumn{1}{c|}{3~$\times$~3}
&\multicolumn{1}{c|}{1}
&\multicolumn{1}{c|}{96 / 96}
&\multicolumn{1}{c|}{\texttt{conv4\_1}} \\

\multicolumn{1}{|c|}{\texttt{conv5}}			
&\multicolumn{1}{c|}{3~$\times$~3}
&\multicolumn{1}{c|}{2}
&\multicolumn{1}{c|}{96 / 128}
&\multicolumn{1}{c|}{\texttt{conv4\_2}} \\		
	
\multicolumn{1}{|c|}{\texttt{conv6}}			
&\multicolumn{1}{c|}{3~$\times$~3}
&\multicolumn{1}{c|}{2}
&\multicolumn{1}{c|}{128 / 192}
&\multicolumn{1}{c|}{\texttt{conv5}} \\
\hline
\end{tabular}}
\end{table*}

\begin{table*}[ht]
\small
\centering
\caption{The network details of the descriptor matching unit (M) in NetE at pyramid level 5. ``\texttt{upconv}'', ``\texttt{f-warp}'', ``\texttt{corr}'', and ``\texttt{loss}'' denote the fractionally strided convolution (so-called deconvolution), feature warping, correlation, and the layer where training loss is applied, respectively. Furthermore, ``\texttt{conv5a}' and ``\texttt{conv5b}'' denote the high-dimensional features of images $I_{1}$ and $I_{2}$ generated from NetC at pyramid level 5.} \label{tab:NetE-C-L5}
\scalebox{1}{
\begin{tabular}{cccccc}
\hline

\multicolumn{1}{|c|}{Layer name} 			
&\multicolumn{1}{c|}{Kernel}
&\multicolumn{1}{c|}{Stride}									
&\multicolumn{1}{c|}{\# Ch. In / Out}
&\multicolumn{1}{c|}{Input} \\
\hline
\multicolumn{1}{|c|}{\texttt{upconv5\_M}}			
&\multicolumn{1}{c|}{4~$\times$~4}
&\multicolumn{1}{c|}{$\frac{1}{2}$}
&\multicolumn{1}{c|}{2 / 2} 
&\multicolumn{1}{c|}{\texttt{flow6\_R}} \\
																		
\multicolumn{1}{|c|}{\texttt{f-warp5\_M}	}		
&\multicolumn{1}{c|}{-}
&\multicolumn{1}{c|}{-}
&\multicolumn{1}{c|}{128, 2 / 128}
&\multicolumn{1}{c|}{\texttt{conv5b},  \texttt{upconv5\_M}} \\

\multicolumn{1}{|c|}{\texttt{corr5\_M}}			
&\multicolumn{1}{c|}{1~$\times$~1}
&\multicolumn{1}{c|}{1}
&\multicolumn{1}{c|}{128, 128 / 49}
&\multicolumn{1}{c|}{\texttt{conv5a}, \texttt{f-warp5\_M}} \\

\multicolumn{1}{|c|}{\texttt{conv5\_1\_M}}			
&\multicolumn{1}{c|}{3~$\times$~3}
&\multicolumn{1}{c|}{1}
&\multicolumn{1}{c|}{49 / 128}
&\multicolumn{1}{c|}{\texttt{corr5\_M}} \\

\multicolumn{1}{|c|}{c\texttt{onv5\_2\_M}}			
&\multicolumn{1}{c|}{3~$\times$~3}
&\multicolumn{1}{c|}{1}
&\multicolumn{1}{c|}{128 / 128}
&\multicolumn{1}{c|}{\texttt{conv5\_1\_M}} \\

\multicolumn{1}{|c|}{\texttt{conv5\_3\_M}}			
&\multicolumn{1}{c|}{3~$\times$~3}
&\multicolumn{1}{c|}{1}
&\multicolumn{1}{c|}{128 / 96}
&\multicolumn{1}{c|}{\texttt{conv5\_2\_M}} \\	
	
\multicolumn{1}{|c|}{\texttt{conv5\_4\_M}}			
&\multicolumn{1}{c|}{3~$\times$~3}
&\multicolumn{1}{c|}{1}
&\multicolumn{1}{c|}{96 / 64}
&\multicolumn{1}{c|}{\texttt{conv5\_3\_M}} \\
	
\multicolumn{1}{|c|}{\texttt{conv5\_5\_M}}			
&\multicolumn{1}{c|}{3~$\times$~3}
&\multicolumn{1}{c|}{1}
&\multicolumn{1}{c|}{64 / 32}
&\multicolumn{1}{c|}{\texttt{conv5\_4\_M}} \\

\multicolumn{1}{|c|}{\texttt{conv5\_6\_M}}			
&\multicolumn{1}{c|}{3~$\times$~3}
&\multicolumn{1}{c|}{1}
&\multicolumn{1}{c|}{32 / 2}
&\multicolumn{1}{c|}{\texttt{conv5\_5\_M}} \\
	
\multicolumn{1}{|c|}{\texttt{flow5\_M}, \texttt{loss5\_M}}			
&\multicolumn{2}{c|}{element-wise sum}
&\multicolumn{1}{c|}{2, 2 / 2}
&\multicolumn{1}{c|}{\texttt{upconv5\_M}, \texttt{conv5\_4\_M}} \\						 
\hline
\end{tabular}}
\end{table*}

\begin{table*}[ht]
\small
\centering
\caption{Network details of sub-pixel refinement module (S) in NetE at pyramid level 5.} \label{tab:NetE-S-L5}
\scalebox{1}{
\begin{tabular}{ccccc}
\hline
\multicolumn{1}{|c|}{Layer name} 			
&\multicolumn{1}{c|}{Kernel}
&\multicolumn{1}{c|}{Stride}									
&\multicolumn{1}{c|}{\# Ch. In / Out}

&\multicolumn{1}{c|}{Input} \\
\hline					
\multicolumn{1}{|c|}{\texttt{f-warp5\_S}}			
&\multicolumn{1}{c|}{-}
&\multicolumn{1}{c|}{-}
&\multicolumn{1}{c|}{128, 2 / 128}
&\multicolumn{1}{c|}{\texttt{conv5b}, \texttt{flow5\_C}} \\

\multicolumn{1}{|c|}{\texttt{conv5\_1\_S}}			
&\multicolumn{1}{c|}{3$\times$3}
&\multicolumn{1}{c|}{1}
&\multicolumn{1}{c|}{258 / 128}
&\multicolumn{1}{c|}{\texttt{concat(conv5a, f-warp5\_S, flow5\_C)}} \\

\multicolumn{1}{|c|}{\texttt{conv5\_2\_S}}			
&\multicolumn{1}{c|}{3~$\times$~3}
&\multicolumn{1}{c|}{1}
&\multicolumn{1}{c|}{128 / 128}
&\multicolumn{1}{c|}{\texttt{conv5\_1\_S}} \\

\multicolumn{1}{|c|}{\texttt{conv5\_3\_S}}			
&\multicolumn{1}{c|}{3~$\times$~3}
&\multicolumn{1}{c|}{1}
&\multicolumn{1}{c|}{128 / 96}
&\multicolumn{1}{c|}{\texttt{conv5\_2\_S}} \\

\multicolumn{1}{|c|}{\texttt{conv5\_4\_S}}			
&\multicolumn{1}{c|}{3~$\times$~3}
&\multicolumn{1}{c|}{1}
&\multicolumn{1}{c|}{96 / 64}
&\multicolumn{1}{c|}{\texttt{conv5\_3\_S}} \\

\multicolumn{1}{|c|}{\texttt{conv5\_5\_S}}			
&\multicolumn{1}{c|}{3~$\times$~3}
&\multicolumn{1}{c|}{1}
&\multicolumn{1}{c|}{64 / 32}
&\multicolumn{1}{c|}{\texttt{conv5\_4\_S}} \\	
	
\multicolumn{1}{|c|}{\texttt{conv5\_6\_S}}			
&\multicolumn{1}{c|}{3~$\times$~3}
&\multicolumn{1}{c|}{1}
&\multicolumn{1}{c|}{32 / 2}
&\multicolumn{1}{c|}{\texttt{conv5\_5\_S}} \\
	
\multicolumn{1}{|c|}{\texttt{flow5\_S}, \texttt{loss5\_S}}			
&\multicolumn{2}{c|}{element-wise sum}
&\multicolumn{1}{c|}{2, 2 / 2}
&\multicolumn{1}{c|}{\texttt{flow5\_C}, \texttt{conv5\_4\_S}} \\						 
\hline
\end{tabular}}
\end{table*}

\begin{table*}[ht]
\small
\centering
\caption{Network details of flow regularization module (R) in NetE at pyramid level 5. ``\texttt{warp}'', ``\texttt{norm}'', ``\texttt{softmax}'', and ``\texttt{f-lcon}'' denote the image warping, L2 norm of the RGB brightness difference between the two input images, normalized exponential operation over each 1~$\times$~1~$\times$~(\#~Ch.~In) column in the 3-D tensor, and feature-driven local convolution, respectively. Furthermore, ``\texttt{conv\_dist}'' highlight the output of the convolution layer is used as the feature-driven distance metric $\mathcal{D}$. ``\texttt{im5a}'' and ``\texttt{im5b}'' denote the down-sized images of $I_{1}$ and $I_{2}$ at pyramidal level 5.} \label{tab:NetE-R-L5}
\scalebox{1}{
\begin{tabular}{ccccc}
\hline

\multicolumn{1}{|c|}{Layer name} 			
&\multicolumn{1}{c|}{Kernel}
&\multicolumn{1}{c|}{Stride}									
&\multicolumn{1}{c|}{\# Ch. In / Out}

&\multicolumn{1}{c|}{Input} \\
\hline				
\multicolumn{1}{|c|}{\texttt{m-flow5\_R}}			
&\multicolumn{2}{c|}{remove mean}
&\multicolumn{1}{c|}{2 / 2}
&\multicolumn{1}{c|}{\texttt{flow5\_S}} \\

\multicolumn{1}{|c|}{\texttt{warp5\_R}}			
&\multicolumn{1}{c|}{-}
&\multicolumn{1}{c|}{-}
&\multicolumn{1}{c|}{3, 2 / 3}
&\multicolumn{1}{c|}{\texttt{im5b}, \texttt{flow5\_S}} \\

\multicolumn{1}{|c|}{\texttt{norm5\_R}}			
&\multicolumn{2}{c|}{L2 norm}
&\multicolumn{1}{c|}{3, 3 / 1}
&\multicolumn{1}{c|}{\texttt{im5a}, \texttt{warp5\_R}} \\

\multicolumn{1}{|c|}{\texttt{conv5\_1\_R}}			
&\multicolumn{1}{c|}{3~$\times$~3}
&\multicolumn{1}{c|}{1}
&\multicolumn{1}{c|}{131 / 128}
&\multicolumn{1}{c|}{\texttt{concat(conv5a, m-flow5\_R, norm5\_R)}} \\

\multicolumn{1}{|c|}{\texttt{conv5\_2\_R}}			
&\multicolumn{1}{c|}{3~$\times$~3}
&\multicolumn{1}{c|}{1}
&\multicolumn{1}{c|}{128 / 128}
&\multicolumn{1}{c|}{\texttt{conv5\_1\_R}} \\

\multicolumn{1}{|c|}{\texttt{conv5\_3\_R}}			
&\multicolumn{1}{c|}{3~$\times$~3}
&\multicolumn{1}{c|}{1}
&\multicolumn{1}{c|}{128 / 64}
&\multicolumn{1}{c|}{\texttt{conv5\_2\_R}} \\	
	
\multicolumn{1}{|c|}{\texttt{conv5\_4\_R}}			
&\multicolumn{1}{c|}{3~$\times$~3}
&\multicolumn{1}{c|}{1}
&\multicolumn{1}{c|}{64 / 64}
&\multicolumn{1}{c|}{\texttt{conv5\_3\_R}} \\	
	
\multicolumn{1}{|c|}{\texttt{conv5\_5\_R}}			
&\multicolumn{1}{c|}{3~$\times$~3}
&\multicolumn{1}{c|}{1}
&\multicolumn{1}{c|}{64 / 32}
&\multicolumn{1}{c|}{\texttt{conv5\_4\_R}} \\	
	
\multicolumn{1}{|c|}{\texttt{conv5\_6\_R}}			
&\multicolumn{1}{c|}{3~$\times$~3}
&\multicolumn{1}{c|}{1}
&\multicolumn{1}{c|}{32 / 32}
&\multicolumn{1}{c|}{\texttt{conv5\_5\_R}} \\	

\multicolumn{1}{|c|}{\texttt{conv5\_dist\_R}}			
&\multicolumn{1}{c|}{3~$\times$~3}
&\multicolumn{1}{c|}{1}
&\multicolumn{1}{c|}{32 / 9}
&\multicolumn{1}{c|}{\texttt{conv5\_6\_R}} \\

\multicolumn{1}{|c|}{\texttt{softmax5\_R}}			
&\multicolumn{1}{c|}{-}
&\multicolumn{1}{c|}{1}
&\multicolumn{1}{c|}{9 / 9}
&\multicolumn{1}{c|}{\texttt{conv5\_dist\_R}} \\

\multicolumn{1}{|c|}{\texttt{f-lcon5\_R}, \texttt{loss5\_R}}			
&\multicolumn{1}{c|}{3~$\times$~3}
&\multicolumn{1}{c|}{1}
&\multicolumn{1}{c|}{9, 2 / 2}
&\multicolumn{1}{c|}{\texttt{softmax5\_R}, \texttt{flow5\_R}} \\					 
\hline
\end{tabular}}
\end{table*}

%-------------------------------------------------------------------------
% references section
%-------------------------------------------------------------------------
% can use a bibliography generated by BibTeX as a .bbl file
% BibTeX documentation can be easily obtained at:
% http://mirror.ctan.org/biblio/bibtex/contrib/doc/
% The IEEEtran BibTeX style support page is at:
% http://www.michaelshell.org/tex/ieeetran/bibtex/
\bibliographystyle{IEEEtran}
% argument is your BibTeX string definitions and bibliography database(s)
\bibliography{egbib}
%
% <OR> manually copy in the resultant .bbl file
% set second argument of \begin to the number of references
% (used to reserve space for the reference number labels box)
%\begin{thebibliography}{1}
%
%\bibitem{IEEEhowto:kopka}
%H.~Kopka and P.~W. Daly, \emph{A Guide to \LaTeX}, 3rd~ed.\hskip 1em plus
%  0.5em minus 0.4em\relax Harlow, England: Addison-Wesley, 1999.
%
%\end{thebibliography}

%
\end{document}